\documentclass{article}

\usepackage[preprint]{neurips_2026}
\usepackage{multirow}
\usepackage{xcolor}
\usepackage{graphicx}
\usepackage{amsmath}
\usepackage{subcaption}

\usepackage{booktabs}

\usepackage[most]{tcolorbox}
\usepackage{enumitem}

\usepackage[table]{xcolor}
\usepackage{pifont}

\newcommand{\cmark}{\textcolor{green!60!black}{\ding{51}}}
\newcommand{\xmark}{\textcolor{red!75!black}{\ding{55}}}
\newcommand{\pmark}{\textcolor{orange!85!black}{\textbf{$\triangle$}}}

\usepackage[utf8]{inputenc} % allow utf-8 input
\usepackage[T1]{fontenc}    % use 8-bit T1 fonts
\usepackage{hyperref}       % hyperlinks
\usepackage{url}            % simple URL typesetting
\usepackage{booktabs}       % professional-quality tables
\usepackage{amsfonts}       % blackboard math symbols
\usepackage{nicefrac}       % compact symbols for 1/2, etc.
\usepackage{microtype}      % microtypography
\usepackage{xcolor}         % colors

\usepackage{algorithm}
\usepackage{algpseudocode}
\usepackage{amssymb}
\usepackage{float}
\usepackage[table]{xcolor}
\usepackage{enumitem}
\usepackage{titletoc}

\usepackage{booktabs}
\usepackage{wrapfig}

\newcommand{\stella}[1]{}
\renewcommand{\stella}[1]{{\color{blue!60}[stella: #1]}}

\usepackage{makecell}

\projecturl{https://aiden0526.github.io/EvoArena/}
\codeurl{https://github.com/Aiden0526/EvoArena}
\dataseturl{https://huggingface.co/collections/Aiden0526/evoarena}

\title{EvoArena: Tracking Memory Evolution\\for Robust LLM Agents in Dynamic Environments}

\vspace{-5em}

\author{%
\textbf{Jundong Xu}$^{1}$\thanks{Equal Contribution} \quad
\textbf{Qingchuan Li}$^{1}$\footnotemark[1] \quad
\textbf{Jiaying Wu}$^{1}$ \quad
\textbf{Yihuai Lan}$^{2}$ \\
\textbf{Shuyue Stella Li}$^{3}$ \quad
\textbf{Huichi Zhou}$^{4}$ \quad
\textbf{Bowen Jiang}$^{5}$ \quad
\textbf{Lei Wang}$^{2}$ \quad
\textbf{Jun Wang}$^{4}$ \\
\textbf{Anh Tuan Luu}$^{6}$ \quad
\textbf{Caiming Xiong}$^{7}$ \thanks{Work Partially Done at Salesforce AI Research} \quad
\textbf{Hae Won Park}$^{8}$ \quad
\textbf{Bryan Hooi}$^{1}$ \quad
\textbf{Zhiyuan Hu}$^{1,8}$ \\
\normalfont
$^1$National University of Singapore \quad
$^2$Singapore Management University \\
\normalfont
$^3$University of Washington \quad
$^4$University College London \quad
$^5$University of Pennsylvania \\
\normalfont
$^6$Nanyang Technological University \quad
$^7$Recursive \quad
$^8$Massachusetts Institute of Technology
}

\begin{document}

\maketitle

\vspace{-5mm}

\begin{figure}[H]
\centering
\vspace{-1em}
\includegraphics[width=0.9\linewidth]{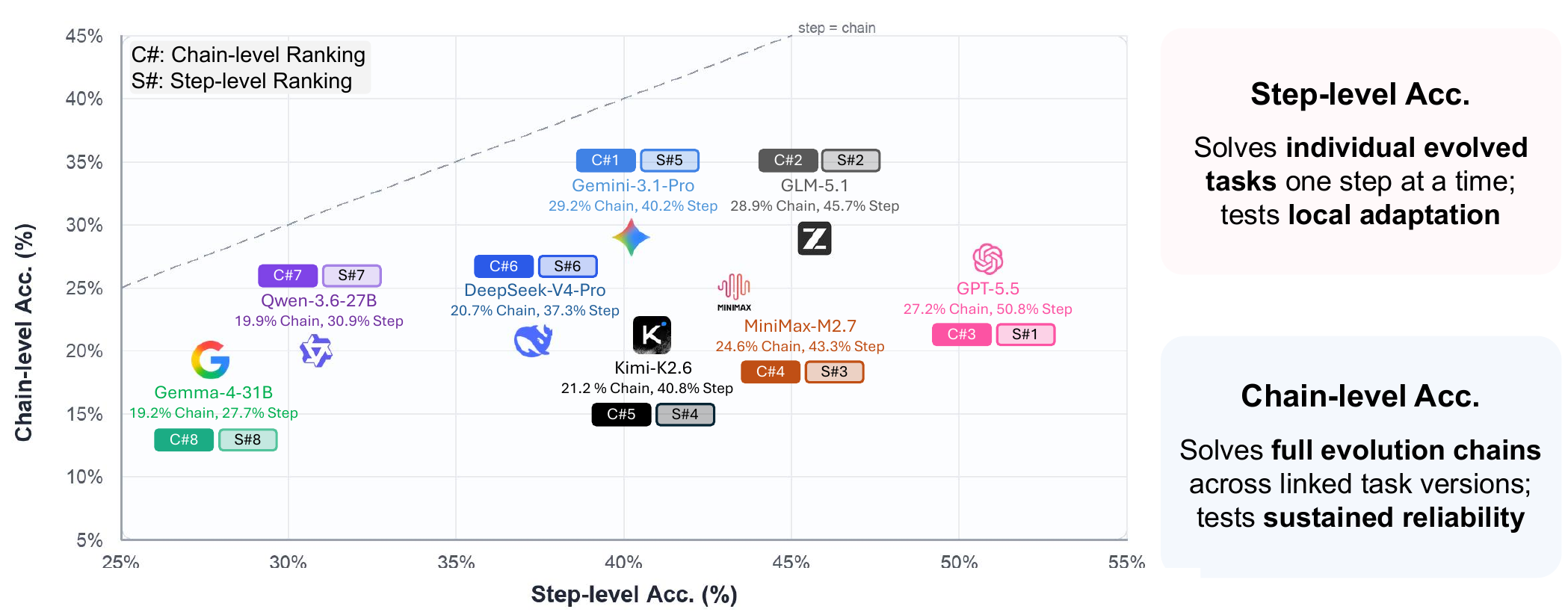}
% \vspace{-4mm}
\caption{\small \textbf{Step accuracy vs. chain accuracy on EvoArena.} The closer to the upper-right corner the better.}
\label{fig:evoarena_overall_performance}
% \vspace{2mm}
\end{figure}

\vspace{-5mm}

\begin{abstract}

Large language model (LLM) agents have achieved strong performance on a wide range of benchmarks, yet most evaluations assume static environments. In contrast, real-world deployment is inherently dynamic, requiring agents to continually align their knowledge, skills, and behavior with changing environments and updated task conditions. To address this gap, we introduce \textbf{EvoArena}, a benchmark suite that models environment changes as sequences of progressive updates across terminal, software, and social domains. 
We further propose \textbf{EvoMem}, a patch-based memory paradigm that records memory evolution as structured update histories, enabling agents to reason about environmental evolution through changes in their memory.
Experiments show that current agents struggle on EvoArena, achieving an average accuracy of 39.6\% across evolving terminal, software, and social-preference domains. EvoMem consistently improves performance, yielding an average gain of 1.5\% on EvoArena and also improving standard benchmarks such as GAIA and LoCoMo by 6.1\% and 4.8\%. 
Beyond individual tasks, EvoMem further improves chain-level accuracy by 3.7\% on EvoArena, where success requires completing a consecutive sequence of related evolutionary subtasks. 
Mechanistic analysis shows that EvoMem improves evidence capture in the memory, indicating better preservation of complete evolving environment states. 
Our results highlight the importance of modeling evolution in both evaluation and memory for reliable agent deployment.

\end{abstract}
\vspace{-3mm}

\newpage

\section{Introduction}
\vspace{-1mm}

LLM agents are increasingly expected to operate in external environments such as web systems, terminal workspaces, software repositories, and personalized assistants. Modern agent systems combine reasoning \cite{react}, tool use \cite{memento-s}, and memory to execute multi-step procedures \cite{xu2025amem}, act on environment state, and adjust their behavior within a task episode, achieving strong performance on benchmarks for web navigation \cite{zhou2023webarena}, software engineering \cite{jimenez2024swebench}, tool use \cite{mialon2024gaia}, and general agentic reasoning \cite{mialon2024gaia}. A harder setting arises when the same agent must remain useful across versions of the same environment, where interfaces, rules, code states, and user preferences continue to change.

Yet this progress leaves a central deployment challenge underexplored: \textbf{agents are usually evaluated on static environment snapshots.} In most benchmarks, the interface, rules, task distribution, and success criteria are fixed once the benchmark is constructed \cite{jimenez2024swebench,liu2024agentbench,mialon2024gaia,zhou2023webarena}. Recent dynamic evaluations improve freshness or interaction realism through refreshed tasks, asynchronous events, or self-evolving instances \cite{gaia2,li2026swe-benchl-ive,benchmark-selfevolv}, but they rarely test persistent environment evolution, where the same setting changes across versions. In deployment, APIs, workflows, codebases, and user preferences continually evolve. A reliable agent must know what changed, what still holds, and how to act under the current version.

To address this gap, we introduce \textbf{EvoArena}, a benchmark suite for evaluating agents under \textbf{persistent environment evolution}. EvoArena organizes each environment into a chain of \textbf{progressively evolving releases}, where the same underlying goal or setting is preserved while interfaces, rules, workflows, code states, or user preferences change over time. This formulation turns environment evolution into a measurable capability: an agent must solve the current task, identify which updates matter, and avoid reusing behaviors tied to obsolete versions. Concretely, EvoArena includes \textbf{Terminal-Bench-Evo} for evolving terminal workflows,  \textbf{SWE-Chain-Evo} for evolving codebases, and \textbf{PersonaMem-Evo} for evolving user preferences. Together, these benchmarks evaluate both \textit{forward adaptation} to new changes and \textit{version compatibility} with still-valid prior knowledge, testing whether agents remain reliable beyond static snapshots.

Using EvoArena, we find that even strong agent systems degrade substantially in evolving environments. One recurring failure mode is \textbf{state collapse}: most memory-based agents maintain memory as \textit{a single latest state}, such as a retrieved memory bank or episodic store \cite{ mem0,xu2025amem, memento-s}. This design is effective when newer information safely supersedes older information, but becomes brittle when different environment versions require different behaviors. A workflow permission update, for instance, may overwrite an earlier rule that still applies to an older release, a different organization, or a future rollback. The agent then loses both the previous behavior and the context explaining when it was valid. This reveals a need for \textbf{version-aware state tracking}: agents must retain the latest memory, recover relevant prior states, and reason over why each update occurred.

To address these limitations, we introduce \textbf{EvoMem}, a lightweight git-like memory paradigm for evolving environments. EvoMem augments a standard memory system with an append-only patch history that records meaningful memory changes. Each patch stores the pre-update memory, post-update memory, update rationale, and supporting evidence from the triggering context. This makes memory evolution \emph{traceable}: agents can inspect what changed, why it changed, and which evidence justified the update. At inference time, the agent retrieves from the latest memory by default, while selectively retrieving relevant patches when a query depends on overwritten states, conflicting evidence, or earlier environment versions. In this way, EvoMem turns memory updates into a compact evidence trail for reasoning over environment evolution.

Empirically, \textbf{EvoMem improves agent robustness across both EvoArena and standard long-horizon agent benchmarks}, with an average 1.5\% performance gain observed across agents and backbone models. 
In addition, EvoMem improves chain-level accuracy by 3.7\%, showing its ability to support agents in completing sequences of related tasks under a continuous environment evolving.
Beyond aggregate performance, our analysis shows why patch histories help: on PersonaMem-Evo, EvoMem yields stronger gains on temporal trajectory and multi-pattern synthesis questions, where agents must retain evolving and dispersed preference evidence. It also improves row-level evidence capture, indicating that patches better preserve complete preference states required for downstream reasoning. These findings suggest a broader implication: reliable agents in changing environments should treat memory as an evolving history of grounded updates, keeping the latest state connected to the prior states, rationales, and evidence that produced it. Overall, we make the following contributions::
\begin{itemize}
    \item We introduce \textbf{EvoArena}, a benchmark suite that evaluates agents across progressively evolving workflow, software, and social environments.
    \item We propose \textbf{EvoMem}, a patch-based memory paradigm that preserves traces of environmental evolution, enabling agents to reason more robustly in evolving environments.
    \item We show that existing agents struggle under environment evolution, while EvoMem consistently improves robustness and evidence retention across diverse settings.
\end{itemize}

\vspace{-3mm}

\section{Related Work}
\vspace{-1mm}

\textbf{Dynamic and Evolving Agent Benchmarks.} \quad
Agent benchmarks increasingly evaluate realistic interaction settings, including web navigation \cite{zhou2023webarena}, software engineering \cite{jimenez2024swebench}, tool use and general agentic reasoning \cite{liu2024agentbench,mialon2024gaia}, device control \cite{rawles2025androidworld}, terminal workflows \cite{terminal-bench}, enterprise tasks \cite{workarena++}, and personalized memory \cite{jiang2025personamem}. Recent dynamic benchmarks improve evaluation freshness or interaction realism through refreshed software tasks, asynchronous events, or generated benchmark variants \cite{gaia2,li2026swe-benchl-ive,benchmark-selfevolv}. Long-horizon personalization benchmarks further study how agents use evolving user context, but often introduce preference change as a single update rather than a multi-step version history \cite{li2026horizonbench}. EvoArena instead evaluates persistent environment evolution: the same setting changes across versions, requiring agents to adapt to new releases while preserving behavior that remains valid. We instantiate this setting by extending terminal, enterprise workflow, software-engineering, and personalization benchmarks into evolving chains.

\begin{table}[t]
\centering
\small
\setlength{\tabcolsep}{4pt}
\renewcommand{\arraystretch}{1.05}
\begin{tabular}{lccccc}
\toprule
\textbf{Benchmark} 
& \textbf{Domain} 
& \textbf{What Evolves?} 
& \textbf{PE} 
& \textbf{IC} 
& \textbf{CE} \\
\midrule

WebArena \cite{zhou2023webarena}
& Web
& Static tasks
& \xmark
& \xmark
& \xmark \\

SWE-bench \cite{jimenez2024swebench}
& Software
& Static issues
& \xmark
& \xmark
& \xmark \\

GAIA \cite{mialon2024gaia}
& Tool / Web
& Static tasks
& \xmark
& \xmark
& \xmark \\

AgentBench \cite{liu2024agentbench}
& Multi-env.
& Static tasks
& \xmark
& \xmark
& \xmark \\

AndroidWorld \cite{rawles2025androidworld}
& Device control
& Static tasks
& \xmark
& \xmark
& \xmark \\

Terminal-Bench \cite{terminal-bench}
& Terminal
& Static tasks
& \xmark
& \xmark
& \xmark \\

WorkArena++ \cite{workarena++}
& Workflow
& Static tasks
& \xmark
& \xmark
& \xmark \\

PersonaMem \cite{jiang2025personamem}
& Personalization
& User memory
& \pmark
& \pmark
& \xmark \\

SWE-bench-Live \cite{li2026swe-benchl-ive}
& Software
& Task refresh
& \xmark
& \xmark
& \xmark \\

GAIA2 \cite{gaia2}
& Tool / Web
& Async events
& \pmark
& \cmark
& \xmark \\

Benchmark Self-Evolving \cite{benchmark-selfevolv}
& General
& Generated tasks
& \xmark
& \xmark
& \xmark \\

HorizonBench \cite{li2026horizonbench}
& Personalization
& Preference changes
& \pmark
& \cmark
& \xmark \\

\midrule
\rowcolor{blue!8}
\textbf{EvoArena}
& \textbf{Multi-domain}
& \textbf{Dynamic Env}
& \cmark
& \cmark
& \cmark \\
\bottomrule
\end{tabular}
\vspace{1mm}
\caption{
\small \textbf{Comparison of EvoArena with related agent benchmarks.} PE denotes persistent environment evolution, where the same environment or user state evolves across versions; IC denotes implicit change, where the agent must infer or handle changes from context or interaction; CE denotes chain evaluation, where success is measured over temporally linked task sequences. \cmark denotes supported, \xmark denotes not supported, and \pmark denotes partial support.
}
\label{tab:evolving_benchmark_comparison}
\vspace{-5mm}
\end{table}

\textbf{Self-Evolving Agents and Memory.} \quad
Self-evolving agents improve their own behavior through reflection \cite{reflexion}, skill accumulation \cite{voyager}, reusable skill refinement \cite{memento-s}, or scaffold adaptation \cite{openhands}. These methods focus on agent-side improvement, while EvoArena studies changes in the external environment. Memory systems such as structured long-term memory \cite{xu2025amem}, production-scale persistent memory \cite{mem0}, and memory-based agent frameworks \cite{langgraph} continuously update stored knowledge. However, they usually consolidate memory toward the latest state, which can obscure overwritten states and the reasons behind updates. EvoMem treats memory updates as evidence, preserving what changed, why it changed, and which context supports each version-dependent behavior.

\begin{figure}[t]
    \centering
    \includegraphics[width=1\linewidth]{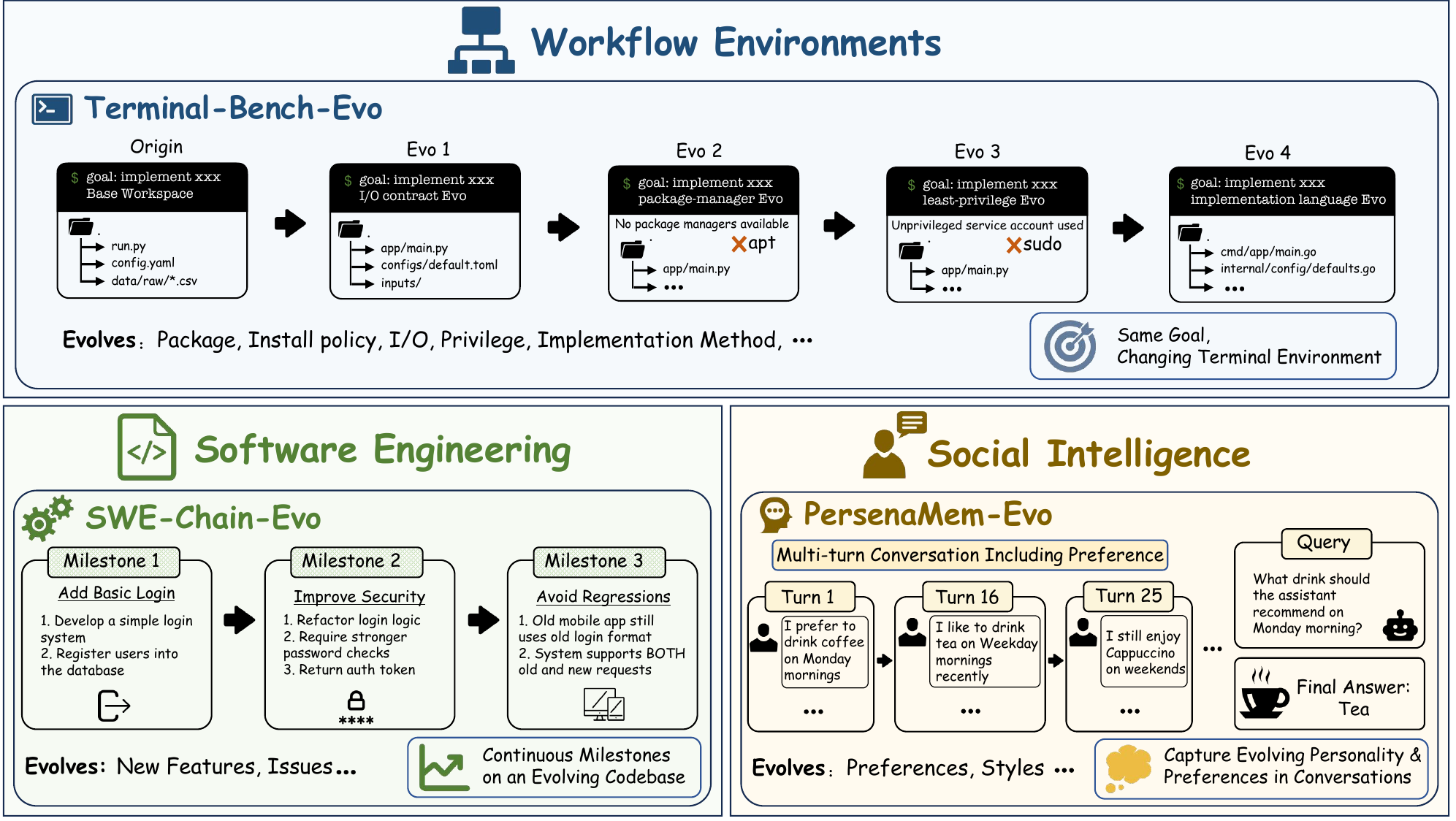}
    \vspace{-2mm}
\caption{\small
\textbf{EvoArena construction.}
We convert static agent benchmarks into versioned evolution chains across executable workflows, software engineering, and social intelligence, testing whether agents can adapt to new changes while preserving still-valid prior behavior.
}
    \vspace{-2mm}
    \label{fig:benchmark}
\end{figure}

\section{EvoArena: Benchmarking Agents under Persistent Environment Evolution}

Real-world agent deployment often occurs in \textbf{evolving environments}, where interfaces, rules, code states, user preferences, and system constraints change over time. Agents must infer these changes from observations, interaction feedback, and task outcomes, then decide which prior knowledge remains valid under the current version. Existing benchmarks largely evaluate agents on static snapshots, leaving this version-aware capability underexplored.

To study this setting, we construct \textbf{EvoArena}, a benchmark suite that evaluates agents under persistent environment evolution. As shown in Figure~\ref{fig:benchmark}, EvoArena covers three representative evolution regimes: \textbf{(1) executable workflow evolution} (Terminal-Bench-Evo), \textbf{(2) software evolution} (SWE-Chain-Evo), and \textbf{(3) preference evolution} (PersonaMem-Evo). These domains capture complementary forms of non-stationarity: terminal workflows change through dependency, interface, path, and validation updates; codebases evolve through continuous implementation milestones; and user preferences shift across long-horizon conversations. Together, they provide a unified testbed for evaluating whether agents can track environment changes, adapt to new versions, and preserve still-valid prior behavior.

\begin{figure}[t]
    \centering
    \includegraphics[width=0.9\linewidth]{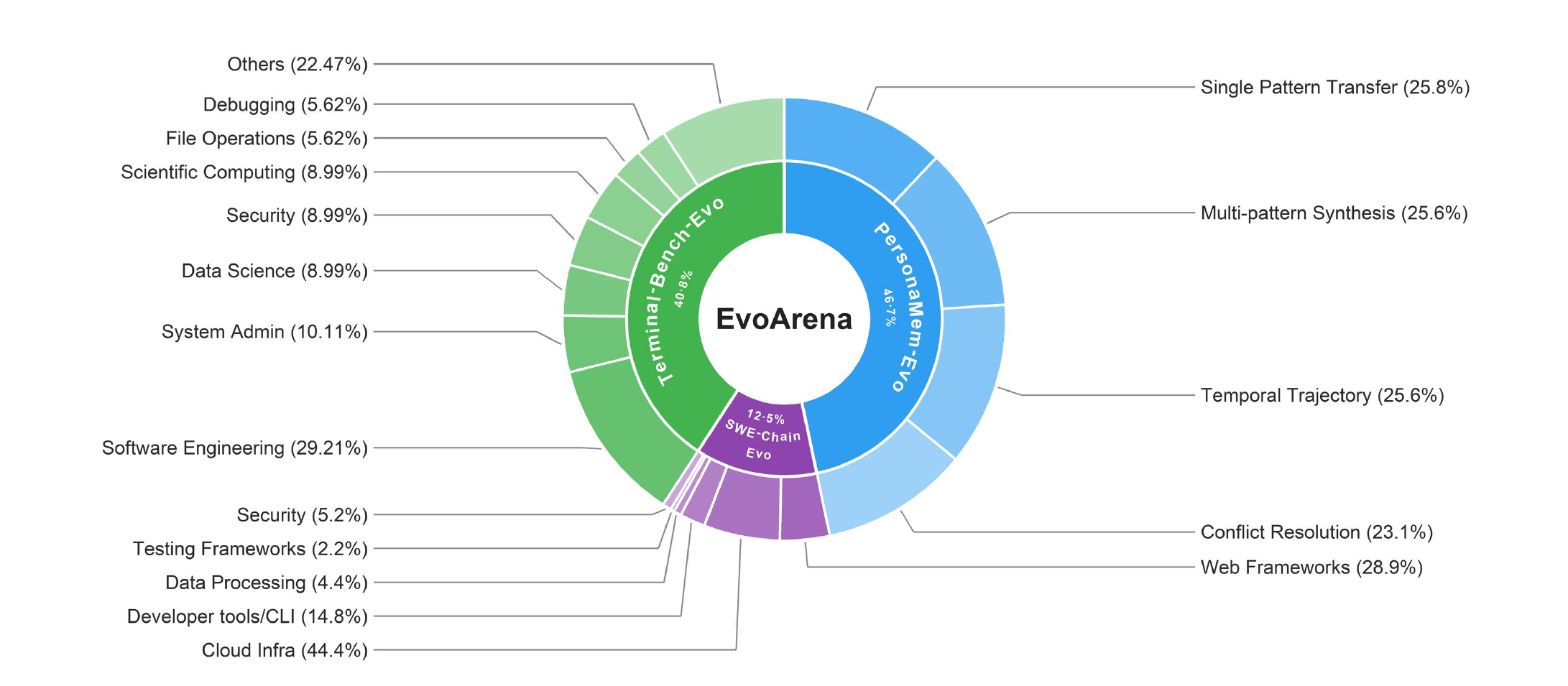}
    
    \caption{\small
    \textbf{Distribution of EvoArena}. The central circle shows domain proportions, and the surrounding panels show the question type distribution within each domain.
    }
    \vspace{-6mm}
    
    \label{fig:evoarena_stats}
\end{figure}

Figure~\ref{fig:evoarena_stats} summarizes the distribution of domain and question across EvoArena, and Table~\ref{tab:evoarena_subset_statistics} summarizes the key statistics for each subset. We describe each subset below and point each construction detail to the corresponding appendix subsubsection.

\subsection{Executable Workflow Evolution: Terminal-Bench-Evo}
\label{sec:terminal_bench_evo}

We choose executable terminal workflows because they expose the concrete surfaces through which deployed agents experience environment drift: files move, command-line interfaces change, dependencies are upgraded, validation scripts tighten, and policy constraints become part of the runnable task. We define an \emph{evolving workflow} as a discrete version chain: an ordered sequence of executable releases $\{v^{(1)}, \ldots, v^{(m)}\}$ that preserve the same Terminal-Bench objective \cite{terminal-bench} but change the surrounding instruction, environment, files, dependencies, interfaces, or tests between releases. Each version is a standalone terminal episode with a fixed container/workspace; the environment does not change during that episode. 
Our construction follows five stages that mirror the subsections below. First, \textbf{workflow-state analysis} extracts the objective, environment, files, dependencies, interfaces, and validation rules of each task. Second, \textbf{evolution-plan design} specifies realistic updates over mutable workflow components while preserving the underlying task objective. Then, \textbf{inherited version realization} materializes these updates as executable releases, where later versions inherit the realized environment of earlier ones. This is followed by \textbf{quality control and oracle validation}, which filters or repairs releases to ensure that each version is executable, internally consistent, and solvable by the reference solution. Finally, \textbf{benchmark assembly, metadata, and metrics} groups validated releases into workflow chains and records the metadata and evaluation units used for step- and chain-level scoring.
We describe each stage in detail below.

\textbf{(1) Workflow-state analysis.} \noindent
For each original task, we first convert the raw Terminal-Bench instance into a structured workflow state containing its objective, executable environment, relevant files, dependencies, interface/I/O contracts, and validation rules. This representation identifies what should remain stable across the chain, such as the user-facing objective, and what can plausibly evolve, such as paths, invocation commands, package versions, or test expectations. The resulting state becomes the input to evolution planning. Appendix~\ref{app:terminal_task_analysis} specifies the workflow-state fields and the mutable execution assumptions extracted from each task.

\textbf{(2) Evolution-plan design.} \noindent
Using the structured state, we design a sequence of candidate updates that preserve the objective while changing the operational procedure. The update taxonomy covers I/O and protocol changes, CLI/API changes, dependency or toolchain updates, workspace and module refactoring, and semantic or policy changes. For example, a later version may move the required output path, replace an invocation flag, upgrade a runtime, reorganize source modules, add a permission constraint, or tighten the validation rule. The output is an ordered plan specifying which workflow component changes at each version. Appendix~\ref{app:terminal_taxonomy} defines the evolution taxonomy and gives the component-level change categories used during planning.

\textbf{(3) Inherited version realization.} \noindent
We then instantiate the evolution plan in chronological order. Each version is produced by jointly editing the instruction, container environment, files/configuration, reference solution, tests, and metadata. Crucially, version $t$ starts from the realized environment of version $t-1$, so earlier path, dependency, permission, interface, and validation changes persist unless the current update explicitly revises them. This inheritance makes the chain a coherent workflow history rather than a set of independent variants, while still leaving each version as a complete executable task. Appendix~\ref{app:terminal_version_construction} describes how inherited environments, instructions, reference solutions, and tests are materialized for each release.

\textbf{(4) Quality control and oracle validation.} \noindent
After realization, we verify that each version is executable, internally consistent, and solvable. The reference solution must pass the version-specific tests, the instruction/environment/tests must agree with one another, and later versions must preserve inherited changes from earlier versions. Versions with ambiguous instructions, incomplete dependencies, inconsistent tests, or broken reference solutions are repaired or removed. This turns candidate releases into validated executable benchmark instances. Appendix~\ref{app:terminal_quality_control} details the consistency checks and filtering criteria, while Appendix~\ref{app:terminal_oracle_validation} reports the oracle-solution validation procedure.

\textbf{Benchmark assembly, evaluation, and metadata.} \noindent
Finally, we assemble the retained versions into workflow chains and store chain position, update type, changed components, environment summaries, validation summaries, and reference-solution information for analysis. The executable unit is one \emph{versioned task}: the agent receives a fixed version-specific instruction and container/workspace, and is scored by that version's tests. We report both \emph{step accuracy}, averaged over versioned tasks, and \emph{chain accuracy}, where a chain is correct only if all of its versions are solved. Appendix~\ref{app:terminal_metadata_stats} lists the stored metadata and dataset statistics, and Appendix~\ref{app:terminal_eval_unit_metrics} defines the step- and chain-level metrics.

\begin{figure}[t]
    \centering
    \includegraphics[width=1\linewidth]{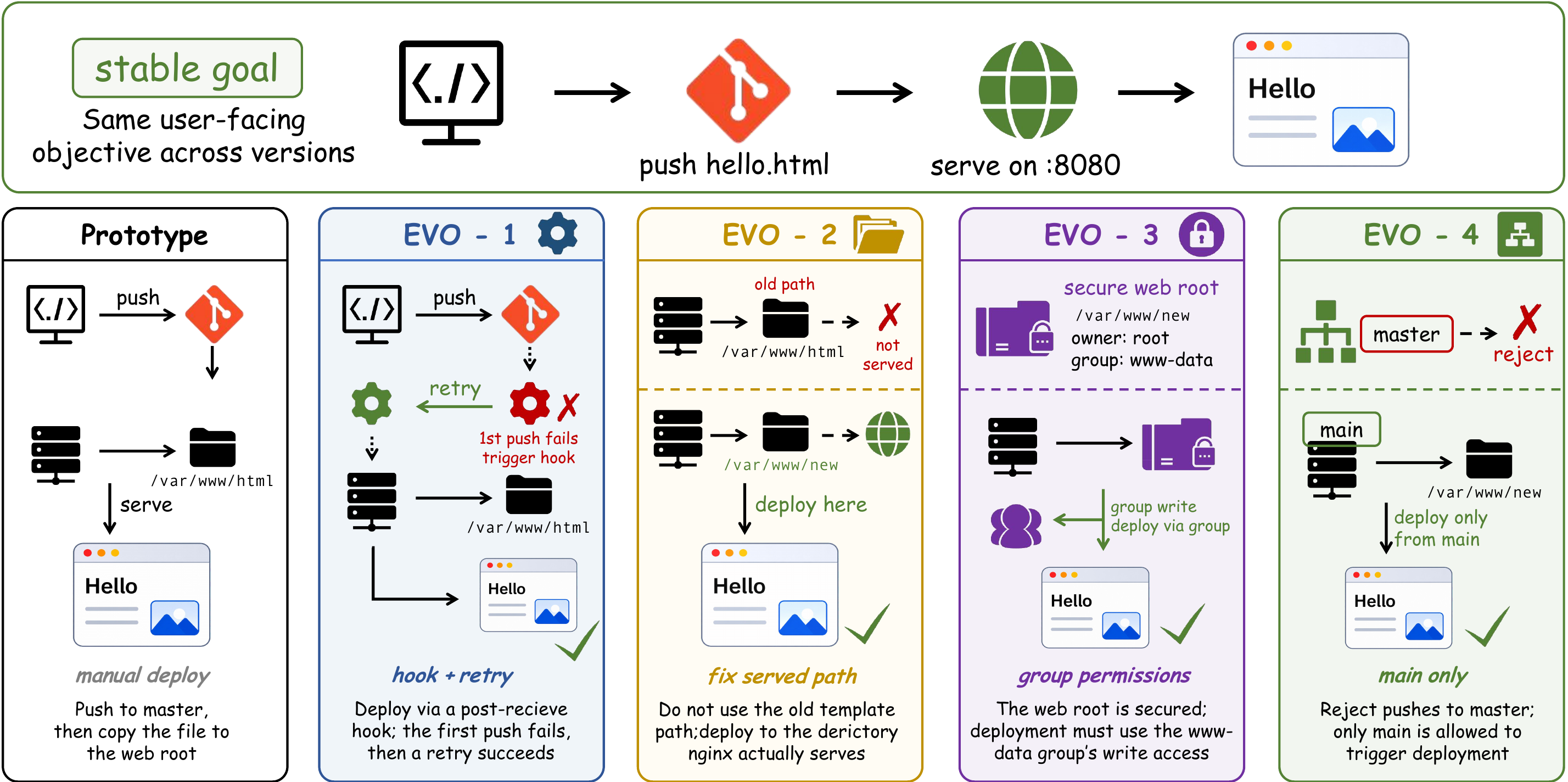}
    \caption{\small \textbf{An example of Terminal-Bench-Evo.}
The end goal stays fixed: push \texttt{hello.html} and serve it on port 8080, while each version changes a key operational constraint: deployment mechanism, served path, permissions, or branch policy.
}
    \vspace{-3mm}
    \label{fig:terminal-example}
\end{figure}

The final \textsc{Terminal-Bench-Evo} dataset contains 89 initial Terminal-Bench tasks, from which 356 evolved task versions are constructed as five-version workflow chains; after quality control removes 4 invalid versions, this results in 352 evolved versioned tasks and 441 total task instances including the initial versions. Chains contain four to five versions, with mean length 4.96 and median length 5. The evolved versioned tasks cover diverse operational changes, with I/O or protocol changes forming the largest group (49.1\%), followed by workspace/module/staging changes (13.4\%), CLI/API changes (10.5\%), dependency/toolchain/capability changes (8.0\%), and semantic, policy, or evaluation-rule changes (4.6\%). The benchmark also spans a broad range of terminal domains, with top categories including software engineering (29.5\%), system administration (10.0\%), security (9.1\%), data science (9.1\%), and scientific computing (8.6\%). In terms of difficulty, most instances are medium or hard, with 268 medium, 152 hard, 20 easy, and 1 expert task. These statistics characterize \textsc{Terminal-Bench-Evo} as executable workflow evolution. An example is shown in Figure~\ref{fig:terminal-example}.

\subsection{Software Evolution: SWE-Chain-Evo}
\label{sec:swe_chain_evo}

We choose software repositories because the codebase itself is an evolving environment: earlier API changes, dependency updates, tests, and implementation decisions remain active when later requirements arrive. We define software evolution as a chronological chain of \emph{milestone releases} $\{m^{(1)}, \ldots, m^{(T)}\}$ within a repository window. Each milestone is a localized, testable development objective implemented by one or more related commits. The history is embedded in the repository state: at step $t$, the agent sees the current accumulated repository snapshot and a milestone instruction, produces a patch, and is evaluated on tests for that milestone. 
Our construction follows six stages. First, \textbf{repository selection} identifies active, diverse, and testable repositories. Second, \textbf{update-window extraction} selects contiguous chronological commit ranges. Then, \textbf{milestone grouping} clusters related commits into coherent development objectives. Next, \textbf{task-description synthesis} converts each milestone into a SWE-bench-style requirement. This is followed by \textbf{Docker evaluation construction}, which builds executable tests and validates reference solutions. Finally, \textbf{chronological chain assembly} orders validated milestones into evolution chains through oracle repository-state progression. Each stage is described in detail below.

\textbf{(1) Repository selection.} \noindent
We start from candidate GitHub repositories and retain projects with recent development activity, non-trivial size, reproducible dependency installation, and automated tests that can be run in a Docker environment. We also require domain diversity across web frameworks, infrastructure, observability, security, developer tools, data processing, and testing frameworks. This defines the repository pool in which software evolution is both realistic and executable. Appendix~\ref{app:swe_repository_collection} gives the repository filtering criteria and domain coverage.

\begin{figure}[t]
    \centering

    \includegraphics[width=\linewidth]{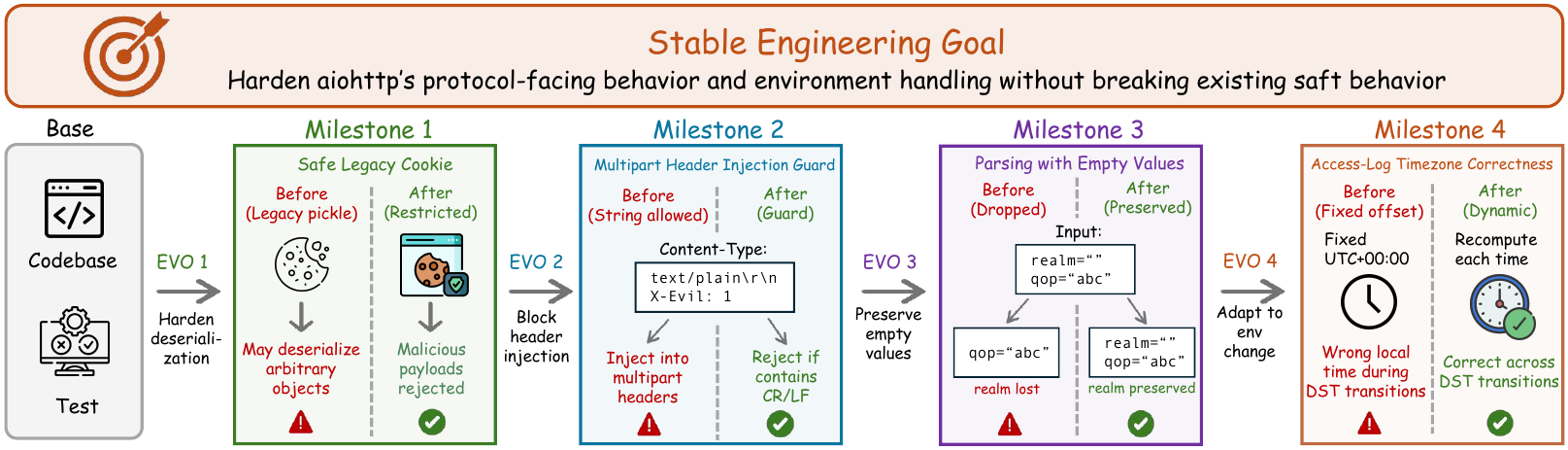}
    \caption{\small
    \textbf{Example evolution chain in SWE-Chain-Evo.}
    The \texttt{aiohttp} chain evolves across four milestones, progressively hardening protocol and environment boundaries while preserving existing compatible behavior.}
    \vspace{-3mm}
    \label{fig:swe_chain_evo_example}
\end{figure}

\textbf{(2) Update-window extraction.} \noindent
For each retained repository, an update window is a contiguous chronological range of commits from which candidate evolution chains can be formed. This keeps later milestones grounded in the same repository history rather than mixing unrelated snapshots, and it provides the local temporal context for grouping commits into milestones. Appendix~\ref{app:swe_milestone_construction} defines update windows and describes how commit ranges are selected before milestone grouping.

\textbf{(3) Milestone grouping.} \noindent
Within each update window, we group nearby commits into milestones using commit-message semantics and manual inspection of code diffs. A valid milestone must implement one coherent objective, such as a feature addition, bug fix, API adjustment, dependency migration, or test-backed maintenance update. This sits between two less useful extremes: release notes often aggregate many unrelated changes, while individual commits can be too small, incomplete, refactoring-only, or noisy. Formatting-only edits, incidental dependency bumps, merge artifacts, and unrelated changes are removed from the candidate milestone patch when they do not contribute to the objective, and the cleaned milestone is kept only if it remains executable and test-valid. Appendix~\ref{app:swe_milestone_construction} also describes commit grouping, noisy-change removal, and milestone coherence checks.

\textbf{(4) Task-description synthesis.} \noindent
Each validated milestone is converted into a SWE-bench-style instance. The task description is written from pre-milestone context and natural development signals such as commit messages, issue or PR text when available, tests, and high-level code-change summaries. We review descriptions to avoid leaking the reference patch, exact code edits, changed files, or solution-specific implementation details unless such information would naturally appear in a user requirement. The agent's expected output is a repository patch, which is applied to the pre-milestone chain state before evaluation. Appendix~\ref{app:swe_task_format} specifies the allowed description signals, leakage controls, and patch-output format.

\textbf{(5) Docker evaluation construction.} \noindent
We build a Docker environment for each instance, install dependencies, configure build and test commands, and validate that the reference milestone patch succeeds in that environment. Fail-to-Pass tests are tests that fail on the pre-milestone state and pass after applying the reference patch; every retained milestone must have at least one such test. Pass-to-Pass tests are regression tests that pass both before and after the reference patch, and are included when stable tests are available. A step is resolved only if the agent patch applies, all selected Fail-to-Pass tests pass, and no selected Pass-to-Pass regression test fails. Appendix~\ref{app:swe_docker_evaluation} defines the test-selection rules, Docker validation, and resolution criterion.

\textbf{Benchmark assembly, evaluation, and metadata.} \noindent
Validated milestones are ordered by their original chronological commit order within each repository window. After step $t$ is evaluated, we apply the \emph{reference} milestone update, not the agent's predicted patch, to form the repository state for step $t+1$. This oracle-state progression isolates adaptation to evolving repository states from compounding errors caused by earlier failed agent patches. 
Thus the benchmark tests whether agents can solve each localized requirement under the current accumulated codebase; memory-based agents may additionally use prior interaction context, but repository-state accumulation is fixed by the reference history.
When the same underlying milestone appears in overlapping chain contexts, it contributes multiple task-step slots but one unique milestone. 
We evaluate agents using two metrics: \emph{Step Accuracy}, which measures whether an agent solves each individual milestone at its chain position, and \emph{Chain Accuracy}, defined as the number of consecutively solved milestones from the beginning of a chain before the first unsolved or unexecuted milestone, divided by the total chain length.
Appendix~\ref{app:swe_chain_assembly} describes chronological ordering, oracle-state progression, and the distinction between task-step slots and unique milestones.

Overall, \textsc{SWE-Chain-Evo} contains 50 evolution chains from 12 repositories, with 493 chain-step instances and 145 unique repository milestones. A
\emph{chain-step instance} is a milestone evaluated at a particular position within an evolution chain, and serves as the unit on which agents are evaluated;
since some underlying repository updates appear in multiple chain contexts, deduplicating by milestone identity yields 145 unique milestones. Chains contain
5--15 steps, with mean length 9.86 and median length 10, including 10 five-step chains, 9 seven-step chains, 10 ten-step chains, 7 twelve-step chains, 7
thirteen-step chains, and 7 fifteen-step chains. Milestones are focused but non-trivial: each milestone is represented as one real commit transition, modify 2.72
files on average, with 38.6\% modifying multiple files, and have a median gold patch size of 53 diff lines. The benchmark spans Go (404 instances, 81.9\%) and
Python (89 instances, 18.1\%).

Each milestone instance includes at least one Fail-to-Pass test, with 7.13 Fail-to-Pass tests on average and a median of 2; Pass-to-Pass tests have mean 25.85
and median 4, with no instances lacking Pass-to-Pass tests. Besides, the chains exhibit substantial cross-step dependency: among the 443 non-initial milestone
instances, 29.8\% modify at least one file touched by an earlier milestone, and 14.2\% modify a file touched by the immediately preceding step. 
This reflects real
software evolution, where later milestones often depend on and revise earlier implementation decisions. Compared with independent SWE-bench-style instances,
\textsc{SWE-Chain-Evo} evaluates whether agents can solve localized software requirements under an evolving repository state where previous API, dependency,
test, and implementation changes remain active. Appendix~\ref{app:swe_dataset_statistics} reports detailed statistics, and an example is shown in
Figure~\ref{fig:swe_chain_evo_example}.

\subsection{Social Intelligence Evolution: PersonaMem-Evo}
\label{sec:personamem_evo}

We choose long-horizon preference interactions because personalization agents must infer user-specific behavior from implicit evidence, and real preferences often change as routines, constraints, experiences, or priorities shift. We define preference evolution as an ordered trajectory of latent preference states $\{p^{(0)}, \ldots, p^{(k)}\}$ for the same user and preference dimension. Earlier preference states remain in the interaction history as historically grounded evidence, while later episodes may revise their strength, scope, style, temporal validity, or applicable conditions. The evaluation unit is one multiple-choice question paired with a long mixed-topic history: the agent must select the answer best supported by the interaction history, not merely by persona stereotypes or commonsense priors. 
Our construction follows six stages. First, \textbf{seed persona expansion and preference cleaning} creates structured profiles and cleaned preference inventories. Second, \textbf{implicit interaction history construction} expresses preferences through realistic multi-turn interactions. Then, \textbf{temporal preference evolution} introduces ordered preference trajectories. Next, \textbf{preference-inference question generation} creates questions that require generalizing observed behavior to new settings. This is followed by \textbf{dual-blind filtering and answer-option construction}, which removes shortcuts and builds balanced options. Finally, \textbf{benchmark assembly, metadata, and evaluation} produces the final benchmark instances with metadata and task-accuracy evaluation. 
The following sections examine each stage in detail.

\textbf{(1) Seed persona expansion and preference cleaning.}
Built on PersonaMem-v2 \cite{jiang2025personamem}, we sample seed personas from PersonaHub \cite{ge2024scaling} and expand each seed into a structured profile with demographic, occupational, lifestyle, personality, and social-context fields. From this profile, we synthesize categorized preferences, including stereotypical, anti-stereotypical, neutral, therapy-related, and health/medical preferences. We then remove duplicated, empty, or inconsistent entries and assign topic labels. This defines the stable user background and the initial preference inventory that later interaction episodes will express or revise. Appendix~\ref{app:persona_seed_preferences} describes seed expansion, preference categories, cleaning rules, and topic labeling.

\begin{figure}[t]
    \centering

    \includegraphics[width=\linewidth]{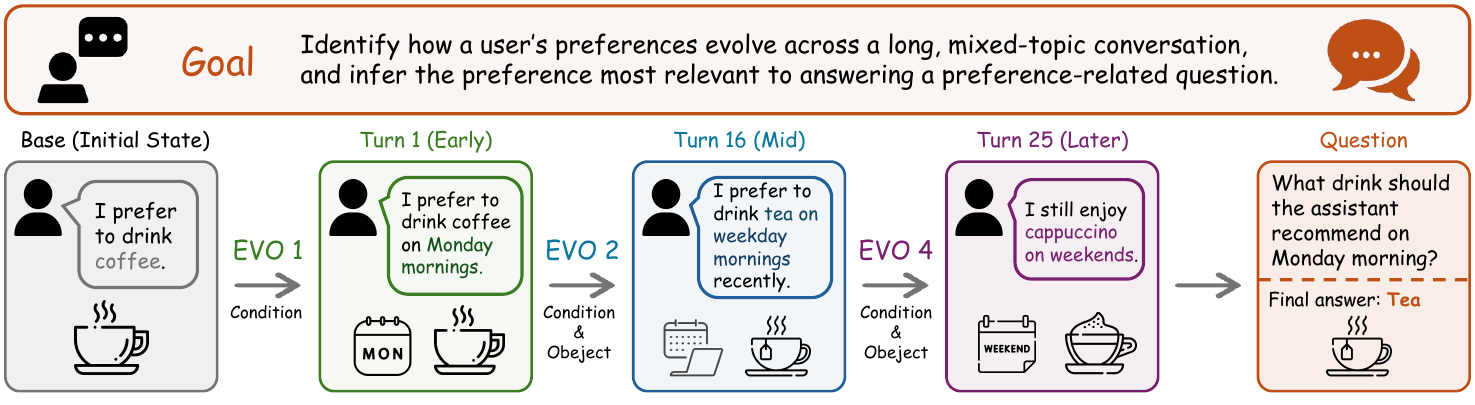}
    \caption{\small
    \textbf{Example evolution chain in PersonaMem-Evo.}
    The conversation history reveals evolving user preferences, and the query requires matching the relevant condition to the most appropriate current preference.
    }
    \label{fig:personamem_evo_example}
    \vspace{-6mm}
\end{figure}

\textbf{(2) Implicit interaction history construction.}
For each preference record, we generate one or more short conversation segments between a user with a AI-assistant in which the preference is implied through the user's request, constraints, wording, repeated behavior, or surrounding context. These segments are instantiated from a fixed set of interaction formats, such as asking the assistant to revise an email, translate a message, answer a knowledge question, give advice about a personal problem, or polish a social-media post. 
Each segment is associated with the preference record that generated it; the record specifies the preference text, its source category (e.g., stereotypical, anti-stereotypical, neutral, health-related, or therapy-related), whether it describes the user or another person, and whether it supersedes or deletes an earlier preference. 
We concatenate the generated segments into mixed-topic long-context histories while preserving segment boundaries, so the agent must recover user-specific evidence from realistic conversational context.
Appendix~\ref{app:persona_inter_format} provides the detailes of the interaction format and the preference record.

\textbf{(3) Temporal preference evolution.}
Each updated state is instantiated as a new implicit interaction episode and annotated with its previous state, updated state, change family, step index, and full trajectory. For example, a user may first prefer intense hiking, later avoid it after an injury, and then prefer light trail walks after recovery. This design requires agents to distinguish the latest preference from outdated but previously valid states. Appendix~\ref{app:persona_temporal_evolution} defines the change families, trajectory sampling, update annotations, and temporal-state construction.

\textbf{(4) Preference-inference question generation.}
From the long histories, we generate questions that require applying observed behavioral patterns to new decision settings. The question types are single-pattern transfer, multi-pattern synthesis, conflict resolution, and temporal trajectory prediction. These categories test whether agents can transfer an uncommon preference to a new domain, combine dispersed preference signals, arbitrate between competing preferences, or extrapolate a preference trajectory. Appendix~\ref{app:persona_ood_generation} describes the definition of each question type, generation prompts, and complexity levels.

\textbf{(5) Dual-blind filtering and answer-option construction.}
Each candidate question is packaged as a balanced four-way multiple-choice item. Distractors are designed to be plausible and may include stereotype-consistent choices, semantically related alternatives, or outdated preference states. We then apply dual-blind filtering: a question is rejected if it can be answered from the persona profile alone or without using the interaction history. This removes shortcuts and makes the retained questions depend on conversational memory instead of demographic priors or answer-option artifacts. Appendix~\ref{app:persona_dual_blind_filtering} defines the persona-only and no-context filters, and Appendix~\ref{app:persona_answer_options} describes distractor construction and answer-option balancing.

\textbf{Benchmark assembly, evaluation, and metadata.} \noindent
Finally, each accepted question is paired with its long-context history and metadata, including persona conversation, topic label, scenario type, reasoning type, complexity level, and, when applicable, preference-change family and trajectory step. To construct preference-evolution chains, we group preferences within the same persona conversation as belonging to the same evolving chain when their cosine similarity exceeds 70\% in chronological order. 
We evaluate agents using two metrics: \emph{Step Accuracy}, which measures accuracy on each individual multiple-choice question, and \emph{Chain Accuracy}, where a chain is counted as correct only if all questions in the corresponding preference-evolution chain are answered correctly. 
Appendix~\ref{app:persona_metadata} lists the benchmark fields, metadata schema, and evaluation format.

Overall, \textsc{PersonaMem-Evo} contains 10 persona-level conversations and 505 preference-inference questions, averaging 50.5 questions per persona. It has four balanced question types : single-pattern transfer (130), multi-pattern synthesis (129), temporal trajectory prediction (129), and conflict resolution (117).
Appendix~\ref{app:persona_dataset_statistics} reports the definition of each question type.
The dataset is built with different difficulty level, with 120 easy (L1), 186 medium (L2), and 199 hard questions (L3).

To make the benchmark useful for diagnosing memory-agent failures, we further annotate each question with its \emph{source preferences}, i.e., the gold observed preference evidence required to answer correctly. 
This measures how many memories must be retrieved and combined: single-pattern transfer uses one source preference, conflict resolution uses two, multi-pattern synthesis uses three, while temporal trajectory questions require 2-10 source preferences with median 6 and mean 6.2, making them a direct test of long-chain preference tracking. 
Besides, the benchmark requires reasoning over long interaction histories: each persona-level chat history contains a median of 597 messages, with a median length of 174.7K tokens, testing whether agents can track evolving preferences from sparse signals distributed across long conversations.
These statistics allow us to distinguish failures in memory retrieval, multi-evidence aggregation, stereotype bias, preference updating, and implicit preference extraction. Appendix~\ref{app:persona_dataset_statistics} reports detailed statistics, and an example is shown in Figure~\ref{fig:personamem_evo_example}.

\begin{table}[t]
\centering
\small
\setlength{\tabcolsep}{2pt}
\renewcommand{\arraystretch}{1.03}
\noindent\makebox[\linewidth][c]{%
\begin{minipage}[t]{0.318\linewidth}
\centering
\textbf{\textsc{Terminal-Bench-Evo}}\\[0.8mm]
\begin{tabular*}{\linewidth}{@{}l@{\extracolsep{\fill}}r@{}}
\toprule
\textbf{Statistics} & \textbf{Numbers} \\
\midrule
Init. tasks / chains & 89 \\
Constructed evol. & 356 \\
Invalid removed & 4 \\
Evolved tasks & 352 \\
Total inst. & 441 \\
Chain length & 4.96 / 5 \\
Len. range & 4--5 \\
\midrule
\multicolumn{2}{@{}l@{}}{\textit{Difficulty}} \\
Easy & 20 \\
Medium & 268 \\
Hard & 152 \\
Expert & 1 \\
\bottomrule
\end{tabular*}
\end{minipage}%
\hspace{0.014\linewidth}%
\begin{minipage}[t]{0.318\linewidth}
\centering
\textbf{\textsc{SWE-Chain-Evo}}\\[0.8mm]
\begin{tabular*}{\linewidth}{@{}l@{\extracolsep{\fill}}r@{}}
\toprule
\textbf{Statistics} & \textbf{Numbers} \\
\midrule
Repositories & 12 \\
Evol. chains & 50 \\
Unique miles. & 145 \\
Total inst. & 493 \\
Avg commits / mile. & 1.00 \\
\midrule
\multicolumn{2}{@{}l@{}}{\textit{Chain steps}} \\
Mean & 9.86 \\
5-step / 7-step & 10 / 9 \\
10-step / 12-step & 10 / 7 \\
13-step / 15-step & 7 / 7 \\
Avg P2P tests & 25.85 \\
Avg F2P tests & 7.13 \\
\bottomrule
\end{tabular*}
\end{minipage}%
\hspace{0.014\linewidth}%
\begin{minipage}[t]{0.318\linewidth}
\centering
\textbf{\textsc{PersonaMem-Evo}}\\[0.8mm]
\begin{tabular*}{\linewidth}{@{}l@{\extracolsep{\fill}}r@{}}
\toprule
\textbf{Statistics} & \textbf{Numbers} \\
\midrule
Persona conv. & 10 \\
Pref. chains & 313 \\
Total inst. & 505 \\
Avg conv. length & 174.7K \\
Avg turns & 597 \\
Q / persona & 50.5 \\
Q type counts & 130/129/129/117 \\
Temp. src. & 6 / 6.2 \\
\midrule
\multicolumn{2}{@{}l@{}}{\textit{Difficulty}} \\
Easy & 120 \\
Medium & 186 \\
Hard & 199 \\
\bottomrule
\end{tabular*}
\end{minipage}%
}
\vspace{1mm}
\caption{\small \textbf{Key statistics of the three EvoArena subsets.} Chain length reports mean / median. P2P and F2P denote Pass-to-Pass and Fail-to-Pass tests.}
\vspace{-5mm}
\label{tab:evoarena_subset_statistics}
\end{table}

\section{EvoMem: Patch-Based Memory Evolution}
\label{sec:method}

\subsection{Overview: Memory as an Evolution Trace}
\label{sec:method_overview}

Existing agent memory systems typically maintain a single latest memory state. This design is effective when newer observations safely supersede older ones. In evolving environments, however, knowledge is often \emph{version-dependent}: a rule updated for a new workflow release may overwrite an earlier rule that remains valid for an older release, another organization, or a future rollback. Updating memory toward the latest state can therefore erase useful prior behavior and the context explaining when that behavior applied.

\begin{figure}[t]
    \centering
    \includegraphics[width=0.92\linewidth]{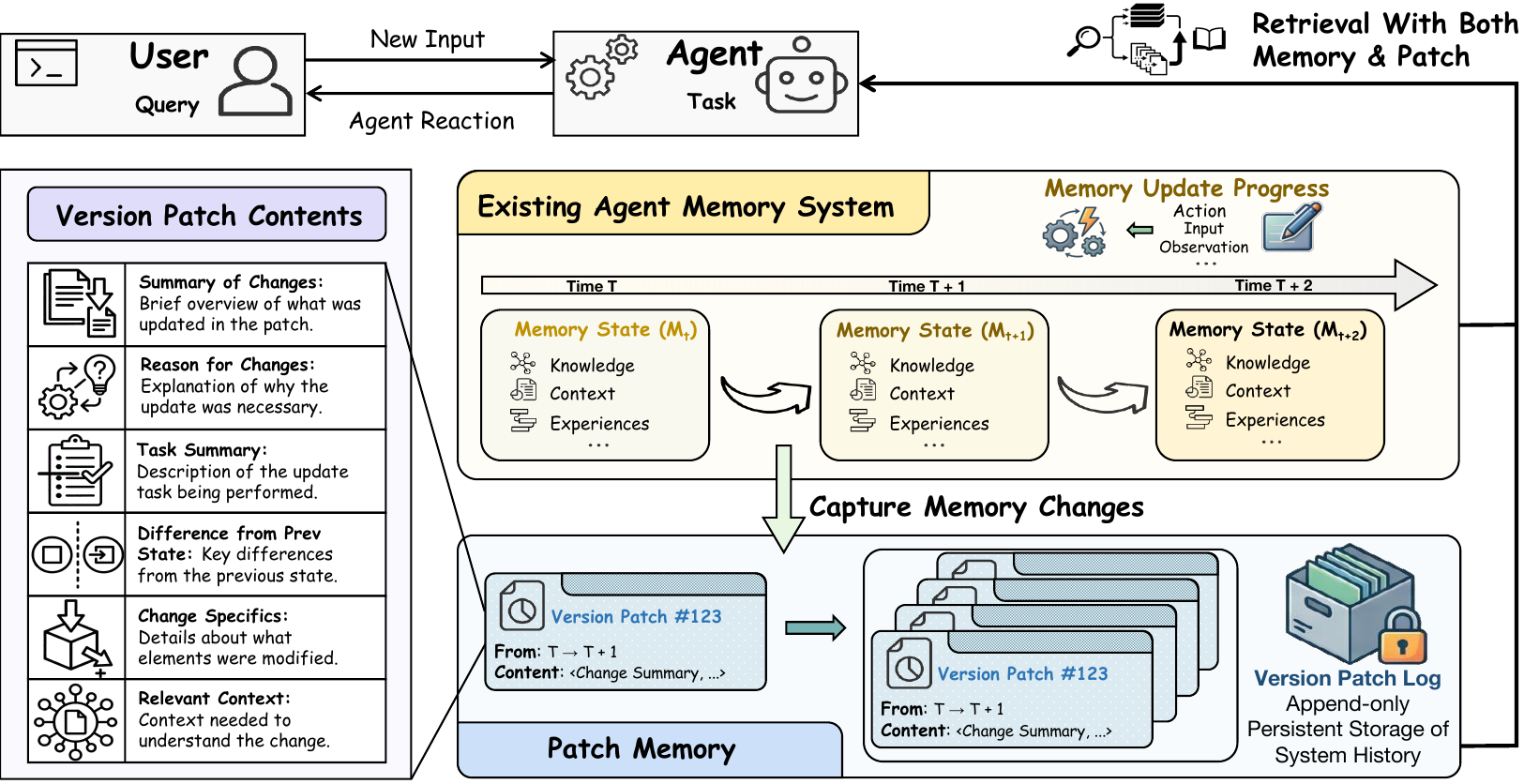}
    \vspace{-2mm}
    \caption{\small
    \textbf{Overview of EvoMem.}
    EvoMem augments a base memory system with an append-only patch history that records behaviorally meaningful memory updates and retrieves relevant patches as versioned evidence at inference time.
    }

    \label{fig:evomem_architecture}
\end{figure}

EvoMem addresses this limitation by augmenting an existing memory system with an explicit trace of memory evolution. As shown in Figure~\ref{fig:evomem_architecture}, EvoMem has two components. First, \emph{patch recording} stores meaningful non-additive memory updates, including what changed, why it changed, and which evidence triggered the update. Second, \emph{patch-augmented retrieval} retrieves relevant historical patches together with the latest memory when a query depends on overwritten states, temporal changes, or version-specific behavior. This design keeps the base memory system intact while making memory updates inspectable and reusable for downstream reasoning.

\subsection{Recording Memory Evolution as Patches}
\label{sec:patch_record}

The goal of patch recording is to preserve the \emph{behaviorally meaningful transitions} that occur when an agent updates its memory. In evolving environments, the latest memory content is often insufficient by itself: the agent may also need to know which earlier state was revised, what triggered the revision, and when the earlier state may still be valid. EvoMem therefore records non-additive memory updates as explicit patches, while leaving the base memory updater unchanged.

Let $x_t$ denote the input observation at time $t$, and let $M_{t-1}$ be the memory maintained by the base agent before observing $x_t$. The base memory updater produces a new memory state
\[
M_t = U(M_{t-1}, x_t),
\]
where $U$ can be any agent-specific update function, such as updating a text memory bank, revising a user profile, modifying a memory graph, or refining a skill file.

EvoMem does not replace this updater. It monitors the transition from $M_{t-1}$ to $M_t$ and captures memory changes that would otherwise be discarded. We compute
\[
\Delta_t = \mathrm{Diff}(M_{t-1}, M_t),
\]
where $\Delta_t$ summarizes the affected memory entries and changed fields. EvoMem creates a patch only for \emph{non-additive} updates, namely updates that revise, overwrite, or reinterpret existing memory. Purely additive observations can remain in the base memory without requiring a patch.

For each non-additive update, EvoMem writes a patch to an append-only patch history:
\[
p_t = \left(\tau_t, C_t^{-}, C_t^{+}, r_t, z_t, e_t\right).
\]
Here, $\tau_t$ denotes temporal metadata such as turn, session, or timestamp; $C_t^{-}$ and $C_t^{+}$ denote the affected memory content before and after the update; $r_t$ records the update rationale; $z_t$ provides a concise semantic summary of the change; and $e_t$ stores supporting evidence such as the triggering interaction, task context, execution feedback, or environment snapshot.

The resulting patch history is
\[
\mathcal{P}_{1:t}=\{p_1,\ldots,p_t\}.
\]
Thus, $M_t$ stores the latest consolidated memory, while $\mathcal{P}_{1:t}$ preserves the intermediate transitions that explain how the memory reached its current state. This separation turns memory from a single mutable store into a versioned evolution trace: the agent can act from the latest memory while retaining access to prior states, rationales, and evidence for version-aware reasoning.

\subsection{Retrieving Versioned Memory Evidence}
\label{sec:patch_retrieval}

Patch recording makes memory evolution available, but the agent still needs a way to use this history at inference time. The goal of patch-augmented retrieval is to expose version-relevant evidence only when it helps resolve the query. In ordinary cases, the latest memory may be sufficient. When the query depends on temporal changes, overwritten information, or update rationales, the agent should also access the patch history that explains how the current memory state was reached.

Given a query $q$, EvoMem first follows the standard memory-agent pipeline and retrieves evidence from the latest memory:
\[
c_{\mathrm{mem}} = R_{\mathrm{mem}}(q, M_T),
\]
where $M_T$ is the final memory after processing all observations, and $R_{\mathrm{mem}}$ is the base retriever.

EvoMem then retrieves relevant patches from the patch history:
\[
\mathcal{P}_q = R_{\mathrm{patch}}(q, \mathcal{P}_{1:T}),
\]
where $R_{\mathrm{patch}}$ returns the top-$k$ patches relevant to $q$. These patches provide versioned evidence about previous memory states, the updates that changed them, and the rationales behind those updates.

The final context is
\[
c(q)=\mathrm{Concat}\left(c_{\mathrm{mem}}, \mathcal{P}_q\right).
\]
This context is provided to the agent for response generation or action selection. The latest memory supplies the current consolidated state, while retrieved patches explain how that state changed and which prior states may still be relevant. As a result, the agent can ground its decision in both the current memory and the evolution trace behind it.

\subsection{Instantiation Across Agents}
\label{sec:instantiation}

EvoMem is a general memory abstraction with agent-specific schemas. Different agents represent and update knowledge in different forms, so each instantiation specifies three components: the base memory state $M_T$, the non-additive updates that should trigger patches, and the retrieval function over patch history. Across all cases, EvoMem follows the same principle: monitor memory updates, preserve meaningful overwritten states, and retrieve historical context when version-specific reasoning is required.

\begin{itemize}[leftmargin=*]
    \item \textbf{Terminus2 \cite{terminal-bench}.}
    Terminus2 is a terminal-native agent without a persistent memory module. We instantiate $M_T$ as distilled task-solving knowledge from prior terminal trajectories. Patches record strategy changes caused by evolving terminal environments, such as updated dependencies, interfaces, paths, or validation rules.  Appendix~\ref{sec:evomem_terminus2} details the chain-scoped terminal memory, transition-patch schema, retrieval-time memory construction, and anti-copying safeguards.

    \item \textbf{OpenHands \cite{openhands}.}
    We instantiate $M_T$ as distilled software-engineering context from previous trajectories, including files, symbols, constraints, and execution outcomes. Patches record superseded implementation strategies, revised logic, and the observed cause of the change, helping the agent reuse prior debugging evidence while avoiding regressions. Appendix~\ref{sec:evomem_openhands} details the feature-level patch records, semantic-structural retrieval score, prompt construction, and the non-invasive integration with the OpenHands loop.

    \item \textbf{A-Mem \cite{xu2025amem}.}
    A-Mem represents memory as semantically organized notes and links. We instantiate patches at the level of note and relation updates, preserving the previous note state, the revised content, and the rationale behind the change. This supports reasoning over evolving user preferences and temporally grounded memory conflicts. Appendix~\ref{sec:evomem_amem} details the graph-diff trigger, patch representation, patch indexing, query-time retrieval, and patch-conditioned answer generation.

    \item \textbf{Memento-Skill \cite{memento-s}.}
    Memento-Skill maintains a global \texttt{TIP.md} file as reusable skill memory. We treat \texttt{TIP.md} as $M_T$ and create patches whenever it is revised, storing the task-specific tip update, triggering failure, and update rationale. This allows the agent to retrieve task-specific corrections that may be diluted in the consolidated skill file. Appendix~\ref{sec:evomem_memento} details the versioned tip-memory family, store-time tip update, hybrid retrieval, prompt injection, and post-failure patching flow.
\end{itemize}

These instantiations show that EvoMem does not require a single memory format. It provides a lightweight interface for turning memory updates into reusable versioned evidence across terminal, software-engineering, conversational-memory, and skill-memory agents. Further implementation details are provided in Appendix~\ref{appendix:evomem_implement}.

\begin{table}[t]
\centering
\setlength{\tabcolsep}{3.0pt}
\renewcommand{\arraystretch}{0.95}
\caption{ \small
\textbf{Main results on EvoArena.}
We report step accuracy for all benchmarks and chain accuracy for benchmarks with executable evolution chains. Chain accuracy requires every step in a chain to be solved.
}
\resizebox{\linewidth}{!}{
\begin{tabular}{lllcccccc}
\toprule
\textbf{Benchmark} & \textbf{Agent} & \textbf{Model} & \multicolumn{3}{c}{\textbf{Step Acc.}} & \multicolumn{3}{c}{\textbf{Chain Acc.}} \\
\cmidrule(lr){4-6}\cmidrule(lr){7-9}
& & & \textbf{Base} & \textbf{+EvoMem} & \textbf{$\Delta$} & \textbf{Base} & \textbf{+EvoMem} & \textbf{$\Delta$} \\
\midrule

\multirow{10}{*}{Terminal-Bench-Evo}
& \multirow{10}{*}{Terminus 2}
& GPT-5.5 & 62.8 & \textbf{65.1} & \textcolor{green!50!black}{+2.3} & 31.8 & \textbf{45.5} & \textcolor{green!50!black}{+13.7} \\
& & Gemini-3.1-Pro & 53.8 & \textbf{56.5} & \textcolor{green!50!black}{+2.7} & 39.3 & \textbf{44.1} & \textcolor{green!50!black}{+4.8} \\
& & Kimi-K2.6 & 40.8 & \textbf{42.9} & \textcolor{green!50!black}{+2.1} & 14.9 & \textbf{22.7} & \textcolor{green!50!black}{+7.8} \\
& & Deepseek-V4-Pro & 37.3 & \textbf{40.4} & \textcolor{green!50!black}{+3.1} & 13.5 & \textbf{22.4} & \textcolor{green!50!black}{+8.9} \\
& & GLM-5.1 & 51.8 & \textbf{55.3} & \textcolor{green!50!black}{+3.5} & 34.2 & \textbf{36.8} & \textcolor{green!50!black}{+2.6} \\
& & MiniMax-M2.7 & 41.0 & \textbf{42.4} & \textcolor{green!50!black}{+1.4} & 18.2 & \textbf{19.5} & \textcolor{green!50!black}{+1.3} \\
& & Qwen3.6-27B & 37.6 & \textbf{40.9} & \textcolor{green!50!black}{+3.3} & 11.1 & \textbf{17.3} & \textcolor{green!50!black}{+6.2} \\
& & Gemma4-31B & 23.4 & \textbf{24.5} & \textcolor{green!50!black}{+1.1} & 9.0 & \textbf{12.4} & \textcolor{green!50!black}{+3.4} \\
\cmidrule(l){3-9}
& & \textbf{Average} & 43.6 & \textbf{46.0} & \textcolor{green!50!black}{\textbf{+2.4}} & 21.5 & \textbf{27.6} & \textcolor{green!50!black}{\textbf{+6.1}} \\
\cmidrule(l){1-9}

\multirow{10}{*}{SWE-Chain-Evo}
& \multirow{10}{*}{OpenHands}
& GPT-5.5 & 49.7 & \textbf{50.9} & \textcolor{green!50!black}{+1.2} & 12.2 & \textbf{16.8} & \textcolor{green!50!black}{+4.6} \\
& & Gemini-3.1-Pro & \textbf{20.5} & 18.1 & \textcolor{red!50!black}{-2.4} & 8.8 & \textbf{10.2} & \textcolor{green!50!black}{+1.4} \\
& & Kimi-K2.6 & \textbf{30.2} & 27.6 & \textcolor{red!50!black}{-2.6} & 8.5 & \textbf{12.1} & \textcolor{green!50!black}{+3.6} \\
& & Deepseek-V4-Pro & 26.7 & \textbf{27.7} & \textcolor{green!50!black}{+1.0} & 8.2 & \textbf{13.3} & \textcolor{green!50!black}{+5.1} \\
& & GLM-5.1 & 34.9 & \textbf{36.1} & \textcolor{green!50!black}{+1.2} & 9.9 & \textbf{12.8} & \textcolor{green!50!black}{+2.9} \\
& & MiniMax-M2.7 & 41.4 & \textbf{42.3} & \textcolor{green!50!black}{+0.9} & 14.7 & \textbf{15.3} & \textcolor{green!50!black}{+0.6} \\
& & Qwen3.6-27B & 11.6 & 11.6 & \textcolor{green!50!black}{+0.0} & 12.2 & \textbf{10.1} & \textcolor{green!50!black}{+2.1} \\
& & Gemma4-31B & 8.5 & \textbf{12.0} & \textcolor{green!50!black}{+3.5} & 5.2 & \textbf{6.3} & \textcolor{green!50!black}{+1.1} \\
\cmidrule(l){3-9}
& & \textbf{Average} & 27.9 & \textbf{28.3} & \textcolor{green!50!black}{\textbf{+0.4}} & 10.0 & \textbf{12.1} & \textcolor{green!50!black}{\textbf{+2.1}} \\
\cmidrule(l){1-9}

\multirow{8}{*}{PersonaMem-Evo}
  & \multirow{8}{*}{A-Mem}
  & GPT-5.5 & 40.0 & \textbf{43.8} & \textcolor{green!50!black}{+3.8} & 37.5 & \textbf{41.2} & \textcolor{green!50!black}{+3.7} \\
  & & Gemini-3.1-Pro & 46.4 & \textbf{48.3} & \textcolor{green!50!black}{+1.9} & 38.8 & \textbf{40.8} & \textcolor{green!50!black}{+2.0} \\
  & & Kimi-K2.6 & 51.5 & \textbf{55.5} & \textcolor{green!50!black}{+4.0} & 40.2 & \textbf{50.0} & \textcolor{green!50!black}{+9.8} \\
  & & Deepseek-V4-Pro & 47.9 & \textbf{51.6} & \textcolor{green!50!black}{+3.7} & 40.4 & \textbf{47.4} & \textcolor{green!50!black}{+7.0} \\
  & & GLM-5.1 & \textbf{50.4} & 47.5 & \textcolor{red!50!black}{-2.9} & \textbf{42.5} & 38.9 & \textcolor{red!50!black}{-3.7} \\
  & & MiniMax-M2.7 & 47.5 & \textbf{47.9} & \textcolor{green!50!black}{+0.4} & 40.9 & \textbf{41.4} & \textcolor{green!50!black}{+0.4} \\
  & & Qwen3.6-27B & 43.5 & \textbf{44.4} & \textcolor{green!50!black}{+1.0} & 36.5 & \textbf{39.9} & \textcolor{green!50!black}{+3.4} \\
    & & Gemma4-31B & 51.1 & \textbf{52.9} & \textcolor{green!50!black}{+1.8} & 43.5 & \textbf{45.8} & \textcolor{green!50!black}{+2.3} \\
  \cmidrule(l){3-9}
  & & \textbf{Average} & 47.3 & \textbf{49.0} & \textcolor{green!50!black}{\textbf{+1.7}} & 40.0 & \textbf{43.2} & \textcolor{green!50!black}{\textbf{+3.2}} \\

\bottomrule
\end{tabular}
}
\vspace{-3mm}
\label{tab:main_results}
\end{table}

\begin{table}[t]
\centering
\setlength{\tabcolsep}{3.5pt}
\renewcommand{\arraystretch}{0.95}
\caption{\small
\textbf{Main results on typical benchmarks.}
We report each benchmark's primary metric: task accuracy for agent benchmarks and exact match for long-horizon memory evaluation.
}
\begin{tabular}{lllccc}
\toprule
\textbf{Benchmark} & \textbf{Agent} & \textbf{Model} & \textbf{Base} & \textbf{+EvoMem} & \textbf{$\Delta$} \\
\midrule

\multirow{9}{*}{GAIA}
& \multirow{9}{*}{Memento-S}
& GPT-5.5 & 83.0 & 83.0 & +0.0 \\
& & Gemini-3.1-Pro & 57.0 & \textbf{65.0} & \textcolor{green!50!black}{+8.0} \\
% & & Kimi-K2.6 & -- & \textbf{--} & \textcolor{green!50!black}{--} \\
& & Gemma4-31B & 45.0 & \textbf{54.0} & \textcolor{green!50!black}{+9.0} \\
& & Deepseek-V4-Pro & 70.0 & \textbf{80.0} & \textcolor{green!50!black}{+10.0} \\
& & GLM-5.1 & 70.0 & \textbf{77.0} & \textcolor{green!50!black}{+7.0} \\
& & Qwen3.6-27B & 70.0 & \textbf{75.0} & \textcolor{green!50!black}{+5.0} \\
\cmidrule(l){3-6}
& & \textbf{Average} & 65.8 & \textbf{72.3} & \textcolor{green!50!black}{\textbf{+6.5}} \\
\cmidrule(l){1-6}

\multirow{9}{*}{LoCoMo}
& \multirow{9}{*}{A-Mem}
& GPT-5.5 & 32.9 & \textbf{33.9} & \textcolor{green!50!black}{+1.0} \\
& & Gemini-3.1-Pro & 21.1 & \textbf{28.6} & \textcolor{green!50!black}{+7.5} \\

& & Gemma4-31B & 52.3 & \textbf{55.2} & \textcolor{green!50!black}{+2.9} \\
& & Deepseek-V4-Pro & 52.0 & \textbf{56.5} & \textcolor{green!50!black}{+4.5} \\
& & Kimi-K2.6 & 54.0 & \textbf{57.7} & \textcolor{green!50!black}{+3.7} \\
& & Qwen3.6-27B & 26.0 & \textbf{26.3} & \textcolor{green!50!black}{+0.3} \\
\cmidrule(l){3-6}
& & \textbf{Average} & 39.7 & \textbf{43.0} & \textcolor{green!50!black}{\textbf{+3.3}} \\

\bottomrule
\end{tabular}
\vspace{-3mm}
\label{tab:typical_results}
\end{table}

\vspace{-3mm}

\section{Experiments}
\subsection{Experimental Setup}

We evaluate whether EvoMem improves agent robustness under environment evolution and long-horizon memory use. Our primary evaluation uses \textbf{EvoArena}, which contains three evolving benchmark subsets: \textsc{Terminal-Bench-Evo}, \textsc{SWE-Chain-Evo}, and \textsc{PersonaMem-Evo}. These subsets capture complementary forms of environment evolution: executable workflow changes, accumulated codebase changes, and evolving user preferences.

To test whether EvoMem remains useful beyond explicitly evolving benchmarks, we also evaluate on two standard agent benchmarks: \textsc{GAIA} \cite{mialon2024gaia} and \textsc{LoCoMo} \cite{locomo}. We report the main task-level metric for each benchmark: task accuracy for \textsc{EvoArena}, LLM-judge accuracy for \textsc{GAIA}, and exact match for \textsc{LoCoMo}. Together, these benchmarks cover terminal interaction, software engineering, tool use, and long-horizon conversational memory.

\textbf{Agents and models.} \quad
We pair each benchmark with an agent suited to its interaction setting. For \textsc{GAIA}, we use \textsc{Memento-Skill} \cite{memento-s}, which supports complex tool-use tasks. For \textsc{LoCoMo} and \textsc{PersonaMem-Evo}, we use \textsc{A-Mem} \cite{xu2025amem}, which maintains long-horizon conversational memory. For \textsc{Terminal-Bench-Evo} and \textsc{SWE-Chain-Evo}, we use coding and system-interaction agents, including \textsc{Terminus2} \cite{terminal-bench} and \textsc{OpenHands} \cite{openhands}. We evaluate across multiple backbone models, including \textsc{GPT-5.4-mini} \cite{openai2026gpt54mini}, \textsc{GPT-5.5} \cite{openai2026gpt55}, \textsc{Gemini-3.1-Pro} \cite{google2026gemini31pro}, \textsc{Qwen3.6-27B} \cite{qwen2026qwen36_27b}, and \textsc{Kimi-K2.6} \cite{moonshot2026kimi_k26}. Additional implementation details are provided in Appendix~\ref{appendix:exp_setting}.

\textbf{Evaluation.} \quad
We report \emph{Step Accuracy} for all benchmarks, where a step denotes the basic evaluation unit of each benchmark: a versioned terminal task in \emph{Terminal-Bench-Evo}, a milestone task in \emph{SWE-Chain-Evo}, a multiple-choice preference question in \emph{PersonaMem-Evo}, and a standard task/question instance in \emph{GAIA} and \emph{LoCoMo}. 
For EvoArena, we additionally report \emph{Chain Accuracy}, which measures whether the agent completes an entire evolution chain without any mistake. Specifically, a \emph{Terminal-Bench-Evo} chain is correct only if all versions derived from the same initial task are solved. A \emph{SWE-Chain-Evo} chain accuracy measures how many milestones are consecutively solved from the beginning of the repository evolution chain before the first unsolved or unexecuted milestone, divided by the total number of milestones in the chain. Lastly a \emph{PersonaMem-Evo} chain is correct only if all questions grouped into the same preference-evolution chain are answered correctly.

\subsection{Main Results}
\label{sec:main_results}

Table~\ref{tab:main_results} summarizes the main results and we highlight three findings.

\textbf{Current agents struggle under persistent environment evolution, especially at the chain level.}
Across \textsc{EvoArena}, base agents achieve only moderate step-level performance, with average accuracies of 43.6\% on \textsc{Terminal-Bench-Evo}, 29.2\% on \textsc{SWE-Chain-Evo}, and 46.5\% on \textsc{PersonaMem-Evo}. Performance drops further under the corresponding chain-level metrics for each subset: base agents achieve only 21.5\% on \textsc{Terminal-Bench-Evo}, 10.6\% on \textsc{SWE-Chain-Evo}, and 39.1\% on \textsc{PersonaMem-Evo}. These results show that strong agent systems still have difficulty tracking evolving execution conditions, accumulated codebase states, and shifting user preferences. More importantly, the substantially lower chain-level performance indicates that solving isolated steps does not necessarily translate into reliability across persistent environment evolution.

\textbf{EvoMem improves robustness across evolving and standard settings.} 
Adding EvoMem consistently improves performance across all EvoArena subsets. On average, EvoMem improves \textsc{Terminal-Bench-Evo} by 2.4\%, \textsc{SWE-Chain-Evo} by 0.5\%, and \textsc{PersonaMem-Evo} by 1.8\%. These gains indicate that preserving memory evolution helps across different forms of environmental change, from terminal interface and validation shifts to software milestones and preference updates. EvoMem also improves performance on standard agent benchmarks, with average gains of 6.1\% on \textsc{GAIA} and 4.8\% on \textsc{LoCoMo}. Across all evaluated settings, EvoMem yields an average improvement of 2.6\% at step-level and 3.7\% at chain-level. These results suggest a broader design implication: robust agent memory should preserve update history as retrievable evidence, especially when task conditions change over time.

\textbf{EvoMem improves even more at the chain level than at the step level.}
EvoMem yields larger gains under chain-level evaluation than under step-level evaluation, highlighting its value for maintaining consistency across dependent task sequences. On \textsc{Terminal-Bench-Evo}, the average improvement increases from 2.4\% at the step level to 6.1\% at the chain level. On \textsc{SWE-Chain-Evo}, the gain increases from 0.5\% step accuracy to 2.9\% chain accuracy, and on \textsc{PersonaMem-Evo}, from 1.8\% step accuracy to 3.0\% chain accuracy. This pattern suggests that EvoMem is especially effective when success requires agents to preserve and reuse evolution-aware knowledge over multiple related environment states.

\section{Analysis: When and Why Does EvoMem Help?}
\label{sec:diagnostic_analysis}

Beyond aggregate performance, we analyze when and why EvoMem helps agents in evolving environments. We examine its behavior across domains as follows.

\subsection{Terminal-Bench-Evo Mechanism Analysis}
\label{sec:terminal_pm_mechanism}

We perform an observational mechanism analysis to understand how EvoMem changes \textsc{Terminus2}'s behavior on \textsc{Terminal-Bench-Evo}. The central hypothesis is that patch memory helps only when the agent operationalizes the
  retrieved transition information: it must notice what changed from previous variants, preserve the useful part of the old procedure, and modify the part that is no longer valid. We therefore analyze not only
  whether patch memory is available, but whether its contents appear in the agent's later reasoning or executed commands.
  
We design four factors for this purpose. \emph{Patch example retrieval} tests whether EvoMem supplies at least one explicit EvoMem transition example, which describes a prior requirement, the evolved requirement, and the observed adaptation between them. \emph{Evolved-requirement coverage} tests whether changed requirements summarized in EvoMem patch reappear in later agent's reasoning or commands, measuring whether the current EVO constraints become salient. \emph{Patch uptake} tests whether terms from retrieved transition EvoMem patches reappear in subsequent reasoning or commands, measuring whether the historical adaptation pattern is incorporated into the agent's plan. \emph{Command-level patch uptake} is stricter: it only counts overlap with executed shell commands, and therefore measures whether patch information affects concrete terminal actions. 
For each factor, we compare baseline and EvoMem accuracy on the same subset of instances. 

\begin{table}[t]
\centering
\small
\caption{ \small \textbf{EvoMem gains stratified by EvoMem-side retrieval and behavioral uptake conditions.} Baseline is evaluated on the same instance subsets but receives no patch memory. Gain gap denotes the difference between the EvoMem improvement under the stronger condition and the weaker condition.}
\label{tab:terminal_pm_gain_gap}
\resizebox{\linewidth}{!}{
\begin{tabular}{llcccc}
\toprule
\textbf{Mechanism factor} & \textbf{Condition} & \textbf{Baseline} & \textbf{EvoMem} & \textbf{$\Delta$} & \textbf{Gain gap} \\
\midrule
\multirow{2}{*}{Patch example retrieval}
& EvoMem retrieved no patch example & 46.6\% & 49.7\% & +3.1\% & \multirow{2}{*}{\textcolor{green!50!black}{+3.4\%}} \\
& EvoMem retrieved patch example(s) & 87.1\% & 93.5\% & +6.5\% &  \\
\midrule
\multirow{2}{*}{Evolved-requirement coverage}
& Low coverage & 41.2\% & 43.3\% & +2.1\% & \multirow{2}{*}{\textcolor{green!50!black}{+3.2\%}} \\
& High coverage & 65.3\% & 70.5\% & +5.3\% &  \\
\midrule
\multirow{2}{*}{Patch uptake}
& No patch uptake & 46.8\% & 49.4\% & +2.6\% & \multirow{2}{*}{\textcolor{green!50!black}{+5.7\%}} \\
& Patch uptake $>0$ & 80.6\% & 88.9\% & +8.3\% &  \\
\midrule
\multirow{2}{*}{Command-level patch uptake}
& No command-level uptake & 46.2\% & 49.4\% & +3.1\% & \multirow{2}{*}{\textcolor{green!50!black}{+3.1\%}} \\
& Command-level uptake $>0$ & 87.5\% & 93.8\% & +6.2\% &  \\
\bottomrule
\end{tabular}
}
\vspace{-2mm}
\label{tab:terminal-analysis}
\end{table}

\paragraph{EvoMem helps when retrieved transitions are operationalized.}
The results support the operationalization hypothesis, as shown in Table~\ref{tab:terminal-analysis}. When EvoMem retrieves no explicit patch example, the gain over baseline is 3.1\%; when a patch example is retrieved, the gain rises to 6.5\%. High evolved-requirement coverage also yields a larger gain than low coverage (5.3\% vs. 2.1\%), indicating that attending to the current variant's changed requirement is important for avoiding stale reuse. The largest gap appears for patch uptake: when the agent does not reuse any patch-transition terms, the gain is 2.6\%, but when patch uptake is nonzero, the gain increases to 8.3\%. Qualitatively, this pattern suggests that EvoMem is not simply adding more context. \textsc{Terminal-Bench-Evo} tasks often preserve much of the earlier procedure while changing a decisive requirement, such as an interface detail, file location, artifact convention, or environment constraint. EvoMem helps when the retrieved patch identifies this local transition and the agent carries it into reasoning or commands: the reusable part of the old strategy is preserved, while the part invalidated by the current evolution is revised.

\subsection{SWE-Chain-Evo Mechanism Analysis}

To understand why EvoMem improves \textsc{SWE-Chain-Evo}, we analyze whether it helps agents preserve behavior introduced by earlier milestones while solving the current code-editing task. This targets a central challenge in software evolution: implementing new requirements without regressing previously correct behavior. We therefore focus on \emph{\texttt{PASS\_TO\_PASS} failures}, where at least one preservation test that should continue passing after the current milestone fails after the agent patch. This metric directly captures whether the agent breaks historical behavior.

\begin{table}[h]
\centering
\small
\setlength{\tabcolsep}{8pt}
\renewcommand{\arraystretch}{1.18}
\caption{\small \textbf{Regression analysis on \textsc{SWE-Chain-Evo}}. Lower is better.}
\label{tab:swe_chain_regression}
\begin{tabular*}{0.78\linewidth}{@{\extracolsep{\fill}}lccc}
\toprule
\textbf{Model} & \textbf{Base} & \textbf{+EvoMem} & \textbf{$\Delta$} \\
\midrule
Qwen3.6-27B & 9.01\% & 6.73\% & \textcolor{green!50!black}{$-2.28\%$} \\
Kimi-K2.6 & 7.14\% & 3.33\% & \textcolor{green!50!black}{$-3.81\%$} \\
Gemini-3.1-Pro & 11.11\% & 8.89\% & \textcolor{green!50!black}{$-2.22\%$} \\
\midrule
\textbf{Average} & \textbf{9.09\%} & \textbf{6.32\%} & \textcolor{green!50!black}{\textbf{$-2.77\%$}} \\
\bottomrule
\end{tabular*}
\vspace{-1mm}
\end{table}

\begin{table}[htbp]
\centering
\caption{
\textbf{\small \textsc{PersonaMem-Evo} accuracy breakdown (\%).}
Results are grouped by reasoning type and difficulty level.
}
\label{tab:personamem_breakdown}

\begin{minipage}[t]{0.53\linewidth}
\centering
\resizebox{\linewidth}{!}{
\begin{tabular}{lccc}
\toprule
\textbf{Question Type} & \textbf{Base} & \textbf{+EvoMem} & \textbf{$\Delta$} \\
\midrule
Conflict Resolution & \textbf{29.5} & 28.6 & \textcolor{red!50!black}{$-0.9$} \\
Single-Pattern Transfer & \textbf{46.2} & 44.4 & \textcolor{red!50!black}{$-1.8$} \\
Multi-Pattern Synthesis & 38.8 & \textbf{44.0} & \textcolor{green!50!black}{$+5.2$} \\
Temporal Trajectory & 46.6 & \textbf{51.7} & \textcolor{green!50!black}{$+5.2$} \\
\midrule
\rowcolor{blue!8}
\textbf{Overall} & 40.5 & \textbf{42.5} & \textcolor{green!50!black}{$+2.0$} \\
\bottomrule
\end{tabular}
}
\textbf{(a) By question type.}
\end{minipage}
\hspace{0.035\linewidth}
\begin{minipage}[t]{0.41\linewidth}
\centering
\resizebox{\linewidth}{!}{
\begin{tabular}{lccc}
\toprule
\textbf{Difficulty} & \textbf{Base} & \textbf{+EvoMem} & \textbf{$\Delta$} \\
\midrule
L1 & 38.9 & \textbf{40.7} & \textcolor{green!50!black}{$+1.8$} \\
L2 & 34.7 & \textbf{38.3} & \textcolor{green!50!black}{$+3.6$} \\
L3 & 46.9 & \textbf{47.5} & \textcolor{green!50!black}{$+0.6$} \\
\midrule
\rowcolor{blue!8}
\textbf{Overall} & 40.5 & \textbf{42.5} & \textcolor{green!50!black}{$+2.0$} \\
\bottomrule
\end{tabular}
}
\textbf{(b) By difficulty level.}
\end{minipage}

\end{table}

\paragraph{EvoMem reduces regressions across backbones.}
Table~\ref{tab:swe_chain_regression} shows that EvoMem consistently reduces \texttt{PASS\_TO\_PASS} failure rates across all evaluated backbones. The average regression rate drops from 9.09\% to 6.32\%, with the largest reduction on Kimi-K2.6 (3.81\%). This indicates that EvoMem helps agents preserve historical code constraints while adapting to new milestonesg. In SWE-Chain-Evo, where later edits build on earlier repository states, this regression reduction indicates that EvoMem helps agents maintain previously established behavior during software evolution.

\subsection{PersonaMem-Evo Mechanism Analysis}

To understand when and why EvoMem improves agent performance, we analyze whether it preserves the evolving evidence needed for downstream reasoning. PersonaMem-Evo provides fine-grained annotations of preference states and evolution trajectories, and its base agent, \textsc{A-Mem}, is explicitly memory-based. 
This setting allows us to study how EvoMem changes memory construction while minimizing confounds from tool-use or coding-specific behaviors.

\textbf{EvoMem helps most when reasoning requires temporal or dispersed evidence.} \quad
Table~\ref{tab:personamem_breakdown} breaks down performance by question type and difficulty. EvoMem improves overall accuracy from 40.5\% to 42.5\%, with the largest gains on \emph{temporal trajectory} and \emph{multi-pattern synthesis} questions, both improving by 5.2\%. This pattern matches the intended role of patch memory. Temporal trajectory questions require tracking how a preference changes over time, while multi-pattern synthesis requires retaining multiple preference signals scattered across the interaction history. In both cases, a single consolidated memory state may lose intermediate evidence, while EvoMem can recover relevant historical states from patch histories. The gains are less uniform on \emph{conflict resolution} and \emph{single-pattern transfer}, where performance decreases slightly. These categories require more than recovering missing evidence: conflict resolution requires inferring priority among competing preferences, and single-pattern transfer requires applying an uncommon or counter-stereotypical preference to a new setting. This suggests that patch histories improve evidence availability, while the final reasoning step remains a bottleneck for harder preference-inference cases.

\begin{table}[t]
  \centering
\caption{
\textbf{\small Preference evidence capture on \textsc{PersonaMem-Evo}.}
Clause-level capture measures preservation of atomic preference evidence, while row-level capture evaluates whether the complete evidence set for each example is retained.
}
  \setlength{\tabcolsep}{5pt}
  \begin{tabular}{lcccccc}
  \toprule
  \multirow{2}{*}{\textbf{Question Type}}
  & \multicolumn{3}{c}{\textbf{Clause-level Capture}}
  & \multicolumn{3}{c}{\textbf{Row-level Capture}} \\
  \cmidrule(lr){2-4}\cmidrule(lr){5-7}
  & \textbf{Base} & \textbf{+EvoMem} & $\boldsymbol{\Delta}$
  & \textbf{Base} & \textbf{+EvoMem} & $\boldsymbol{\Delta}$ \\
  \midrule
  Single-Pattern Transfer & \textbf{74.4} & \textbf{74.4} & +0.0 & \textbf{74.4} & \textbf{74.4} & +0.0 \\
  Multi-Pattern Synthesis & 79.9 & \textbf{81.6} & \textcolor{green!50!black}{+1.7} & 56.0 & \textbf{59.5} & \textcolor{green!50!black}{+3.5} \\
  Conflict Resolution & 82.9 & \textbf{83.8} & \textcolor{green!50!black}{+0.9} & 66.7 & \textbf{68.6} & \textcolor{green!50!black}{+1.9} \\
  Temporal Trajectory & 98.5 & \textbf{99.2} & \textcolor{green!50!black}{+0.7} & 92.2 & \textbf{96.6} & \textcolor{green!50!black}{+4.4} \\
  \midrule
\rowcolor{blue!8}
  Overall & 89.4 & \textbf{90.3} & \textcolor{green!50!black}{+0.9} & 72.5 & \textbf{74.9} & \textcolor{green!50!black}{+2.4} \\
  \bottomrule
  \end{tabular}
  \vspace{-2mm}
  \label{tab:persona_capture_by_ood}
\end{table}

\textbf{Patch histories improve complete evidence preservation.} \quad
To test the mechanism more directly, we measure whether the constructed memory store preserves ground-truth preference evidence. We report \emph{clause-level capture}, which checks whether each atomic preference clause appears in at least one memory unit, and \emph{row-level capture}, which requires all clauses needed for an example to be preserved. Table~\ref{tab:persona_capture_by_ood} shows that EvoMem improves clause-level capture from 89.4\% to 90.3\% and row-level capture from 72.5\% to 74.9\%. The larger row-level gain indicates that EvoMem better preserves complete preference states needed for downstream reasoning. Category-level results further support this explanation: the largest row-level capture gains occur on \emph{temporal trajectory} (4.4\%) and \emph{multi-pattern synthesis} (3.5\%), which are also the categories with the strongest accuracy gains. These results suggest that EvoMem improves performance by preserving coherent versioned evidence, especially when the answer depends on how preferences evolve across interactions.

\vspace{-3mm}

\subsection{Efficiency--Accuracy Trade-off}
\label{sec:efficiency_accuracy}

\begin{figure}[t]
\centering
\includegraphics[width=0.9\linewidth]{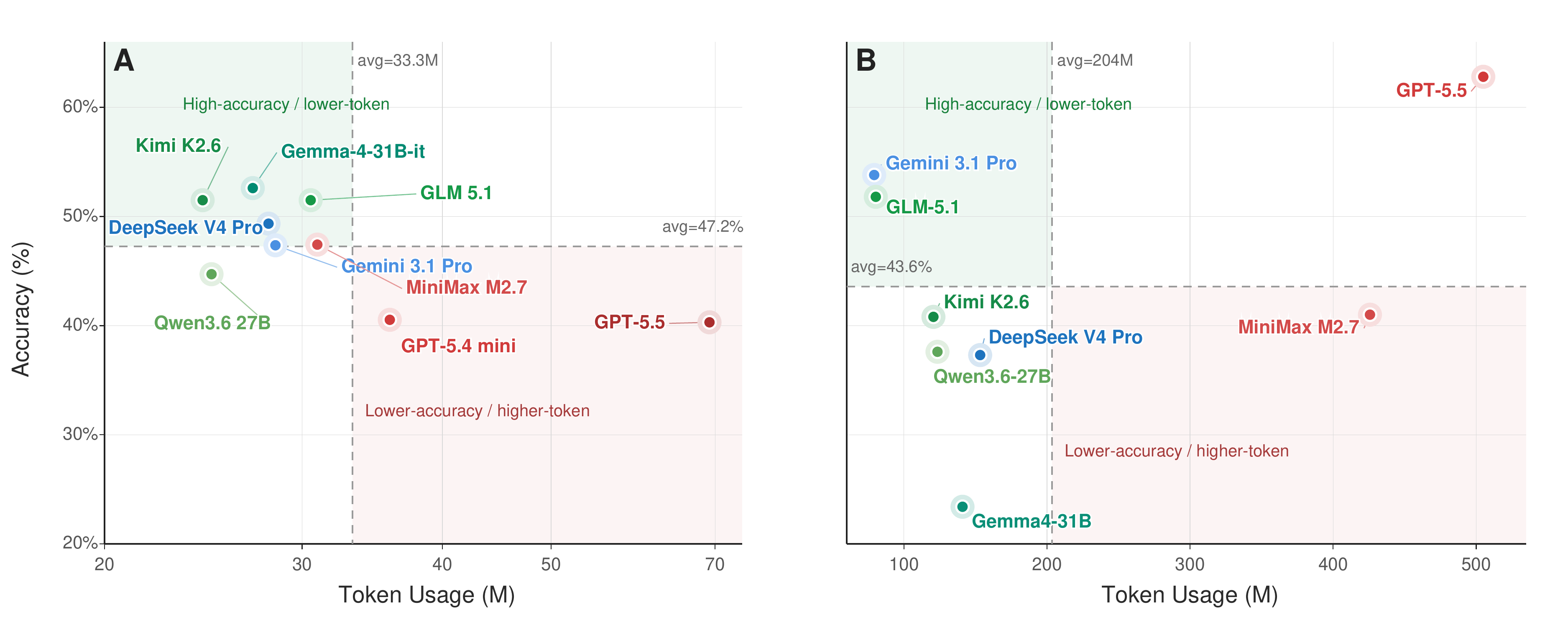}
\caption{\small \textbf{Accuracy versus total token usage across evaluated backbone models.} Total token usage is measured in millions of tokens, and lower usage indicates higher inference efficiency. Dashed lines mark the cross-model averages.}
\vspace{-4mm}
\label{fig:token_accuracy_efficiency}
\end{figure}

\paragraph{PersonaMem-Evo Efficiency.}
Figure~\ref{fig:token_accuracy_efficiency}(a) compares model accuracy against total token usage on PersonaMem-Evo, using token consumption as a proxy for inference efficiency. The results show that higher token usage does not reliably translate into higher accuracy. Gemma-4-31B-it achieves the best accuracy (52.6\%) while using fewer tokens than the cross-model average (27.1M vs. 28.8M), and Kimi K2.6 reaches similarly strong accuracy (51.5\%) with the lowest token usage among the evaluated models (24.5M). GLM 5.1 also attains high accuracy (51.5\%) but requires above-average token usage (30.5M). By contrast, GPT-5.4-Mini consumes more tokens (35.9M) while obtaining lower accuracy (40.5\%), and GPT-5.5 further amplifies this pattern, using 69.2M tokens while achieving 40.3\% accuracy. These results show that longer trajectories do not necessarily translate into better task performance. Overall, PersonaMem-Evo demonstrates that efficiency and accuracy should be evaluated jointly rather than treating token budget as a substitute for capability.

\paragraph{Terminal-Bench-Evo Efficiency.}
Figure~\ref{fig:token_accuracy_efficiency}(b) also reports the efficiency--accuracy trade-off on Terminal-Bench-Evo, where token budgets are substantially larger than in PersonaMem-Evo. GPT-5.5 achieves the highest terminal accuracy (62.8\%) but consumes by far the most tokens (505.0M), indicating that its strong performance comes with a substantial inference cost. In contrast, Gemini 3.1 Pro and GLM-5.1 achieve strong accuracy (53.8\% and 51.8\%, respectively) while remaining well below the terminal average token usage (79.2M and 80.3M vs. 203.6M). Kimi K2.6 and Qwen3.6-27B also use below-average token budgets (120.6M and 123.4M), but their accuracies remain below the terminal average of 43.6\%. Meanwhile, MiniMax M2.7 consumes a very large token budget (425.8M) while achieving only 41.0\% accuracy, and Gemma4-31B obtains the lowest terminal accuracy (23.4\%) despite moderate token usage (141.0M). These results further confirm that larger token budgets alone do not guarantee stronger task performance, making joint evaluation of accuracy and inference efficiency essential.

\section{Conclusion}

In this paper, we introduce EvoArena, a benchmark for evaluating LLM agents in environments where tasks, codebases, workflows, and user preferences evolve over time. We show that current agents remain limited under such non-stationary conditions, and propose EvoMem, a patch-based memory paradigm that records memory changes and their update context. Across EvoArena and standard agent benchmarks, EvoMem improves performance and better preserves evolving evidence. We hope this work supports future research on agents that can track changes, recover historical context, and adapt reliably in real-world evolving settings.

\newpage

\bibliographystyle{plain}
\bibliography{ref}

\newpage

\appendix

\startcontents[appendix]

\section*{List of Appendices}
\printcontents[appendix]{}{1}{}
\clearpage

\section{Limitations and Future Directions}
\label{sec:scope_future}

EvoArena and EvoMem take a first step toward evaluating and improving LLM agents in evolving environments. Our benchmark focuses on three representative forms of environment evolution: executable workflow changes, software repository evolution, and long-horizon preference shifts. These domains are intentionally chosen because they expose different version-aware capabilities, including adapting to updated interfaces, reasoning over accumulated codebase changes, and preserving temporally grounded user preferences.

This scope also highlights the broader significance of the problem. Environment evolution is a general challenge for deployed agents, and similar dynamics arise in robotics, embodied interaction, scientific workflows, multi-agent collaboration, and other long-running systems. Extending EvoArena to these settings would enable the community to study how agents track physical state changes, scientific protocol updates, collaborative dependencies, and evolving multi-agent roles. We view these extensions as an important direction for building agents that remain reliable across changing real-world conditions.

\section{Broader Impact}
\label{app:broader_impact}

This work studies how LLM agents behave in evolving environments, where workflows, codebases, tools, and user preferences change over time. As LLM agents are increasingly deployed in long-running settings, failures to track such changes can lead to brittle behavior, outdated decisions, or incorrect reuse of prior procedures. EvoArena provides a benchmark suite for evaluating these risks, and EvoMem offers a lightweight memory mechanism for preserving update histories as retrievable evidence. We expect this work to support the development of more reliable agents that can adapt to changing task conditions while retaining relevant prior knowledge.

A positive impact of this work is improved evaluation for agent robustness. Many current evaluations focus on fixed task snapshots, which can overestimate agent reliability in deployment settings where rules, interfaces, and user needs evolve. By modeling versioned environment changes, EvoArena can help researchers diagnose whether agents understand the current state of an environment and whether they can avoid applying obsolete behaviors. This may benefit applications such as software maintenance, enterprise workflow automation, long-horizon personal assistants, and other settings where agents must operate across repeated updates.

EvoMem may also improve transparency in memory-based agents. By recording what memory changed, why it changed, and which evidence triggered the update, patch histories make memory evolution more inspectable. This can help developers audit agent behavior, trace failures to specific memory updates, and distinguish current knowledge from previously valid knowledge. Such traceability is especially important in settings where agent actions affect user-facing decisions, organizational procedures, or software systems.

This work also raises potential risks. More persistent and adaptive memory can strengthen agent capabilities in ways that may be misused, for example by helping agents maintain long-term strategies in adversarial or manipulative tasks. Memory histories may also contain sensitive user information if deployed without appropriate privacy safeguards. In addition, patch retrieval can surface outdated or context-specific information if the agent fails to judge whether a prior state remains applicable. These risks highlight the need for careful access control, data minimization, retention policies, and explicit auditing of memory contents in real deployments.

EvoArena itself is intended for research evaluation and should be used with attention to dataset scope and domain coverage. The benchmark focuses on representative forms of environment evolution, including terminal workflows, software repositories, and long-horizon user preferences. Future work should extend this evaluation to additional domains, larger deployment horizons, and human-centered validation of agent behavior under changing conditions.

Overall, our results suggest that reliable agent deployment requires treating memory as an evolving record of grounded updates. We hope EvoArena and EvoMem encourage future work on agents that can adapt to changing environments in a traceable, auditable, and privacy-conscious manner.

\section{Declaration of LLM Usage}
\label{llm_usage}
We used LLMs to assist with writing, polishing, and coding support. All technical ideas, experimental design, analysis, and final manuscript decisions were made by the authors.

\section{Additional Analysis}

\subsection{PersonaMem-Evo Analysis}

\begin{table}[!htbp]
\centering
\begin{minipage}[t]{0.56\linewidth}
\centering
\resizebox{\linewidth}{!}{
\begin{tabular}{lccc}
\toprule
\textbf{OOD Type} & \textbf{Base} & \textbf{+EvoMem} & \textbf{$\Delta$} \\
\midrule
Conflict Resolution & 17.1\% & 24.8\% & $+7.6$ \\
Multi-Pattern Synthesis & 49.1\% & 57.8\% & $+8.6$ \\
Single-Pattern Transfer & 41.9\% & 43.6\% & $+1.7$ \\
Temporal Trajectory & 50.9\% & 46.6\% & $-4.3$ \\
\bottomrule
\end{tabular}
}
\vspace{1mm}

\textbf{(a) By OOD reasoning type.}
\end{minipage}
\hspace{0.035\linewidth}
\begin{minipage}[t]{0.38\linewidth}
\centering
\resizebox{\linewidth}{!}{
\begin{tabular}{lccc}
\toprule
\textbf{Difficulty} & \textbf{Base} & \textbf{+EvoMem} & \textbf{$\Delta$} \\
\midrule
L1 & 37.0\% & 41.7\% & $+4.6$ \\
L2 & 36.5\% & 37.7\% & $+1.2$ \\
L3 & 45.8\% & 50.3\% & $+4.5$ \\
\bottomrule
\end{tabular}
}
\vspace{1mm}

\textbf{(b) By difficulty level.}
\end{minipage}

\vspace{2mm}
\caption{PersonaMem-Evo performance breakdown by reasoning type and difficulty level using GPT-5.5.}
\label{tab:personamem_gpt55_breakdown}
\end{table}

To analyze where EvoMem helps under a stronger backbone, we further break down PersonaMem-Evo performance using GPT-5.5 by question type and difficulty level. Table~\ref{tab:personamem_gpt55_breakdown} shows that EvoMem improves across most categories, with the largest gains on \emph{multi-pattern synthesis} (+8.6\%) and \emph{conflict resolution} (+7.6\%). This suggests that, with a stronger model, the additional patch evidence can be more effectively used not only to preserve dispersed preference signals, but also to arbitrate among competing memories. EvoMem also improves \emph{single-pattern transfer} (+1.7\%), while performance decreases on \emph{temporal trajectory} (-4.3\%), indicating that retrieving historical patches can sometimes introduce ambiguity when the model must identify the latest valid preference state. The difficulty breakdown shows consistent gains across L1, L2, and L3, with larger improvements on L1 (+4.6\%) and L3 (+4.5\%).

\section{EvoArena Dataset Construction Details}

\subsection{Terminal-Bench-Evo: Task Analysis and Evolution Taxonomy}
\label{app:terminal}
\label{app:terminal_analysis}

\textbf{Goal.}
\emph{Terminal-Bench-Evo} is designed to evaluate whether an agent can adapt to an evolving executable terminal environment while solving the same underlying workflow objective. Unlike static terminal benchmarks where the task instruction, environment, and validation rules remain fixed, Terminal-Bench-Evo emphasizes \emph{temporal workflow evolution}: dependencies, toolchains, directory layouts, interfaces, I/O contracts, and validation requirements may change across versions. Agents must therefore infer the current environment state and adjust their solution, instead of reusing a previously valid procedure blindly.

Built on top of Terminal-Bench~\cite{terminal-bench}, Terminal-Bench-Evo converts all 89 original tasks into evolving workflow chains. For each original task $x_i$, we construct a chronologically ordered chain
\[
C_i = \{v_i^{(1)}, \ldots, v_i^{(m_i)}\},
\]
where each version $v_i^{(t)}$ contains a version-specific task instruction, executable environment, and validation tests. All versions in $C_i$ preserve the same high-level objective as $x_i$, but differ in the surrounding execution conditions, such as environment configuration, command-line interface, filesystem structure, dependency setup, input/output contract, or evaluation logic.

Each $v_i^{(t)}$ is an executable benchmark episode with a fixed instruction and environment; the environment does not change during that episode. The chain $C_i$ provides the release history used to organize inherited changes and to compute chain-level reliability. We summarize the construction pipeline in Algorithm~\ref{alg:terminal_bench_evo}. The pipeline consists of original-task conversion, version-update design, executable-environment realization, version-specific validation, quality filtering, and benchmark assembly. The following paragraphs describe the evaluation unit, evolution taxonomy, environment realization, metadata, validation procedure, and dataset statistics.

\begin{algorithm}[t]
\caption{Construction of Terminal-Bench-Evo}
\label{alg:terminal_bench_evo}
\begin{algorithmic}[1]
\Require Original Terminal-Bench tasks $\mathcal{X}=\{x_1,\ldots,x_{89}\}$; evolution taxonomy $\mathcal{G}$; validation executor $\mathcal{V}$
\Ensure Terminal-Bench-Evo benchmark $\mathcal{B}$
\State $\mathcal{B}\leftarrow \emptyset$
\For{each original task $x_i \in \mathcal{X}$}
    \State $a_i \leftarrow \textsc{AnalyzeTask}(x_i)$
    \Comment{Extract objective, environment, interfaces, I/O contracts, dependencies, and tests}
    \State $\Pi_i \leftarrow \textsc{ConstructEvolutionPlan}(x_i, a_i, \mathcal{G})$
    \State $E_i^{(0)} \leftarrow \textsc{InitializeEnvironment}(x_i)$
    \State $C_i \leftarrow \emptyset$
    \For{$t=1$ to $|\Pi_i|$}
        \State $\pi_i^{(t)} \leftarrow \Pi_i[t]$
        \State $(I_i^{(t)}, E_i^{(t)}, T_i^{(t)}, M_i^{(t)}) \leftarrow
        \textsc{RealizeVersion}(E_i^{(t-1)}, \pi_i^{(t)})$
        \Comment{Instantiate instruction, environment, tests, and metadata}
        \If{$\textsc{ValidateVersion}(I_i^{(t)}, E_i^{(t)}, T_i^{(t)}, \mathcal{V})$}
            \State $v_i^{(t)} \leftarrow (I_i^{(t)}, E_i^{(t)}, T_i^{(t)}, M_i^{(t)})$
            \State $C_i \leftarrow C_i \oplus v_i^{(t)}$
        \EndIf
    \EndFor
    \State $C_i \leftarrow \textsc{VerifyChainConsistency}(C_i)$
    \State $\mathcal{B} \leftarrow \mathcal{B} \cup \{C_i\}$
\EndFor
\State \Return $\mathcal{B}$
\end{algorithmic}
\end{algorithm}

\subsubsection{Evaluation Unit and Metrics}
\label{app:terminal_eval_unit_metrics}
Terminal-Bench-Evo separates the executable unit from the chain-level aggregation. A single benchmark episode corresponds to one versioned task instance $v_i^{(t)}$: the agent is given the version-specific instruction and container/workspace, then its terminal actions are evaluated by the tests attached to that version. Thus, ``evolving'' does not mean that the filesystem or dependency state mutates unexpectedly during one run; it means that the same workflow objective is released as a sequence of discrete, executable versions.

We report two metrics for this setting. \emph{Step accuracy} is the average success rate over all versioned task instances, so each solved version contributes one correct step. \emph{Chain accuracy} is stricter: a workflow chain $C_i$ is counted as correct only if all of its versions are solved under the evaluation protocol. Step accuracy measures adaptation to the current executable state, while chain accuracy measures whether an agent remains reliable across the full inherited workflow history. In memory-based experiments, versions from the same chain can be executed in chronological order so that earlier observations are available to later episodes; different chains remain independent.

\subsubsection{Original Task Analysis}
\label{app:terminal_task_analysis}
For each original task $x_i$, we first analyze its executable workflow components before constructing the evolution chain. Specifically, we identify the task objective, required commands, dependency and toolchain assumptions, relevant source and configuration files, input and output contracts, filesystem layout, and validation logic. This analysis yields a structured task state
\[
s_i = (o_i, e_i, f_i, d_i, c_i, r_i),
\]
where $o_i$ denotes the high-level objective, $e_i$ the executable environment, $f_i$ the relevant files and directory structure, $d_i$ the dependency and toolchain requirements, $c_i$ the interface and I/O contracts, and $r_i$ the validation rules.

The structured state $s_i$ serves as the basis for constructing version updates. Each update modifies one or more surrounding workflow components while preserving $o_i$, ensuring that different versions remain instances of the same underlying task instead of unrelated terminal problems. In addition to recurring evolution dimensions shared across many tasks, we also curate task-specific evolutions when the original workflow admits a natural semantic change, such as migrating the required implementation language from R to Python or changing the form in which the final artifact must be produced. These task-specific updates are recorded under the \emph{Other} category in our evolution statistics.

\subsubsection{Evolution Taxonomy}
\label{app:terminal_taxonomy}
Based on the structured task state $s_i$, we assign each non-initial version update a primary evolution label $g_i^{(t)} \in \mathcal{G}$. The taxonomy is organized around the main components of an executable terminal workflow that may naturally change over time: problem-facing contracts such as input/output paths, formats, schemas, or protocols; invocation interfaces such as CLI/API arguments and configuration entries; execution substrates such as dependencies, runtimes, package managers, toolchains, and environment capabilities; workspace organization such as directory layouts, module paths, imports, and staging flows; and task semantics or evaluation policies such as stricter constraints, security requirements, updated target rules, or additional edge cases.

We choose these dimensions because they jointly cover the main layers that define a terminal task beyond its high-level objective: what data the task consumes and produces, how the workflow is invoked, what execution environment it assumes, how the workspace is organized, and how correctness is evaluated. Each dimension can evolve independently while the underlying objective remains unchanged, making it suitable for constructing temporally evolving versions of the same task without turning them into unrelated new tasks. Updates that are natural to a specific workflow but do not fit the recurring categories are retained as task-specific evolutions and grouped under the \emph{Other} category. The full category distribution is reported in Table~\ref{tab:terminal_bench_evo_categories}.

\begin{table}[t]
\centering
\small
\begin{tabular}{lrr}
\toprule
\textbf{Evolution category} & \textbf{Count} & \textbf{Ratio} \\
\midrule
I/O-output path contract & 118 & 33.5\% \\
CLI/API surface change & 37 & 10.5\% \\
Format/protocol change & 35 & 9.9\% \\
Language/version/toolchain & 25 & 7.1\% \\
I/O-input path contract & 20 & 5.7\% \\
Staging/promotion flow & 19 & 5.4\% \\
Naming/import/module contract & 16 & 4.5\% \\
Workspace/source layout & 12 & 3.4\% \\
Security/policy hardening & 9 & 2.6\% \\
Evaluation target rule & 7 & 2.0\% \\
Environment capability toggle & 3 & 0.9\% \\
Other task-specific evolution & 51 & 14.5\% \\
\midrule
Total & 352 & 100.0\% \\
\bottomrule
\end{tabular}
\caption{Distribution of evolution categories in Terminal-Bench-Evo. Counts are computed over non-initial version updates, excluding the first version of each workflow chain.}
\label{tab:terminal_bench_evo_categories}
\end{table}

\begin{table}[t]
\centering
\small
\begin{tabular}{lrr}
\toprule
\textbf{Workflow component group} & \textbf{Count} & \textbf{Ratio} \\
\midrule
I/O or protocol contract & 173 & 49.1\% \\
Workspace, module, or staging flow & 47 & 13.4\% \\
CLI/API invocation surface & 37 & 10.5\% \\
Dependency, toolchain, or capability & 28 & 8.0\% \\
Semantic, policy, or evaluation rule & 16 & 4.6\% \\
Other task-specific evolution & 51 & 14.5\% \\
\midrule
Total & 352 & 100.0\% \\
\bottomrule
\end{tabular}
\caption{Grouped evolution statistics for Terminal-Bench-Evo. Groups aggregate the fine-grained categories in Table~\ref{tab:terminal_bench_evo_categories}.}
\label{tab:terminal_bench_evo_component_groups}
\end{table}

\subsection{Terminal-Bench-Evo: Version Construction and Validation}
\label{app:terminal_validation}

\subsubsection{Version-Chain Construction}
\label{app:terminal_version_construction}
For each original task $x_i$, we use an agent-assisted construction pipeline to generate a set of candidate version updates grounded in the structured task state $s_i$. Each candidate describes a plausible evolution of the original workflow, such as modifying the invocation interface, dependency setup, filesystem layout, I/O contract, execution constraints, or validation requirements. We inspect these candidates for semantic plausibility and objective preservation, retaining only updates that represent natural changes to the surrounding terminal workflow while keeping the high-level objective $o_i$ unchanged.

For each retained update, we instantiate the full task materials, including the revised instruction, executable environment, validation tests, reference solution, and metadata. The pipeline provides initial task materials and solutions when applicable, while reference solutions for difficult or failure-prone versions are constructed or corrected by human experts to ensure reliability. The retained updates are then ordered from simpler to more complex changes and composed into a single evolution chain. During chain assembly, version $t$ is constructed by applying its update to the realized environment of version $t-1$. Unchanged files, dependencies, paths, permissions, interface conventions, and validation assumptions are inherited, while only the components specified by the current update are patched. Each stage therefore reflects the accumulated workflow history, but is still required to remain a complete and independently solvable terminal task. After assembling the chain, we run oracle validation to verify executability and test correctness, iteratively repair environment or validation issues, and finally perform a semantic consistency check to ensure that the full chain forms a coherent workflow evolution rather than a collection of unrelated variants.

\subsubsection{Quality Control}
\label{app:terminal_quality_control}
We apply multiple rounds of quality control to ensure that each EVO stage is a valid, executable, and independently solvable Terminal-Bench task instance, beyond a prompt-only variant. For each retained update, we instantiate the full task materials, including the revised instruction, executable environment, source and configuration files, dependency specifications, input/output paths, validation tests, reference solution, and metadata when applicable. We then check that the instruction, environment, and tests are mutually consistent, and that the update preserves the original high-level objective while introducing a meaningful workflow change.

\subsubsection{Oracle Validation and Solvability Verification}
\label{app:terminal_oracle_validation}
Each realized version is verified through oracle execution. The reference solution is run in the constructed environment, and its outputs are evaluated using the version-specific validation tests. We reject or repair versions with build failures, nondeterministic behavior, ambiguous instructions, incomplete dependencies, inconsistent validation logic, or mismatches between the task specification and expected outputs. After composing versions into a chain, we further check that later EVO stages correctly inherit earlier modifications while remaining complete standalone tasks. This process ensures that every retained version is executable, testable, and solvable, and that each chain forms a coherent workflow evolution instead of a collection of unrelated variants.

\subsubsection{Metadata and Dataset Statistics}
\label{app:terminal_metadata_stats}
For each versioned task instance, we store structured metadata describing its position and role in the evolution chain, including the original Terminal-Bench task identifier, chain identifier, EVO stage index, primary evolution category, changed workflow components, instruction-change summary, environment-change summary, validation-test summary, and reference-solution information. These annotations support both dataset analysis and error analysis by making it possible to identify which types of workflow evolution are most challenging for agents.

Terminal-Bench-Evo contains 89 evolving workflow chains and 441 versioned task instances, with an average of 4.96 versions per chain and a median chain length of 5. Each chain contains four to five validated versions: 85 chains contain five versions and 4 chains contain four versions. Table~\ref{tab:terminal_bench_evo_categories} reports the fine-grained evolution categories over non-initial version updates, Table~\ref{tab:terminal_bench_evo_component_groups} reports grouped workflow-component statistics, and Table~\ref{tab:terminal_bench_evo_statistics} summarizes the overall dataset statistics.

\begin{table}[t]
\centering
\small
\begin{tabular}{lr}
\toprule
\textbf{Statistic} & \textbf{Value} \\
\midrule
Original Terminal-Bench tasks & 89 \\
Evolving workflow chains & 89 \\
Versioned task instances & 441 \\
Non-initial version updates & 352 \\
Average versions per chain & 4.96 \\
Median versions per chain & 5 \\
Minimum versions per chain & 4 \\
Maximum versions per chain & 5 \\
Four-version chains & 4 \\
Five-version chains & 85 \\
Executable task environments & 441 \\
\bottomrule
\end{tabular}
\caption{Summary statistics of Terminal-Bench-Evo.}
\label{tab:terminal_bench_evo_statistics}
\end{table}

\subsection{SWE-Chain-Evo: Repository and Milestone Construction}
\label{app:swe}
\label{app:swe_milestone}

\paragraph{Goal and task formulation.}
SWE-Chain-Evo is designed to evaluate agents under continuous software evolution, where the evolving environment is the codebase itself. In this setting, APIs, dependencies, tests, implementation choices, and architectural constraints introduced by earlier changes become part of the environment that later tasks must operate on. SWE-Chain-Evo therefore organizes software tasks into temporally ordered chains that require agents to adapt to accumulated codebase changes.

Formally, each chain is defined as
\[
C_i = \{m_i^{(1)}, m_i^{(2)}, \ldots, m_i^{(T_i)}\},
\]
where each milestone \(m_i^{(t)}\) represents a semantically coherent development objective in repository \(i\). Given the repository state \(S_i^{(t-1)}\) before the milestone and a task instruction \(q_i^{(t)}\), the agent must produce a code modification that satisfies the milestone requirement. After applying the reference milestone update, the repository transitions to \(S_i^{(t)}\), which becomes the starting environment for the next task. This formulation preserves the temporal and dependency structure of real software development, allowing us to test whether agents can adapt to new requirements while maintaining compatibility with previously accumulated codebase changes.

\subsubsection{Repository Collection and Domain Coverage}
\label{app:swe_repository_collection}
We manually curated 50 candidate GitHub repositories spanning different software domains, programming ecosystems, and codebase environments, while requiring them to be actively maintained and sufficiently well-tested. From these candidates, we retained high-quality SWE-Chain-Evo instances from 26 repositories after milestone construction, executability checks, and test-stability filtering. The final repositories cover web frameworks, cloud infrastructure and distributed systems, observability, security and networking, developer tools and code quality, data processing, scientific computing, and testing frameworks. For dataset characterization, each repository is assigned a single primary domain label, although some repositories may naturally span multiple categories. Table~\ref{tab:swe_chain_domains} summarizes the final repository coverage.

\begin{table}[t]
\centering
\small
\resizebox{\textwidth}{!}{
\begin{tabular}{lrrrl}
\toprule
Domain & \#Repos & \#Chains & \#Milestones & Repositories \\
\midrule
Web frameworks & 5 & 14 & 39 & aiohttp, express, gin, fiber, echo \\
\makecell[l]{Cloud infrastructure / distributed systems /\\ observability} 
& 10 & 20 & 60 
& \makecell[l]{argo-cd, containerd, coredns, loki, terraform, istio,\\ nats-server, opentelemetry-collector, prometheus, traefik} \\
Security / networking & 3 & 3 & 7 & go-jose, vault, mitmproxy \\
Developer tools / CLI / code quality & 5 & 8 & 20 & click, pipx, black, cobra, sqlfluff \\
Data processing / workflow / scientific computing & 2 & 2 & 6 & dagster, xarray \\
Testing frameworks & 1 & 1 & 3 & pytest \\
\midrule
Total & 26 & 48 & 135 & -- \\
\bottomrule
\end{tabular}
}
\caption{Repository and domain coverage of SWE-Chain-Evo. Repository domains are assigned using a single primary label for dataset characterization.}
\label{tab:swe_chain_domains}
\end{table}

\subsubsection{Milestone Construction}
\label{app:swe_milestone_construction}
For each retained repository, we extract continuous update windows from its commit history and construct milestones within each window. A milestone is designed to capture a single coherent development objective with clear semantic scope. We first group nearby commits by the semantic relatedness of their commit messages, and then inspect the corresponding code changes to verify that they contribute to the same functionality, bug fix, refactoring goal, or maintenance objective. This grouping process is manually checked to ensure that each milestone corresponds to a realistic and semantically coherent software change. Changes that do not contribute to the milestone objective, such as unrelated formatting edits, incidental dependency updates, merge artifacts, or other noisy modifications, are filtered out.

After grouping, we manually verify each candidate milestone for semantic coherence, executability, and evaluation reliability. For valid milestones, we synthesize a SWE-bench-style task description from the pre-milestone repository state, specifying the intended functionality or maintenance goal without exposing the reference patch. Milestones that are ambiguous, overly broad, too trivial, or difficult to evaluate reliably are discarded. Algorithm~\ref{alg:swe_chain_milestone} summarizes this construction process.

\begin{algorithm}[t]
\caption{Milestone Construction for SWE-Chain-Evo}
\label{alg:swe_chain_milestone}
\begin{algorithmic}[1]
\Require Repository $r$; commit history $H_r$; validation pipeline $\mathcal{V}$
\Ensure Milestone set $M_r$

\State $M_r \gets \emptyset$
\State $W_r \gets \textsc{ExtractContinuousWindows}(H_r)$

\For{each update window $w \in W_r$}
    \State $G_w \gets \textsc{GroupByCommitSemantics}(w)$
    \For{each candidate commit group $g \in G_w$}
        \State $\Delta_g \gets \textsc{InspectCodeChanges}(g)$

        \If{$\neg\,\textsc{CoherentObjective}(g,\Delta_g)$}
            \State \textbf{continue}
        \EndIf

        \State $g' \gets \textsc{FilterIrrelevantChanges}(g,\Delta_g)$

        \If{$\textsc{AmbiguousOrUnreliable}(g')$}
            \State \textbf{continue}
        \EndIf

        \State $q \gets \textsc{SynthesizeTaskDescription}(g')$

        \If{$\textsc{ManualVerify}(q,g')$ \textbf{and} $\textsc{ExecutableAndStable}(q,g';\mathcal{V})$}
            \State $M_r \gets M_r \cup \{(q,g')\}$
        \EndIf
    \EndFor
\EndFor

\State \Return $M_r$
\end{algorithmic}
\end{algorithm}

\subsection{SWE-Chain-Evo: Evaluation Packaging and Chain Assembly}
\label{app:swe_evaluation}

\subsubsection{Task Format and Problem Statement Construction}
\label{app:swe_task_format}
Each milestone is converted into a SWE-bench-style task instance. We aggregate the milestone's commit messages, code-change summary, and local repository context into an initial problem statement, and then manually refine it to ensure that it faithfully describes the intended functionality, bug fix, refactoring goal, or maintenance requirement. The final instance exposes only the pre-milestone repository snapshot and the natural-language problem statement to the agent; the reference patch is reserved for dataset construction and evaluation.

\subsubsection{Docker Evaluation Environment}
\label{app:swe_docker_evaluation}
For evaluation, we build Docker-based environments for the milestone instances using the RepoLaunch pipeline from SWE-Evo. The pipeline installs repository dependencies, configures build and test commands, and runs automated test discovery to identify Fail-to-Pass and Pass-to-Pass tests. Fail-to-Pass tests validate behavior that should be introduced or repaired by the milestone, while Pass-to-Pass tests check that existing behavior is preserved. When the automatic pipeline misses necessary test signals, we inspect the code semantics and manually supplement tests if the target behavior is clear. We remove instances with flaky builds, ambiguous failures, incomplete fields, missing dependencies, or unstable test outcomes.

\subsubsection{Chain Assembly}
\label{app:swe_chain_assembly}
After individual milestones are packaged and validated, we assemble them into temporally ordered chains within the same repository. Each chain is selected from a continuous update window and follows the repository's natural commit order. Given a chain \(C_i = \{m_i^{(1)}, \ldots, m_i^{(T_i)}\}\), the repository snapshot before \(m_i^{(1)}\) serves as the initial state. After each step, the corresponding reference milestone update is applied to obtain the repository state for the next step. Thus, later tasks are evaluated on codebases that already include accumulated changes from earlier milestones.

We manually review each assembled chain for temporal consistency, semantic continuity, executable dependency, and appropriate difficulty. Chains are discarded if later milestones do not depend on or naturally follow the preceding repository state, if the sequence mixes unrelated development objectives, or if any step cannot be reliably evaluated in the constructed Docker environment. This ensures that SWE-Chain-Evo evaluates cumulative codebase evolution: agents must satisfy new requirements while respecting constraints introduced by earlier development steps.

\subsubsection{Dataset Statistics}
\label{app:swe_dataset_statistics}
Table~\ref{tab:swe_chain_overall_stats} summarizes the overall statistics of SWE-Chain-Evo. The dataset contains 48 milestone chains from 26 repositories, with 135 milestone steps in total. Each chain contains 2--4 milestones, with an average of 2.81 and a median of 3 milestones per chain. Most chains contain three steps, reflecting our goal of constructing short but genuinely sequential software-evolution trajectories.

\begin{table}[t]
\centering
\small
\begin{tabular}{lr}
\toprule
Statistic & Value \\
\midrule
Candidate repositories & 50 \\
Retained repositories & 26 \\
Chains & 48 \\
Milestones & 135 \\
Average milestones per chain & 2.81 \\
Median milestones per chain & 3 \\
Milestone range per chain & 2--4 \\
Average commits per milestone & 1.54 \\
Median commits per milestone & 1 \\
Commit range per milestone & 1--6 \\
Average modified files per milestone & 1.99 \\
Median modified files per milestone & 1 \\
Modified-file range per milestone & 1--12 \\
Average changed lines per milestone & 36.45 \\
Median changed lines per milestone & 18 \\
Changed-line range per milestone & 1--602 \\
Average Fail-to-Pass tests per milestone & 3.12 \\
Average Pass-to-Pass tests per milestone & 6.16 \\
\bottomrule
\end{tabular}
\caption{Summary statistics of SWE-Chain-Evo. Patch statistics are computed from the solution code patch. Changed lines count added plus deleted lines, excluding diff headers.}
\label{tab:swe_chain_overall_stats}
\end{table}

Table~\ref{tab:swe_chain_length_stats} reports the chain-length distribution. The majority of chains contain three milestones, while shorter and longer chains are included to cover different degrees of codebase evolution. At the milestone level, most tasks are compact but non-trivial: 89 of the 135 milestones contain a single commit, while the largest milestone contains 6 commits.

\begin{table}[t]
\centering
\small
\begin{tabular}{lr}
\toprule
Chain length & \#Chains \\
\midrule
2 milestones & 12 \\
3 milestones & 33 \\
4 milestones & 3 \\
\bottomrule
\end{tabular}
\caption{Chain-length distribution in SWE-Chain-Evo.}
\label{tab:swe_chain_length_stats}
\end{table}

For evaluation, all 135 milestones include at least one Fail-to-Pass test, with an average of 3.12 tests per milestone, a median of 1, and a range of 1--30. Pass-to-Pass tests are available for most milestones, with an average of 6.16 tests, a median of 3, and a range of 0--61; only 4 milestones do not include Pass-to-Pass tests. These statistics indicate that each milestone has a direct target-behavior signal, while most also include regression checks for preserving existing functionality.

\subsection{PersonaMem-Evo: Persona and Interaction History Construction}
\label{sec:persona_mem_evo_construction}
\label{app:persona_history}

\textbf{Goal.}
\emph{PersonaMem-Evo} is designed to evaluate whether an agent can infer, track, and generalize a user's evolving latent preferences from long-horizon interaction histories. Unlike factual memory benchmarks where target information is often explicitly stated, PersonaMem-Evo emphasizes \emph{implicit behavioral evidence}: user preferences are revealed through choices, constraints, repeated requests, contextual behavior, and decision patterns across realistic conversations.

Built on top of PersonaMem-v2~\cite{jiang2025personamem}, PersonaMem-Evo extends static preference memory with structured temporal preference evolution and preference inference. For each user persona $u_i$, we construct a long-context interaction history $H_i$ and a set of multiple-choice questions
\[
Q_i^{\mathrm{ood}} = \{q_{i1}^{\mathrm{ood}}, \ldots, q_{im_i}^{\mathrm{ood}}\},
\]
where answering each question requires abstracting user-specific behavioral patterns from $H_i$ and applying them to a novel decision setting.

We summarize the construction pipeline in Algorithm~\ref{alg:persona_mem_evo}. The pipeline consists of seed persona sampling, persona expansion, preference synthesis, implicit interaction generation, temporal preference evolution, question generation, dual-blind filtering, and benchmark assembly. Algorithm~\ref{alg:preference_evolution} describes the construction of multi-step preference trajectories, and Algorithm~\ref{alg:ood_generation} describes question generation with dual-blind validation.

\subsubsection{Seed Persona Expansion and Preference Cleaning}
\label{app:persona_seed_preferences}
Following PersonaMem-v2, we sample seed personas from PersonaHub~\cite{ge2024scaling}, specifically the \texttt{Persona\_Hub\_200000.jsonl} subset. PersonaHub provides a large-scale repository of synthetic personas, which allows the benchmark to cover diverse demographic, occupational, lifestyle, and personality backgrounds. Each seed persona $u_i^{(0)}$ is expanded into a richer structured profile
\[
u_i = \mathcal{E}\!\left(u_i^{(0)}\right),
\]
where $\mathcal{E}(\cdot)$ denotes the persona expansion operator. The expanded profile contains demographic attributes, occupation, personality traits, lifestyle information, social background, and personal context. These fields provide diverse grounding for generating realistic behavioral preferences while keeping the benchmark synthetic and privacy-safe.

From the expanded persona $u_i$, we synthesize a preference set
\[
P_i =
P_i^{\mathrm{st}}
\cup
P_i^{\mathrm{anti}}
\cup
P_i^{\mathrm{neu}}
\cup
P_i^{\mathrm{th}}
\cup
P_i^{\mathrm{med}},
\]
where $P_i^{\mathrm{st}}$, $P_i^{\mathrm{anti}}$, and $P_i^{\mathrm{neu}}$ denote stereotypical, anti-stereotypical, and neutral preferences, respectively, while $P_i^{\mathrm{th}}$ and $P_i^{\mathrm{med}}$ denote therapy-related and health/medical preferences. Anti-stereotypical preferences are especially important because they reduce the reliability of demographic priors and encourage models to rely on behavioral evidence from the interaction history.

Generated preferences are post-processed using validation and deduplication routines. We remove empty preferences, duplicated entries, and semantically inconsistent preferences across categories. Each preference is also assigned a coarse topical label $z \in \mathcal{Z}$, which supports semantic coverage analysis and downstream stratification.

\begin{algorithm}[t]
\caption{Construction of PersonaMem-Evo}
\label{alg:persona_mem_evo}
\begin{algorithmic}[1]
\Require Seed personas $\mathcal{U}^{(0)}=\{u_1^{(0)},\ldots,u_N^{(0)}\}$; LLM generator $\mathcal{G}$; validator $\mathcal{V}$
\Ensure OOD benchmark set $\mathcal{B}^{\mathrm{ood}}$

\State $\mathcal{B}^{\mathrm{ood}} \gets \emptyset$
\For{each seed persona $u_i^{(0)} \in \mathcal{U}^{(0)}$}
    \State $u_i \gets \textsc{ExpandPersona}(u_i^{(0)};\mathcal{G})$
    \State $P_i \gets \textsc{GeneratePreferences}(u_i;\mathcal{G})$
    \State $P_i \gets \textsc{CleanAndDeduplicate}(P_i;\mathcal{V})$
    \State $P_i \gets \textsc{AssignTopics}(P_i;\mathcal{G})$

    \State $\mathcal{C}_i \gets \emptyset$
    \For{each preference $p \in P_i$}
        \State $c \gets \textsc{GenerateImplicitConversation}(u_i,p;\mathcal{G})$
        \State $\mathcal{C}_i \gets \mathcal{C}_i \cup \{c\}$

        \If{$\textsc{EligibleForEvolution}(p)$}
            \State $(\phi,\mathcal{C}^{\mathrm{evo}}) \gets \textsc{SamplePreferenceEvolution}(u_i,p;\mathcal{G})$
            \State $\mathcal{C}_i \gets \mathcal{C}_i \cup \mathcal{C}^{\mathrm{evo}}$
        \EndIf
    \EndFor

    \State $H_i \gets \textsc{BuildLongContextHistory}(\mathcal{C}_i)$
    \State $Q_i^{\mathrm{ood}} \gets \textsc{GenerateOODQuestions}(u_i,H_i;\mathcal{G},\mathcal{V})$

    \For{each validated OOD question $q \in Q_i^{\mathrm{ood}}$}
        \State $\mathcal{B}^{\mathrm{ood}} \gets \mathcal{B}^{\mathrm{ood}} \cup \{(H_i,q)\}$
    \EndFor
\EndFor

\State \Return $\mathcal{B}^{\mathrm{ood}}$
\end{algorithmic}
\end{algorithm}

\subsubsection{Implicit Interaction History Construction}
\label{app:persona_implicit_interactions}
Given the preference set $P_i$, we generate a collection of interaction episodes
\[
\mathcal{C}_i = \{c_{i1}, \ldots, c_{in_i}\}.
\]
Each episode $c$ is a multi-turn dialogue
\[
c = \left[(r_1, x_1), (r_2, x_2), \ldots, (r_T, x_T)\right],
\]
where $r_t \in \{\texttt{user}, \texttt{assistant}\}$ denotes the speaker role and $x_t$ denotes the utterance. For each preference $p \in P_i$, the generator creates one or more episodes in which $p$ is expressed indirectly through user behavior. Interaction scenarios include daily-life conversations, email-style communication, writing assistance, professional communication, recommendation requests, repeated knowledge queries, troubleshooting, translation, and multimodal image-grounded conversations.

To avoid reducing the benchmark to explicit preference lookup, the generator is instructed to express the target preference implicitly through choices, constraints, tone, repeated requests, or contextual behavior, without stating it as an isolated declarative fact. With some probability, an episode is framed as describing another person's preference, yielding a label
\[
\texttt{who} \in \{\texttt{self}, \texttt{others}\}.
\]
This distinction tests whether models can separate user-specific memories from information about other entities.

Some short snippets are expanded into longer multi-turn interactions to improve conversational realism. In addition, a subset of episodes is followed by explicit forgetting requests, where the user asks the assistant not to retain or use a previously mentioned preference. These cases test whether memory systems can handle user-controlled preference removal or suppression.

The generated episodes are assembled into a persona-level long-context history
\[
H_i = \Pi(\mathcal{C}_i),
\]
where $\Pi(\cdot)$ denotes the history construction operator. In practice, $\Pi$ preserves episode boundaries while concatenating episodes into a single mixed-topic interaction stream. The resulting histories are rendered under 32k and 128k context settings.

\subsection{PersonaMem-Evo: Preference Evolution and OOD Benchmark Construction}
\label{app:persona_ood}

\subsubsection{Temporal Preference Evolution}
\label{app:persona_temporal_evolution}
A central property of PersonaMem-Evo is that user preferences are not assumed to be static. For a large subset of self-related preferences, we sample a structured evolution plan
\[
\phi = (f, k, \tau),
\]
where $f \in \mathcal{F}$ is a change family, $k$ is the number of sequential updates, and
\[
\tau = \left(p^{(0)}, p^{(1)}, \ldots, p^{(k)}\right)
\]
is the resulting preference trajectory. Here, $p^{(0)}$ is the initial preference, and each updated state $p^{(t)}$ is generated conditioned on the previous state $p^{(t-1)}$ and a family-specific transition rule.

We sample $k \in \{1,\ldots,5\}$ updates for each evolved preference chain. Approximately 85\% of eligible self-related preferences receive a structured \textsc{ChangePlan}. The goal is not to create arbitrary binary reversals, but to model natural preference evolution caused by new experiences, changing constraints, updated routines, contextual conditions, or temporal validity.

\begin{table}[t]
\centering
\small
\begin{tabular}{p{0.22\linewidth}p{0.12\linewidth}p{0.27\linewidth}p{0.30\linewidth}}
\toprule
\textbf{Change family} & \textbf{Ratio} & \textbf{Transition pattern} & \textbf{Example} \\
\midrule
Same-object attitude revision
& 30\%
& The attitude toward the same object is revised due to new context or experience.
& Prefers intense hiking $\rightarrow$ avoids intense hiking after an injury $\rightarrow$ prefers light trail walks after recovery. \\

Object replacement
& 30\%
& The preferred object changes within the same semantic category.
& Espresso $\rightarrow$ pour-over coffee $\rightarrow$ matcha. \\

Conditional preference shift
& 13\%
& A general preference becomes conditional on a situational constraint.
& Likes reading $\rightarrow$ reads only when alone $\rightarrow$ reads mostly on weekends. \\

Attribute shift
& 13\%
& The preferred attribute changes while the broader category remains related.
& Prefers red cars $\rightarrow$ prefers blue cars $\rightarrow$ prefers SUVs. \\

Temporal-validity shift
& 14\%
& The preference becomes valid only under a time span, routine, or recency condition.
& Exercises daily $\rightarrow$ exercises only in summer $\rightarrow$ exercises in the morning. \\
\bottomrule
\end{tabular}
\caption{Five change families used to construct multi-step preference trajectories.}
\label{tab:persona_mem_evo_change_families}
\end{table}

For each updated preference state $p^{(t)}$, we synthesize a new conversational episode that implicitly reflects the updated state. We also store metadata including the change family $f$, total trajectory length $k$, current step $t$, previous preference state $p^{(t-1)}$, updated preference state $p^{(t)}$, and the full trajectory $\tau$. Thus, the final history $H_i$ contains not only stable user traits but also temporally grounded evidence of how preferences evolve over time. This design supports evaluation of current-state recovery, outdated-state disambiguation, and reasoning over the direction of preference change.

\begin{algorithm}[t]
\caption{Sampling Multi-Step Preference Evolution}
\label{alg:preference_evolution}
\begin{algorithmic}[1]
\Require Persona $u_i$; initial preference $p$; generator $\mathcal{G}$
\Ensure Change plan $\phi=(f,k,\tau)$; trajectory conversations $\mathcal{C}^{\mathrm{evo}}$

\State Sample change family $f \sim \mathrm{Cat}(\mathcal{F})$
\State Sample number of updates $k \in \{1,\ldots,5\}$
\State Initialize trajectory $\tau \gets (p^{(0)})$, where $p^{(0)}=p$
\State $\mathcal{C}^{\mathrm{evo}} \gets \emptyset$

\For{$s=1$ to $k$}
    \State $(p^{(s)}, \eta^{(s)}) \gets \textsc{ApplyChangeFamily}(p^{(s-1)},f,u_i;\mathcal{G})$
    \State $c^{(s)} \gets \textsc{GenerateImplicitConversation}(u_i,p^{(s)},\eta^{(s)};\mathcal{G})$
    \State Annotate $c^{(s)}$ with $(f,k,s,p^{(s-1)},p^{(s)},\eta^{(s)})$
    \State $\tau \gets \tau \oplus p^{(s)}$
    \State $\mathcal{C}^{\mathrm{evo}} \gets \mathcal{C}^{\mathrm{evo}} \cup \{c^{(s)}\}$
\EndFor

\State $\phi \gets (f,k,\tau)$
\State \Return $\phi,\mathcal{C}^{\mathrm{evo}}$
\end{algorithmic}
\end{algorithm}

\subsubsection{OOD Preference-Inference Question Generation}
\label{app:persona_ood_generation}
PersonaMem-Evo includes an OOD preference-inference stage to test whether a model can generalize from observed behavioral patterns to new decision settings. These questions cannot be answered from the persona summary alone or from generic commonsense knowledge; instead, they require abstraction over the long interaction history.

Each OOD question is represented as
\[
q = (x^{q}, a^{+}, \mathcal{A}^{-}, t, d),
\]
where $x^q$ is the query, $a^+$ is the correct answer, $\mathcal{A}^{-}$ is a set of distractors, $t$ is the OOD question type, and $d$ is the complexity level. We consider four OOD types:
\[
\mathcal{T} =
\left\{
\begin{array}{l}
\texttt{single-pattern-transfer},\\
\texttt{multi-pattern-synthesis},\\
\texttt{conflict-resolution},\\
\texttt{temporal-trajectory}
\end{array}
\right\},
\qquad
\mathcal{D} = \{\mathrm{L1}, \mathrm{L2}, \mathrm{L3}\}.
\]

The four OOD types test complementary reasoning abilities:
\begin{itemize}
    \item \textbf{Single-pattern transfer}: transfer one uncommon or counter-stereotypical preference pattern to a new domain.
    \item \textbf{Multi-pattern synthesis}: combine multiple observed preferences, including at least one counter-stereotypical pattern, into a narrower inference.
    \item \textbf{Conflict resolution}: infer the user's implicit priority structure when multiple preferences conflict.
    \item \textbf{Temporal trajectory prediction}: infer the directional evolution of a user's preferences and extrapolate that trajectory to a new domain.
\end{itemize}

The three complexity levels capture different forms of generalization. L1 corresponds to direct transfer, L2 requires cross-domain linking, and L3 requires more abstract meta-reasoning over multiple behavioral signals. These levels define qualitatively different reasoning categories, with increasing abstraction across levels.

\begin{algorithm}[t]
\caption{OOD Question Generation with Dual-Blind Filtering}
\label{alg:ood_generation}
\begin{algorithmic}[1]
\Require Persona $u_i$; long-context history $H_i$; generator $\mathcal{G}$; validator $\mathcal{V}$
\Ensure Validated OOD question set $Q_i^{\mathrm{ood}}$

\State $Q_i^{\mathrm{ood}} \gets \emptyset$
\State Define OOD types $\mathcal{T}$ and complexity levels $\mathcal{D}$

\For{each question type $t \in \mathcal{T}$}
    \For{each complexity level $d \in \mathcal{D}$}
        \For{$r = 1$ to $\textsc{TargetCount}(t,d)$}
            \State $q \gets \textsc{DraftOODQuestion}(u_i,H_i,t,d;\mathcal{G})$
            \If{$q=\varnothing$}
                \State \textbf{continue}
            \EndIf

            \State $q \gets \textsc{BuildAnswerOptions}(q;\mathcal{G})$
            \If{$\neg\,\textsc{BalancedOptions}(q)$}
                \State \textbf{continue}
            \EndIf

            \If{$\textsc{PersonaBlindCorrect}(q,u_i;\mathcal{V})$}
                \State \textbf{continue}
            \EndIf

            \If{$\textsc{NoContextCorrect}(q;\mathcal{V})$}
                \State $q \gets \textsc{AdversarialRewrite}(q;\mathcal{G})$
                \If{$q=\varnothing$ \textbf{or} $\textsc{NoContextCorrect}(q;\mathcal{V})$}
                    \State \textbf{continue}
                \EndIf
            \EndIf

            \State $Q_i^{\mathrm{ood}} \gets Q_i^{\mathrm{ood}} \cup \{q\}$
        \EndFor
    \EndFor
\EndFor

\State \Return $Q_i^{\mathrm{ood}}$
\end{algorithmic}
\end{algorithm}

\subsubsection{Dual-Blind Filtering}
\label{app:persona_dual_blind_filtering}
To ensure that accepted OOD questions require conversational memory, we apply a dual-blind validation procedure that filters shortcut-solvable cases. Given a candidate question $q$, we define two validators:
\[
V_{\mathrm{pers}}(q,u_i) \in \{0,1\},
\qquad
V_{\emptyset}(q) \in \{0,1\}.
\]
Here, $V_{\mathrm{pers}}(q,u_i)=1$ indicates that the question can be answered correctly using only the persona profile $u_i$ without the conversation history, while $V_{\emptyset}(q)=1$ indicates that the question can be answered correctly with neither the persona profile nor the history. We accept a question only if
\[
V_{\mathrm{pers}}(q,u_i)=0
\quad \text{and} \quad
V_{\emptyset}(q)=0.
\]
The first constraint filters out items answerable from demographic priors, explicit persona facts, or stereotypes. The second constraint filters out items answerable from commonsense reasoning, answer-option artifacts, or stylistic imbalance.

When a candidate fails the no-context validation but is otherwise useful, we apply adversarial option rewriting. The rewrite makes the correct answer less guessable from generic priors and makes distractors more plausible, while preserving the intended behavioral inference. The rewritten item is accepted only if it passes both blind checks.

\subsubsection{Answer-Option Construction}
\label{app:persona_answer_options}
Each accepted question is packaged as a four-way multiple-choice item:
\[
\mathcal{A}(q)=\{a^{+}\}\cup\mathcal{A}^{-},
\qquad
|\mathcal{A}(q)|=4.
\]
The generator enforces answer-option balance so that no candidate is trivially identifiable by length, specificity, formatting, or vagueness. Distractors are designed to be plausible and semantically close to the correct answer. For questions involving temporal evolution, at least one distractor is often selected from an earlier preference state, making it necessary to identify the currently valid preference while accounting for historically valid alternatives. For anti-stereotypical preferences, at least one distractor is designed to be stereotype-consistent, creating a deliberate trap for models that rely on demographic priors without grounding their answers in behavioral evidence.

\subsubsection{Benchmark Assembly and Metadata}
\label{app:persona_metadata}
Finally, each validated OOD question $q_{ij}^{\mathrm{ood}}$ is paired with the corresponding long-context history $H_i$ to form a benchmark instance
\[
b_{ij}^{\mathrm{ood}}
=
\left(
H_i,
x_{ij}^{q},
a_{ij}^{+},
\mathcal{A}_{ij}^{-},
t_{ij},
d_{ij},
m_{ij}
\right),
\]
where $m_{ij}$ denotes metadata associated with the instance. The metadata includes the source persona, source preference, topic label, scenario type, whether the preference belongs to the user or another person, whether the preference is updated, the associated conversation snippet, and, when applicable, the change family, trajectory length, trajectory step, full change plan, OOD type, and complexity level.

By construction, PersonaMem-Evo isolates a setting in which success depends on remembering, disambiguating, and generalizing user-specific behavioral evidence, while filtering cases that can be solved through explicit fact lookup or demographic shortcuts.

\subsubsection{Interaction Formats and Preference Records}
\label{app:persona_inter_format}

For each generated preference, we synthesize one or more short dialogue segments between a simulated user and an AI assistant. Each segment follows a predefined interaction format, such as email revision, translation, knowledge-seeking, recommendation, troubleshooting, or consultation-style advice. The format controls the surface form of the exchange, while the target preference remains implicit: it is expressed through the user's constraints, wording, examples, repeated requests, or surrounding context, without using a direct statement such as ``I like X.''

Each segment is paired with a structured \emph{preference record}, which specifies the latent preference expressed by the dialogue and its relation to other records. A record contains the preference text, source category, owner, temporal status, and the generated conversation. The source category indicates where the preference originates in the generation pipeline, e.g., stereotypical, anti-stereotypical, neutral, health-related, therapy-related, multimodal, or OOD-inference. The owner field indicates whether the preference describes the simulated user (\texttt{self}) or another person mentioned by the user (\texttt{others}). The temporal fields specify whether the preference is static, updates a previous preference, or corresponds to an explicit forgetting request.

Concretely, each record contains fields such as
\[
\texttt{preference},\quad
\texttt{pref\_type},\quad
\texttt{who},\quad
\texttt{updated},\quad
\texttt{prev\_pref},\quad
\texttt{conversations}.
\]
For preference-evolution cases, we additionally store trajectory-level metadata:
\[
\texttt{change\_family},\quad
\texttt{change\_k},\quad
\texttt{change\_step},\quad
\texttt{change\_plan}.
\]
Here, \texttt{conversations} contains the actual dialogue turns, while the remaining fields specify the latent preference and its temporal relation to other preference records. During benchmark assembly, these dialogue segments are concatenated into mixed-topic long-context histories while preserving their boundaries and metadata for question generation and analysis.

\subsubsection{Dataset Statistics}
\label{app:persona_dataset_statistics}

We report diagnostic statistics for the 10-persona \textsc{PersonaMem-Evo} evaluation subset used in our analysis. This subset contains 505 OOD preference-inference questions from 10 personas, with 49--51 questions per persona. The question types are approximately balanced across single-pattern transfer, multi-pattern synthesis, conflict resolution, and temporal trajectory prediction. The subset also includes a broad difficulty distribution, with 120 L1, 186 L2, and 199 L3 questions.

\begin{table}[!htbp]
\centering
\small
\begin{tabular}{lrrr}
\toprule
\textbf{Statistic} & \textbf{Count} & \textbf{Mean / Median} & \textbf{Range} \\
\midrule
Personas & 10 & -- & -- \\
Questions & 505 & 50.5 / 51 per persona & 49--51 \\
\midrule
Single-pattern transfer & 130 & -- & -- \\
Multi-pattern synthesis & 129 & -- & -- \\
Conflict resolution & 117 & -- & -- \\
Temporal trajectory prediction & 129 & -- & -- \\
\midrule
L1 questions & 120 & -- & -- \\
L2 questions & 186 & -- & -- \\
L3 questions & 199 & -- & -- \\
\bottomrule
\end{tabular}
\caption{Scale, question-type, and difficulty statistics for the 10-persona \textsc{PersonaMem-Evo} subset.}
\label{tab:personamem_evo_10p_scale}
\end{table}

We further characterize how much preference evidence each question requires. A \emph{source preference} is a gold evidence preference listed in \texttt{source\_preferences}; it corresponds to an observed user preference that should be used to answer the question. This is a mention-level count, so a single question may require multiple source-preference mentions. In total, the 505 questions contain 1,553 source-preference mentions. As shown in Table~\ref{tab:personamem_evo_source_count}, single-pattern transfer questions require one source preference, conflict-resolution questions require two, and multi-pattern synthesis questions require three. Temporal trajectory questions are the most evidence-intensive, requiring 2--10 source preferences with median 6, making them a strong test of temporal memory and multi-evidence composition.

\begin{table}[!htbp]
\centering
\small
\begin{tabular}{lrrrrr}
\toprule
\textbf{Question Type} & \textbf{\#Questions} & \textbf{Min} & \textbf{Median} & \textbf{Mean} & \textbf{Max} \\
\midrule
Single-pattern transfer & 130 & 1 & 1 & 1.0 & 1 \\
Conflict resolution & 117 & 2 & 2 & 2.0 & 2 \\
Multi-pattern synthesis & 129 & 3 & 3 & 3.0 & 3 \\
Temporal trajectory prediction & 129 & 2 & 6 & 6.2 & 10 \\
\bottomrule
\end{tabular}
\caption{Number of source preferences required per question.}
\label{tab:personamem_evo_source_count}
\end{table}

The source-preference composition is designed to support fine-grained diagnosis of memory-agent failures. Among the 1,553 source-preference mentions, the subset contains comparable numbers of stereotypical and anti-stereotypical preferences, which helps test whether agents rely on observed user behavior instead of persona-level priors. We also distinguish static preferences from changed preferences. Among source preferences matched to raw metadata, 801 mentions correspond to changed preferences and 664 correspond to static preferences. Changed preferences are further divided into object replacement, same-object attitude revision, conditional preference shift, temporal-validity shift, and attribute shift, allowing us to analyze which kinds of preference updates are most challenging.

\begin{table}[!htbp]
\centering
\small
\begin{minipage}[t]{0.48\linewidth}
\centering
\begin{tabular}{lr}
\toprule
\textbf{Source Category} & \textbf{Mentions} \\
\midrule
Stereotypical preference & 574 \\
Anti-stereotypical preference & 564 \\
Neutral preference & 223 \\
Health / medical preference & 35 \\
Therapy-background preference & 1 \\
Other matched source mention & 68 \\
Unmatched exact string & 88 \\
\midrule
Total source mentions & 1,553 \\
Matched to raw metadata & 1,465 \\
Matched to observed dialogue & 1,397 \\
\bottomrule
\end{tabular}
\vspace{1mm}

\textbf{(a) Source-preference composition.}
\end{minipage}
\hfill
\begin{minipage}[t]{0.42\linewidth}
\centering
\begin{tabular}{lr}
\toprule
\textbf{Preference Type} & \textbf{Mentions} \\
\midrule
Static preference & 664 \\
\midrule
Object replacement & 273 \\
Same-object attitude revision & 177 \\
Conditional preference shift & 179 \\
Temporal-validity shift & 111 \\
Attribute shift & 61 \\
\bottomrule
\end{tabular}
\vspace{1mm}

\textbf{(b) Static and changed preferences.}
\end{minipage}
\caption{Source-preference composition and change-family statistics. Counts are mention-level.}
\label{tab:personamem_evo_source_composition}
\end{table}

Finally, we characterize the length of the generated interaction histories. Each benchmark question is answered against the full persona-level chat history instead of a short declarative profile. As shown in Table~\ref{tab:personamem_evo_context_length}, histories are long: the median persona has 597 messages and 174.7K tokens. This makes \textsc{PersonaMem-Evo} a challenging setting for memory agents, since relevant implicit preference evidence must be extracted from long, mixed-topic conversations.

\begin{table}[!htbp]
\centering
\small
\begin{tabular}{lrrrr}
\toprule
\textbf{Chat-History Statistic} & \textbf{Min} & \textbf{Median} & \textbf{Mean} & \textbf{Max} \\
\midrule
Messages per persona & 113 & 597.0 & 634.7 & 1,077 \\
User messages per persona & 57 & 308.0 & 325.6 & 551 \\
Assistant messages per persona & 55 & 288.5 & 308.1 & 525 \\
Words per persona & 25,177 & 134,552 & 136,110 & 267,777 \\
Tokens per persona & 31,719 & 174,684 & 178,292 & 367,935 \\
\bottomrule
\end{tabular}
\caption{Generated chat-history length statistics for the 10-persona subset.}
\label{tab:personamem_evo_context_length}
\end{table}

Together, these statistics show that \textsc{PersonaMem-Evo} tests more than factual recall. Source-preference counts measure evidence aggregation, source composition tests resistance to stereotype shortcuts, change-family statistics test preference updating, and long histories impose a realistic memory-retrieval burden.

\section{EvoMem Implementation Details}
\label{appendix:evomem_implement}

\subsection{Terminus2 with EvoMem}
\label{sec:evomem_terminus2}

\textbf{Overview.}
We implement EvoMem for Terminus2 as a chain-scoped patch memory layer for terminal-native task solving. This chain memory instantiates $M_T$ for Terminus2: it is synthesized from prior terminal trajectories and supplied as an external memory context. Terminus2 keeps its original terminal interaction loop: it observes the task, issues shell actions, reads terminal feedback, and decides when to finish. EvoMem augments the context that Terminus2 receives before a task and updates a persistent chain memory after a task has produced a terminal trajectory.

The implementation is designed for sequential task chains. Earlier tasks in the same chain provide internal evidence for synthesizing reusable procedural memory, while later tasks often introduce small but consequential changes in input location, output contract, environment, tool availability, or artifact format. EvoMem records these shifts as compact transition patches that pair a strategy-level adaptation with the task or environment condition that caused it. Importantly, EvoMem does not expose raw previous trajectories, verifier outcomes, reference patches, or task-specific solution artifacts to later tasks. Prior executions are used only to synthesize compact, sanitized procedural summaries and transition-level adaptation hints. At the next task, EvoMem retrieves only the chain memory that is applicable to the current instruction and environment, so the agent receives targeted guidance about how the solution strategy should adapt.

\paragraph{Chain-scoped terminal memory.}
Let
\[
\mathcal{X}=\{x_1,x_2,\ldots,x_T\}
\]
denote a sequence of related terminal tasks in one chain. After processing the first $t$ tasks, EvoMem maintains two chain-level stores:
\[
\mathcal{L}_t=\{\ell_1,\ell_2,\ldots,\ell_t\},
\qquad
\Pi_t=\{\pi_2,\pi_3,\ldots,\pi_t\}.
\]
Here $\ell_i$ is a compact execution ledger for task $x_i$, and $\pi_i$ is a transition patch that summarizes how the strategy changed from $x_{i-1}$ to $x_i$. The ledger records abstract reusable procedure information from a prior terminal solution: the task goal, a short strategy summary, relevant artifact types, high-level command patterns, ineffective strategy notes, and output requirements. The transition patch records the non-additive change across neighboring tasks: what requirement or environment condition changed, which previous assumption was superseded, what adaptation was observed, and which concrete values from the previous task should not be copied blindly.

This representation separates a base recipe from revision evidence. A base ledger gives the agent an initial solving template for the chain, while patches explain how that template should be revised when the current task differs from previous ones. This is especially important for evolving terminal tasks, where the high-level workflow may remain stable but a path, input convention, output format, or toolchain constraint changes.

\paragraph{Store-time update.}
After Terminus2 finishes a task $x_t$, EvoMem extracts a structured terminal history $h_t$ from the agent trajectory. This history contains the agent's messages, shell commands, and terminal observations in a normalized event format. EvoMem also collects a lightweight environment snapshot $e_t$, including available command-line tools and a compact sketch of the task workspace. The environment snapshot is best-effort and is used only to characterize applicability; failure to obtain it does not change the underlying Terminus2 loop.

For $t>1$, EvoMem compares the current instruction with the previous task instruction:
\[
\delta_t=\mathrm{Diff}(x_{t-1},x_t).
\]
The update summarizer then receives the current instruction, the previous instruction, the instruction delta, the terminal history, the environment snapshot, and bounded evidence about changed artifacts. It produces a ledger and, when a previous task exists, a transition patch:
\[
(\ell_t,\pi_t)=U(x_t,x_{t-1},\delta_t,h_t,e_t).
\]
The ledger $\ell_t$ captures abstract reusable execution knowledge from the current task, while removing task-specific answer artifacts. The patch $\pi_t$ captures how the current task revises the previous chain behavior and why that revision was needed under the new task or environment conditions. If the current task is the first task in a chain, EvoMem writes only the ledger and treats it as the initial chain recipe.

\paragraph{Patch representation.}
Each transition patch is represented as
\[
\pi_t =
\left(
c_t,
a_t,
o_t,
g_t,
d_t,
r_t
\right),
\]
where $c_t$ denotes the type of task or environment change, $a_t$ is the superseded assumption, $o_t$ is the observed adaptation, $g_t$ is a generalized pattern for future tasks, $d_t$ lists concrete prior details that should not be copied unless the current instruction requires them, and $r_t$ records the concise cause of the revision. The implementation infers $c_t$ from instruction differences, environment signals, and execution evidence, without using task-specific hand labels. Examples include changes in input source, output destination, formatting contract, working directory convention, data layout, and toolchain or environment requirements.

The patch is intentionally written as a transition example for adaptation across related tasks. This distinction matters because Terminus2 operates in a live terminal environment: a previous strategy may be useful as evidence, but the current instruction remains authoritative. EvoMem therefore preserves only abstract command patterns when they are useful as reusable procedural guidance, and pairs them with patch fields that describe which parts must be re-derived from the current task.

\paragraph{Retrieval-time construction.}
Before solving a new task $x_{t+1}$, EvoMem first checks whether the chain contains reusable memory. If no ledger or relevant patch exists, the original Terminus2 instruction is passed through unchanged. This avoids prompting the agent to search for memory when no useful memory has been created.

When chain memory exists, EvoMem builds a retrieval query from the current task instruction, the change summary against the previous task, and the current environment snapshot:
\[
q_{t+1}=Q(x_{t+1},\delta_{t+1},e_{t+1}).
\]
The retriever selects a base ledger and a small set of relevant transition patches:
\[
C_{t+1}=\mathrm{Retrieve}(q_{t+1},\mathcal{L}_t,\Pi_t).
\]
Patch matching combines two signals. First, EvoMem compares the current task-contract change with the change types summarized in previous patches, so that output-format changes retrieve relevant output-format adaptations. Second, it compares task text, environment signals, and salient paths or artifacts at a lexical level. Retrieved patches are then ordered according to their position in the task chain, preserving the temporal structure of the evolving task sequence.

\paragraph{Prompt construction and container hydration.}
Terminus2 can be sensitive to long prompt prefixes and transcript-like memory. We therefore do not paste the full chain memory directly into the main task instruction. Instead, EvoMem renders the selected context into a compact memory document and materializes it inside the task container before Terminus2 starts solving. The prompt receives only a short memory reference telling the agent that chain memory is available and that the current task instruction is authoritative:
\[
P(x_{t+1})=
\mathrm{Prompt}\!\left(x_{t+1},\mathrm{Ref}(C_{t+1})\right).
\]
If container hydration is unavailable, EvoMem falls back to an inline compact rendering under the same length budget.

The rendered memory contains no raw terminal transcript or previous task answer. It contains three sanitized components. The first summarizes current differences from the previous task. The second provides a base execution ledger, including abstract recipe steps or high-level command patterns when available. The third lists relevant transition examples, each phrased as a conditional adaptation pattern. This format gives Terminus2 enough information to adapt prior experience while avoiding a large, noisy replay of earlier trajectories.

\paragraph{Safety and anti-copying safeguards.}
EvoMem includes safeguards to prevent stale memory from overriding the current task. First, the prompt explicitly states that retrieved memory is guidance and that the current instruction is authoritative. Second, transition patches distinguish generalized adaptation patterns from concrete old values. Prior output fragments, answer literals, and task-specific prefixes are redacted or abstracted before they are exposed to the agent. Third, patches include ``do-not-copy'' evidence, which marks old paths, values, or output fragments as examples that should be reused only if the current instruction independently calls for them.

These safeguards are important for terminal tasks because a memory item can be partially correct but obsolete in a later environment. EvoMem is therefore not a replay buffer. It is a patch-conditioned context mechanism: the agent is expected to use prior execution as evidence, infer the current task's own contract, and adapt the solution accordingly.

\begin{algorithm}[H]
\caption{EvoMem: Chain-Scoped Patch Memory for Terminus2}
\label{alg:evomem_terminus2}
\begin{algorithmic}[1]
\Require Task chain $\mathcal{X}=\{x_1,\ldots,x_T\}$; Terminus2 solver $M$; chain ledgers $\mathcal{L}$; transition patches $\Pi$
\Ensure Completed terminal trajectories and updated chain memory

\State Initialize $\mathcal{L}\gets\emptyset$, $\Pi\gets\emptyset$
\For{$t=1$ to $T$}
    \State Obtain environment snapshot $e_t$
    \If{$\mathcal{L}$ or $\Pi$ contains reusable chain memory}
        \State Compute instruction delta $\delta_t$ against the previous chain task
        \State Build query $q_t\leftarrow Q(x_t,\delta_t,e_t)$
        \State Retrieve base ledger and transition patches $C_t\leftarrow \mathrm{Retrieve}(q_t,\mathcal{L},\Pi)$
        \State Materialize compact sanitized memory context for the task container
        \State Compose prompt $P_t\leftarrow \mathrm{Prompt}(x_t,\mathrm{Ref}(C_t))$
    \Else
        \State Set $P_t\leftarrow \mathrm{Prompt}(x_t)$
    \EndIf
    \State Run Terminus2: $(y_t,h_t)\leftarrow M(P_t)$
    \State Extract normalized terminal history $h_t$
    \State Synthesize sanitized ledger $\ell_t\leftarrow U_{\mathrm{ledger}}(x_t,h_t,e_t)$
    \State $\mathcal{L}\leftarrow\mathcal{L}\cup\{\ell_t\}$
    \If{$t>1$}
        \State Compute transition patch $\pi_t\leftarrow U_{\mathrm{patch}}(x_{t-1},x_t,\delta_t,h_t,e_t)$
        \State $\Pi\leftarrow\Pi\cup\{\pi_t\}$
    \EndIf
\EndFor
\end{algorithmic}
\end{algorithm}

\paragraph{Relation to Terminus2.}
The integration is intentionally non-invasive. Terminus2 remains responsible for planning, terminal command generation, observation handling, and completion decisions. EvoMem adds only a persistent chain memory store, a retrieval-and-rendering step before execution, and a post-run summarization step after the terminal trajectory is available:
\[
\text{Terminus2+EvoMem}
=
\text{Terminus2}
+
\text{Chain Memory}
+
\text{Transition Patch Retrieval}.
\]
This preserves Terminus2's terminal-native behavior while allowing it to reuse abstract procedural experience across evolving tasks without exposing raw prior task executions or blindly repeating procedures that were valid only for earlier chain members.

\subsection{OpenHands with EvoMem}
\label{sec:evomem_openhands}

\textbf{Overview.}
We implement EvoMem for OpenHands as an external patch-memory layer for sequential software-engineering tasks. OpenHands preserves its original autonomous coding loop: it reads the task, inspects the repository, edits code, runs commands, and produces a patch. EvoMem does not modify the OpenHands agent internals. Instead, it wraps the task interface: before each task, it retrieves compact historical code context and prepends it to the task description; after the task, it distills the resulting trajectory and patch into feature-level memory records.

For OpenHands, $M_T$ is instantiated as a distilled code-context memory:
\[
M_T=\{\rho_1,\rho_2,\ldots,\rho_T\},
\]
where each record $\rho_i$ summarizes a prior implementation change. The purpose is not to replay old patches verbatim, but to provide historical debugging context: which files and symbols were involved, what behavior changed, why the change was needed, which constraints should be preserved, and what code evidence supports the summary. When a later task supersedes an earlier implementation assumption, the corresponding record captures the old behavior, the revised logic, and the concrete behavioral cause of the revision.

\paragraph{Feature-level patch records.}
Given a task $x_t$, OpenHands produces a trajectory $h_t$ and a code patch $d_t$. EvoMem first groups the patch into feature-level units using changed files, code hunks, and execution evidence around file reads, code edits, tests, and command observations. For each feature group $g$, a summarizer constructs a patch record
\[
\rho =
\left(
f,\; s,\; b^{-},\; b^{+},\; c,\; r,\; \Delta
\right),
\]
where $f$ denotes affected files, $s$ denotes relevant symbols or feature tags, $b^{-}$ and $b^{+}$ summarize behavior before and after the change, $c$ lists constraints to preserve, $r$ gives the concrete reason for the revision, and $\Delta$ stores bounded code evidence. The record is intentionally compact: it preserves the information needed for future debugging while avoiding long trajectory replay.

This representation matches the software-engineering setting. A prior task may have modified a helper, parser, API boundary, or configuration path in a way that remains relevant to later tasks. If a later instruction touches overlapping code, the retrieved record reminds the agent of the previous behavioral intent and reduces the risk of undoing earlier logic.

\paragraph{Retrieval.}
At inference time, EvoMem builds a query from the current task description. The query includes visible file hints, natural-language feature terms, and symbol-like identifiers:
\[
q_t=Q(x_t).
\]
For each prior record $\rho_i$, EvoMem estimates relevance using both semantic and structural evidence. The semantic component matches the task query against the record's feature title, tags, symbols, behavioral summary, and code-context text. The structural component compares file-path segments and exact file overlap:
\[
s(q_t,\rho_i)
=
\alpha\,s_{\mathrm{text}}(q_t,\rho_i)
+
\beta\,s_{\mathrm{path}}(q_t,\rho_i)
+
\gamma\,s_{\mathrm{exact}}(q_t,\rho_i).
\]
The top-ranked records are rendered under a context budget. If no useful prior memory exists, EvoMem leaves the OpenHands task prompt unchanged.

\paragraph{Prompt construction.}
The selected records are rendered as a short patch-memory block and prepended to the original task description:
\[
P(x_t)=\mathrm{Prompt}\!\left(x_t,\{\rho_{i_1},\ldots,\rho_{i_k}\}\right).
\]
Each rendered record contains the affected files, the revision intent, constraints that should be preserved, and a bounded code excerpt when available. The prompt explicitly states that the current task remains authoritative and that prior patches should be used only as contextual evidence. Thus, EvoMem supplies historical debugging memory without changing the OpenHands planner, editor, tool interface, or repository interaction loop.

\begin{algorithm}[H]
\caption{EvoMem: Patch Memory for OpenHands}
\label{alg:evomem_openhands}
\begin{algorithmic}[1]
\Require Sequential tasks $\mathcal{X}=\{x_1,\ldots,x_T\}$; OpenHands solver $\mathcal{A}$; memory $M_0=\emptyset$
\Ensure Updated patch memory $M_T$

\For{$t=1$ to $T$}
    \State Set $M_t\leftarrow M_{t-1}$
    \State Build query $q_t\leftarrow Q(x_t)$ from visible files, feature terms, and symbols
    \State Retrieve relevant records $R_t\leftarrow \mathrm{Retrieve}(q_t,M_{t-1})$
    \If{$R_t$ is nonempty}
        \State Compose prompt $P_t\leftarrow \mathrm{Prompt}(x_t,R_t)$
    \Else
        \State Set $P_t\leftarrow \mathrm{Prompt}(x_t)$
    \EndIf
    \State Run OpenHands: $(d_t,h_t)\leftarrow \mathcal{A}(P_t)$
    \State Segment $d_t$ into feature-level groups $\mathcal{G}_t$
    \For{each group $g\in\mathcal{G}_t$}
        \State Summarize $(g,d_t,h_t)$ into patch record $\rho_g$
        \State $M_t\leftarrow M_t\cup\{\rho_g\}$
    \EndFor
\EndFor
\end{algorithmic}
\end{algorithm}

\paragraph{Relation to OpenHands.}
EvoMem is a lightweight augmentation to OpenHands:
\[
\text{OpenHands+EvoMem}
=
\text{OpenHands}
+
\text{Patch Record Store}
+
\text{Semantic-Structural Retrieval}.
\]
The base agent remains responsible for repository exploration, code editing, command execution, and patch generation. EvoMem only determines which historical code-context records should be shown before the next task and how the newly produced patch should be summarized for future reuse.

\subsection{Memento-Skill with EvoMem}
\label{sec:evomem_memento}

% \subsection{EvoMem: Versioned Tip Patching on Top of Memento-S}
% \label{sec:evomem_memento}

\textbf{Overview.}
We implement EvoMem as a lightweight memory layer on top of the existing Memento-Skill task-solving loop. The key design choice is to keep the base skill router and skill execution API unchanged, while augmenting only the learning-tip subsystem. Concretely, Memento-Skill continues to select and execute skills in its original workflow. EvoMem intervenes at two points: before solving a task, it retrieves historically relevant tip versions and injects them into the prompt; after a task failure yields reusable feedback, it writes a new versioned tip patch into memory. This preserves the original Memento-Skill control flow while enabling continual accumulation and selective reuse of task-solving experience.

\paragraph{Versioned tip memory.}
In Memento-S, reusable experience is stored in a global \texttt{TIP.md} file. EvoMem replaces this single mutable file with a family of versioned tip memories. Let
\[
\mathcal{F}_T=\{v_1,v_2,\ldots,v_T\}
\]
denote the tip-version family after $T$ updates. Each version is represented as
\[
v_t=(\tau_t,m_t,p_t),
\]
where $\tau_t$ is the full tip-text snapshot, $m_t$ is retrieval-oriented metadata, and $p_t$ is a parent pointer indicating lineage. In our implementation, the family is stored under the global identifier \texttt{global\_learning\_tips}; each version is serialized as a JSON file under \texttt{.memgit/families/tips/global\_learning\_tips/versions/}.

The metadata $m_t$ records the evidence and rationale behind a version:
\[
m_t =
\left(
r_t,\Delta_t,S_t^{+},S_t^{-},\Gamma_t^{+},\Gamma_t^{-},\Sigma_t
\right).
\]
Here, $r_t$ denotes the update reasoning, $\Delta_t$ summarizes the difference from the previous version, $S_t^{+}$ and $S_t^{-}$ denote compressed success and failure case summaries, $\Gamma_t^{+}$ and $\Gamma_t^{-}$ summarize successful and failed trajectories, and $\Sigma_t$ contains selection signals used as retrieval cues. Although we use the term ``patch'', the implementation stores full replacement snapshots rather than textual diffs. The patch semantics are captured by the lineage pointer $p_t$ and the difference metadata $\Delta_t$, while inference can directly read the full snapshot $\tau_t$ without replaying an edit chain.

\paragraph{Store-time update.}
EvoMem writes a new tip version only when Memento-Skill observes a reusable learning signal, typically after an incorrect attempt followed by judge feedback and execution-trace analysis. Let $x$ denote the task, $y$ the system output, and $j$ the failure analysis produced from the judge and trace. Given the previous tip version $v_T=(\tau_T,m_T,p_T)$, EvoMem uses an LLM-based updater to synthesize a new tip snapshot and metadata:
\[
(\tau_{T+1},m_{T+1}) = U(x,j,\tau_T,m_T),
\]
where $U(\cdot)$ is prompted to produce concise, task-specific guidance, organize it by task type, and generate retrieval-friendly selection signals.

The new version is written in two steps. First, EvoMem derives a child version from the current latest version:
\[
v_{T+1} \leftarrow \textsc{Derive}(v_T),
\]
which copies the parent lineage and initializes the child metadata. Second, EvoMem records the new tip snapshot and metadata:
\[
v_{T+1} \leftarrow \textsc{Record}(v_{T+1},\tau_{T+1},m_{T+1}).
\]
The family then evolves monotonically:
\[
\mathcal{F}_{T+1}=\mathcal{F}_T\cup\{v_{T+1}\}.
\]
This update rule maintains versioned tip histories, allowing different historical tips to remain available for different task regimes.

\paragraph{Inference-time retrieval.}
At inference time, EvoMem converts the incoming task into a retrieval query $q$ using stable task descriptors already available in Memento-S, such as the question text, answer type, category, subject, difficulty, and attachment names:
\[
q = Q(x).
\]
For each tip version $v_t$, EvoMem constructs a retrieval document
\[
d_t=\texttt{concat}(\tau_t,m_t),
\]
which includes the tip text, update reasoning, difference summary, success and failure summaries, trajectory summaries, and selection signals. The first-stage retrieval score is BM25:
\[
s_t^{\mathrm{bm25}}=\mathrm{BM25}(q,d_t).
\]
Versions are ranked within the tip family, and the top-$k$ candidates are selected. We use $k=2$ by default.

The implementation also supports optional hybrid reranking. Let $c_t$ be the TF-IDF cosine similarity between $q$ and $d_t$, and let $\hat{s}_t^{\mathrm{bm25}}$ denote the BM25 score normalized by the maximum score among retrieved candidates. The hybrid score is
\[
s_t =
\lambda \hat{s}_t^{\mathrm{bm25}} + (1-\lambda)c_t,
\]
where $\lambda\in[0,1]$ and the default value is $\lambda=0.5$. A relative threshold $\rho$ can further filter weak matches:
\[
\frac{s_t}{\max_j s_j}\ge \rho,
\]
with $\rho=0$ by default. Unless otherwise stated, we use the default BM25 retrieval setting.

\paragraph{Prompt construction and execution.}
The selected versions are rendered into a prompt block containing the version identifier, selection signals, a truncated tip snapshot, representative success and failure summaries, and the update note. The final Memento-Skill prompt is
\[
P(x)=\texttt{Prompt}\!\left(x,\{v_{t_1},\ldots,v_{t_k}\}\right),
\]
where the retrieved versions provide contextual guidance for the current task and do not override the task instruction. Memento-Skill then solves the task using its original solver:
\[
(y,\mathrm{trace}) = M(P(x)).
\]
If the task succeeds, the output is returned directly. If the task fails, EvoMem may create a new tip version using the store-time update rule above; when retry is enabled, the newly written version can be pinned into the next prompt.

\begin{algorithm}[H]
\caption{EvoMem: Versioned Tip Patching on Top of Memento-S}
\label{alg:evomem_memento}
\begin{algorithmic}[1]
\Require Task $x$; Memento-Skill solver $M$; tip family $\mathcal{F}_T=\{v_1,\ldots,v_T\}$; top-$k$; interpolation weight $\lambda$; threshold $\rho$
\Ensure Answer $y$

\State Build retrieval query $q \leftarrow Q(x)$
\For{each tip version $v_t \in \mathcal{F}_T$}
    \State Build retrieval document $d_t \leftarrow \texttt{concat}(\tau_t,m_t)$
    \State Compute BM25 score $s_t^{\mathrm{bm25}} \leftarrow \mathrm{BM25}(q,d_t)$
\EndFor
\State Select top-$k$ candidates by $s_t^{\mathrm{bm25}}$

\If{hybrid reranking is enabled}
    \For{each selected candidate $v_t$}
        \State Compute TF-IDF cosine score $c_t \leftarrow \mathrm{cos}(q,d_t)$
        \State Normalize BM25 score $\hat{s}_t^{\mathrm{bm25}}$
        \State $s_t \leftarrow \lambda \hat{s}_t^{\mathrm{bm25}} + (1-\lambda)c_t$
    \EndFor
    \State Re-rank candidates by $s_t$
\Else
    \State Set $s_t \leftarrow s_t^{\mathrm{bm25}}$
\EndIf

\State Retain candidates satisfying $s_t/\max_j s_j \ge \rho$
\State Compose prompt $P(x)$ by appending retained tip versions to the original Memento-Skill prompt
\State Run Memento-S: $(y,\mathrm{trace}) \leftarrow M(P(x))$

\If{$y$ is correct}
    \State \Return $y$
\Else
    \State Obtain failure analysis $j$ from judge feedback and execution trace
    \State $(\tau_{T+1},m_{T+1}) \leftarrow U(x,j,\tau_T,m_T)$
    \State $v_{T+1} \leftarrow \textsc{Derive}(v_T)$
    \State $v_{T+1} \leftarrow \textsc{Record}(v_{T+1},\tau_{T+1},m_{T+1})$
    \State $\mathcal{F}_{T+1} \leftarrow \mathcal{F}_T \cup \{v_{T+1}\}$
    \If{retry is enabled}
        \State Compose retry prompt with $v_{T+1}$ pinned
        \State $(y,\mathrm{trace}) \leftarrow M(P_{\mathrm{retry}}(x))$
    \EndIf
    \State \Return $y$
\EndIf
\end{algorithmic}
\end{algorithm}

\paragraph{Relation to Memento-S.}
EvoMem is intentionally non-invasive. It does not replace the Memento-Skill skill selector, modify the skill execution API, or alter the base solving loop. Instead, it modifies only the prompt construction stage and the post-failure feedback stage. This separates capability routing from experience reuse: the Memento-Skill skill router determines which tool or skill family to use, while EvoMem determines which historical reasoning patches are most relevant for the current task.

\subsection{A-Mem with EvoMem}
% \subsection{EvoMem: Patch-Augmented A-Mem}
\label{sec:evomem_amem}

\textbf{Overview.}
We implement EvoMem as a patch-augmented extension of A-Mem. A-Mem serves as the base memory system, maintaining a structured memory graph that stores the latest consolidated user information. EvoMem keeps this mechanism unchanged, but adds a patch layer that records non-additive memory revisions caused by later turns. This design targets a limitation of long-horizon personalization: after many updates, the current memory may preserve the latest belief but obscure how the belief changed, when it changed, and which earlier state was superseded. EvoMem therefore maintains both the current A-Mem graph and a temporally indexed history of memory edits.

Formally, let
\[
G_t=(V_t,E_t)
\]
denote the A-Mem memory graph after processing turn $x_t$. Each node $v\in V_t$ is a memory note containing textual content and structured attributes such as context, keywords, and tags; each edge $e\in E_t$ represents an association between notes. A-Mem updates this graph by inserting new information, revising existing notes, and updating note links. EvoMem augments this process with patch construction, patch indexing, and patch-conditioned retrieval. Algorithm~\ref{alg:evomem_amem} summarizes the full procedure.

\begin{algorithm}[H]
\caption{EvoMem: Patch-Augmented A-Mem}
\label{alg:evomem_amem}
\begin{algorithmic}[1]
\Require Conversation stream $\mathcal{X}=\{x_1,\ldots,x_T\}$; question $q$
\Ensure Answer $\hat{y}$

\State Initialize A-Mem graph $G_0 \gets \emptyset$
\State Initialize patch memory $\Pi \gets \emptyset$

\For{$t=1$ to $T$}
    \State $G_{t-1}^{\mathrm{old}} \gets G_{t-1}$
    \State $(G_t,\tau_t) \gets \textsc{AMemUpdate}(G_{t-1}, x_t)$
    \State $\Delta_t \gets \textsc{GraphDiff}(G_{t-1}^{\mathrm{old}}, G_t)$
    \If{$\Delta_t$ is non-additive and $x_t$ is user-driven}
        \State $\pi_t \gets \textsc{SummarizePatch}(x_t,\tau_t,\Delta_t)$
        \State $\Pi \gets \Pi \cup \{\pi_t\}$
        \State $\textsc{IndexPatch}(\pi_t)$
    \EndIf
\EndFor

\State $z(q) \gets \textsc{KeywordRewrite}(q)$
\State $C^{\mathrm{cur}} \gets \textsc{RetrieveCurrent}(G_T,z(q),k)$
\State $P(q) \gets \textsc{RetrievePatches}(\Pi,z(q),m)$
\State $P(q) \gets \textsc{SortByTemporalOrder}(P(q))$
\State $C^{\mathrm{patch}} \gets \textsc{FormatPatches}(P(q))$
\State $C(q) \gets [C^{\mathrm{cur}}; C^{\mathrm{patch}}]$
\State $\hat{y} \gets f_\theta(q,C(q))$
\State \Return $\hat{y}$
\end{algorithmic}
\end{algorithm}

\paragraph{Patch construction during memory ingestion.}
For each incoming turn $x_t$, EvoMem first applies the original A-Mem update routine:
\[
(G_t,\tau_t)=\mathcal{M}(G_{t-1},x_t),
\]
where $\mathcal{M}$ denotes the A-Mem insertion procedure and $\tau_t$ is the evolution trace produced during memory editing. EvoMem then compares the graph before and after the update:
\[
\Delta_t=\mathrm{Diff}(G_{t-1},G_t).
\]
In our implementation, $\Delta_t$ tracks newly created notes, updated notes, changed fields, and modified links. Changed fields include note content, context, keywords, tags, and graph relations.

EvoMem creates a patch only for non-additive updates. Purely additive updates, where a new note is added without modifying existing memory, are already preserved in the current graph and do not require a patch. We therefore create patches only when the update changes existing memory semantics or graph structure. In practice, this includes two main cases: \texttt{overwrite\_update}, where note content or metadata is revised, and \texttt{link\_rewrite\_update}, where the neighborhood of an existing note is rewired. We also suppress patches triggered by assistant turns, so that patches primarily reflect user-driven memory evolution and avoid capturing assistant-side paraphrases or generated explanations.

\paragraph{Patch representation.}
When a non-additive update is detected, EvoMem materializes a patch record
\[
\pi_t =
\left(
\mathrm{id}_t,
\mathrm{meta}_t,
\Delta_t,
S_t,
R_t
\right),
\]
where $\mathrm{id}_t$ is a unique patch identifier, $\mathrm{meta}_t$ stores turn-level metadata, $\Delta_t$ records the structured graph edit, $S_t$ is an LLM-generated summary of the update, and $R_t$ contains before/after evidence for affected notes. The metadata includes session index, turn index, timestamp, and speaker role.

More concretely, each patch stores the trigger turn, patch type, affected notes, note-level before/after states, edge rewrites, an update summary, an estimated preference domain, an inferred change type, and a temporal order key. For each changed note $v$, EvoMem records a local edit block
\[
r_t(v)=\left(v,\mathcal{F}_t(v),\mathbf{b}_t(v),\mathbf{a}_t(v)\right),
\]
where $\mathcal{F}_t(v)$ is the set of changed fields, and $\mathbf{b}_t(v)$ and $\mathbf{a}_t(v)$ denote the note state before and after the update. This exposes what was revised instead of retaining only the final consolidated note.

\paragraph{Patch indexing and storage.}
EvoMem stores the patch history separately from the A-Mem graph. For a sample $i$, let
\[
\Pi_i=\{\pi_1,\pi_2,\ldots,\pi_{T_i}\}
\]
denote its patch set. Each patch is converted into a compact retrieval document containing the temporal position, inferred preference domain, previous state, updated state, change type, and a short trigger snippet. We embed these patch documents with the same sentence encoder used by the retriever; in our implementation, we use \texttt{all-MiniLM-L6-v2}. The embedded patches are inserted into a separate patch retriever.

Thus, EvoMem maintains two retrieval spaces:
\[
\mathcal{R}_{\mathrm{mem}}: q \mapsto \text{current memory evidence from } G_T,
\]
\[
\mathcal{R}_{\mathrm{patch}}: q \mapsto \text{historical revision evidence from } \Pi_i.
\]
The base graph is optimized for retrieving the latest consolidated memory, while the patch store is optimized for retrieving historical changes that explain preference evolution, conflicts, or overwritten states.

\paragraph{Query-time retrieval.}
At inference time, EvoMem combines current graph evidence with relevant patches. Given a question $q$, we first rewrite it into a retrieval query
\[
z(q)=\mathrm{LLMKeyword}(q),
\]
using a lightweight keyword-generation prompt. We then retrieve current memory evidence from A-Mem:
\[
C^{\mathrm{cur}}(q)=\mathcal{R}_{\mathrm{mem}}(z(q),k),
\]
where $k$ is the number of retrieved notes. In our default evaluation configuration, we set $k=10$.

In parallel, EvoMem retrieves the top-$m$ relevant patches:
\[
P(q)=\mathcal{R}_{\mathrm{patch}}(z(q),m),
\]
with default $m=2$. By default, patch retrieval is embedding-based. The implementation also supports an optional hybrid scorer that interpolates dense similarity and BM25:
\[
s_{\mathrm{hyb}}(\pi,q)
=
\alpha s_{\mathrm{dense}}(\pi,q)
+
(1-\alpha)s_{\mathrm{BM25}}(\pi,q),
\]
where $\alpha=0.7$ in the hybrid setting. Unless otherwise stated, we use the standard embedding-based patch retrieval and disable hybrid retrieval, patch-node reranking, and gating.

After retrieval, selected patches are sorted by temporal order. This enforces a simple precedence rule: when multiple patches refer to the same preference domain, later patches should be considered more recent evidence than earlier ones.

\paragraph{Patch-conditioned answer generation.}
EvoMem renders the selected patches into a textual evidence block and concatenates them with current memory evidence:
\[
C(q)=\left[C^{\mathrm{cur}}(q); C^{\mathrm{patch}}(q)\right].
\]
Each patch block includes the session and turn indices, patch type, temporal order, trigger utterance, update summary, inferred change pattern, and key before/after evidence. The final answer is generated by conditioning the answering model on the question and the augmented context:
\[
\hat{y}=f_\theta(q,C(q)).
\]

In our evaluation harness, EvoMem uses an always-on patch mode: whenever relevant patches are retrieved, they are injected into the answer context. The implementation also supports a gated variant in which the model first decides whether patch details are necessary, but we do not enable this option in the default experiments.

\paragraph{Relation to A-Mem.}
EvoMem is a lightweight augmentation of A-Mem. The original A-Mem system still performs note creation, note evolution, link updates, and graph-based retrieval. EvoMem adds three components:
\[
\text{EvoMem}
=
\text{A-Mem}
+
\text{Diff Module}
+
\text{Patch Store}
+
\text{Patch Retrieval}.
\]
This preserves A-Mem's ability to organize consolidated memory while adding an explicit representation of memory evolution. Consequently, EvoMem can answer questions that depend not only on what the user currently prefers, but also on how a preference changed, which earlier state was replaced, and which revision should be treated as the latest valid belief.

\section{Experiments Setting}
\label{appendix:exp_setting}

In this section, we provide benchmark-specific implementation details. Since each benchmark requires different interaction protocols and agent capabilities, we organize the settings by dataset.

\subsection{Terminal-Bench-Evo}

For \emph{Terminal-Bench-Evo}, we use Terminus2 as the baseline terminal agent. The baseline does not use persistent memory. For Terminus2 + EvoMem, we keep the same Terminus2 configuration and add chain-scoped patch memory. Tasks within the same evolution chain are run sequentially so that memory from earlier variants is available to later variants; different chains are independent and may be run in parallel. The rendered memory context is kept compact with a budget of 3500 characters. Both baseline and EvoMem use the same model configuration and agent execution budget.

\subsection{SWE-Chain-Evo}
For \emph{SWE-Chain-Evo}, we use OpenHands with the CodeActAgent scaffold as the baseline software-engineering agent. The baseline disables patch memory. For OpenHands + EvoMem, we keep the same OpenHands inference setup with maximum iterations set to 100 and one run per milestone, and prepend retrieved patch-memory context to the task description. Milestones within a chain are processed sequentially, while baseline and EvoMem use the same task order and model configuration. The patch-memory context budget is set to 2200 characters, and patch summarization uses temperature $0.0$.

\subsection{PersonaMem-Evo}

\label{app:setting_persona_mem_evo}

For \emph{PersonaMem-Evo}, we use A-Mem \cite{xu2025amem} as the baseline memory agent with memory retrieval top-$k=12$. The A-Mem baseline does not use patch retrieval. For \emph{A-Mem + EvoMem}, we keep the same memory retrieval budget, set patch retrieval top-$k=3$, and use a patch similarity threshold of $0.4$. For answer generation, we use temperature $0.0$ to reduce sampling variance in multiple-choice prediction. The memory-building temperature is set to $0.7$ for both the baseline and EvoMem settings.

\subsection{GAIA}
\label{app:setting_gaia}

For \emph{GAIA}, we use \emph{Memento-Skill} \cite{memento-s} as the base agent. We evaluate on the repository split \texttt{gaia\_data/data/split\_by\_level\_60\_40/train}, which contains 100 GAIA instances balanced across difficulty levels. We use this split because it provides a larger and more balanced evaluation set than the held-out test split. Although the original Memento-Skill workflow uses this training split to learn skills and evaluates them on the test split, our goal is different: we compare the same Memento-Skill agent with and without EvoMem on the same set of tasks. Thus, the experiment isolates whether EvoMem improves reuse of task-solving experience, while keeping the task set fixed across compared agents.

Both the baseline and EvoMem settings use the same \texttt{read-write-optimize} protocol, with one feedback round (\texttt{--optimize-attempts 1}) and the post-update unit-test gate enabled. Final answer correctness is judged using \texttt{gpt-5.4-mini}. The baseline uses \texttt{--learning-tips-mode baseline}, which injects only the persistent global \texttt{TIP.md} file and does not retrieve task-specific patch memories. EvoMem uses the hybrid tip pipeline: it keeps the same global \texttt{TIP.md} guidance and additionally retrieves the top-$k=2$ task-versioned tips from the MemGit store using BM25 relevance to the current task. Therefore, EvoMem changes only the memory retrieval layer, while the task-solving and optimization loop remain identical across settings.

\subsection{LoCoMo}
\label{app:setting_locomo}
For \emph{LoCoMo}, we use A-Mem as the baseline memory agent with memory retrieval top-$k=10$. The A-Mem baseline does not use patch retrieval. For \emph{A-Mem + EvoMem}, we keep the same memory retrieval budget, set patch retrieval top-$k=2$, and use a patch similarity threshold of $0.0$. For answer generation, we use temperature $0.5$ for category-5 questions and the default temperature otherwise. The memory-building temperature is set to $0.7$ for both the baseline and EvoMem settings.

\section{Example of EvoArena}

\subsection{Terminal-Bench-Evo} \begin{tcolorbox}[ colback=gray!4, colframe=black!75, title={Example: Terminal-Bench-Evo configure-git-webserver workflow}, fonttitle=\bfseries, breakable, enhanced, before skip=4pt, after skip=4pt, enlargepage flexible, pad at break*=1mm ] \small \textbf{Evaluation setting.} The agent operates in a Linux terminal environment and must configure a small git-backed web deployment service. The stable objective is to let a user push an HTML file to a remote git repository and then retrieve the deployed page from a web server. Across variants, the same high-level service goal is preserved, but the required deployment mechanism, filesystem target, permission model, and branch policy evolve. \vspace{0.5em} \textbf{Stable workflow objective.} \begin{quote} Configure the server so that pushing \texttt{hello.html} to the git repository automatically makes the file available from the web server on port 8080. \end{quote} \vspace{0.5em} \textbf{Evo stages.} \begin{itemize}[leftmargin=1.5em] \item \textbf{Prototype: Basic git-to-web deployment.} The agent configures a git server and a web server so that a user can clone the repository, commit \texttt{hello.html}, push to \texttt{master}, and then retrieve the page through HTTP. \item \textbf{EVO-1: Hook-based deployment with transient rejection.} The service objective remains the same, but deployment must now happen through a git \texttt{post-receive} hook rather than manual file copying. The environment provides hook templates, and the first push is expected to fail because of the deployment logic before a retry succeeds. The deployed web root must also contain a marker showing that deployment happened through the hook. \item \textbf{EVO-2: Reconcile serving path and deployment path.} The workflow still requires automatic deployment after git push, but the environment introduces an intentional mismatch between template assumptions and the actual web-serving directory. The agent must align the hook deployment target with the directory served by nginx, rather than blindly using the previous web root. \item \textbf{EVO-3: Secure web root and group-permission deployment.} The task adds a permission policy. The web root moves to a secure directory owned by \texttt{root:www-data} with group- write deployment semantics. The git user must deploy through group membership, so a solution that only changes paths without fixing ownership and permissions is insufficient. \item \textbf{EVO-4: Main-branch-only deployment.} The git workflow changes from \texttt{master} to \texttt{main}. Pushes to \texttt{master} must be rejected server-side, and deployment should occur only for \texttt{main}. The secure web root and group-permission model from the previous stage must still be preserved. \end{itemize} \vspace{0.5em} \textbf{Why this is an evolution chain.} \begin{quote} A solution that manually copies files into the web root may satisfy the prototype but fails once deployment must be mediated by git hooks. A hook that deploys to the old web root fails when nginx serves a different directory. A path-only adaptation fails when the task introduces group-based permission constraints, and a \texttt{master}-based deployment fails once the server enforces a \texttt{main}-only branch policy. Thus, the user-facing goal remains stable, but the correct terminal strategy must be patched across deployment mechanism, filesystem layout, permission model, and branch semantics. \end{quote} \end{tcolorbox}

 \subsection{SWE-Chain-Evo}

  \begin{tcolorbox}[
      colback=gray!4,
      colframe=black!75,
      title={Example: SWE-Chain-Evo aiohttp protocol-boundary hardening chain},
      fonttitle=\bfseries,
      breakable,
      enhanced,
      before skip=4pt,
      after skip=4pt,
      enlargepage flexible,
      pad at break*=1mm
  ]
  \small

  \textbf{Evaluation setting.}
  The agent works on \texttt{aiohttp}, a Python HTTP client/server library. The chain contains four software-engineering
milestones. Each milestone targets a different boundary where external input enters the library: persisted cookies, multipart
form headers, digest authentication challenges, and access-log timestamp formatting. The stable engineering objective is to
harden protocol-facing behavior while preserving existing compatibility.

  \vspace{0.5em}
  \textbf{Stable workflow objective.}
  \begin{quote}
  Improve aiohttp's handling of externally supplied protocol data and runtime environment details without breaking existing safe
behavior.
  \end{quote}

  \vspace{0.5em}
  \textbf{Evo stages.}

  \begin{itemize}[leftmargin=1.5em]
      \item \textbf{Milestone 1: Safe legacy cookie loading.}
      \texttt{CookieJar.load()} supports existing saved cookie files, but the legacy pickle path may deserialize arbitrary
Python objects. The agent must add a restricted legacy loader that accepts only expected cookie container types, rejects
malicious payloads, and keeps safe JSON cookie save/load behavior working.

      \item \textbf{Milestone 2: Multipart header-injection guard.}
      \texttt{FormData.add\_field()} already validates non-string \texttt{content\_type} values, but string values containing
carriage returns or newlines can still be accepted and later injected into multipart headers. The agent must reject CR/LF-
bearing content types while preserving the existing type-error behavior for non-string inputs.

      \item \textbf{Milestone 3: Digest challenge parsing with empty values.}
      Digest authentication challenge parsing drops fields whose value is the empty string. Since RFC 7616 permits an empty
realm, the parser should retain fields such as \texttt{realm=""} instead of discarding them. The agent must preserve empty-
string challenge values without changing normal challenge parsing.

      \item \textbf{Milestone 4: Access-log timezone correctness.}
      \texttt{AccessLogger} formats timestamps using a constant timezone offset, which can produce incorrect logs across
daylight-saving transitions. The agent must recompute and cache the local timezone from the current localtime offset while
preserving existing access-log atom formatting.
  \end{itemize}

  \vspace{0.5em}
  \textbf{Why this is an evolution chain.}
  \begin{quote}
  The milestones are not isolated toy edits: they form a coherent robustness chain over aiohttp's protocol and environment
boundaries. A solution pattern that only adds input rejection is appropriate for malicious cookie payloads and header injection,
but it would be wrong for digest authentication, where the correct behavior is to preserve an allowed empty value. Similarly,
timestamp correctness requires adapting runtime-derived state rather than parsing user protocol input. Thus, the chain preserves
a stable engineering theme while forcing the agent to distinguish unsafe input rejection, compatibility-preserving parsing, and
environment-sensitive formatting.
  \end{quote}

  \end{tcolorbox}

\subsection{PersonaMem-Evo}

\begin{tcolorbox}[
    colback=gray!4,
    colframe=black!75,
    title={Example: PersonaMem-Evo OOD item with temporal preference evolution},
    fonttitle=\bfseries,
    breakable,
    enhanced,
    before skip=4pt,
    after skip=4pt,
    enlargepage flexible,
    pad at break*=1mm
]
\small

\textbf{Evaluation setting.}
The full history for this persona contains hundreds of turns (over $2\times 10^5$ tokens). We show only the relevant snippets and a few distracting snippets; the model receives the full history.

\vspace{0.5em}
\textbf{Relevant context excerpts.}

\begin{itemize}[leftmargin=1.5em]
    \item \textbf{Earlier preference state.}
    \begin{quote}
    \textit{User:} ``I'm trying to make grocery shopping less chaotic \ldots\ Do you have suggestions for supermarkets that work best if I want straightforward staples, clear labeling, and not too many random specialty ingredients?''

    \textit{Assistant:} ``\ldots\ Mainstream supermarkets with predictable layouts \ldots\ clear aisle signage \ldots\ easy-to-find regular ingredients rather than too many specialty items \ldots''
    \end{quote}

    \item \textbf{Later preference state.}
    \begin{quote}
    \textit{User:} ``I've been trying to branch out with groceries more \ldots\ I keep seeing sumac, tahini, preserved lemons, and different kinds of chili crisp at the international market near me. Can you give me a practical guide to what these taste like and how to use them?''

    \textit{Assistant:} ``Absolutely \ldots\ those are all very usable ingredients \ldots\ Sumac: tangy, lemony \ldots\ Tahini: nutty, earthy \ldots\ preserved lemons \ldots\ fermented sauces \ldots''
    \end{quote}
\end{itemize}

\vspace{0.5em}
\textbf{Distracting context excerpts.}

\begin{itemize}[leftmargin=1.5em]
    \item
    \begin{quote}
    \textit{User:} ``Why do some brunch places feel heavier than others even when the menu sounds similar?''
    \end{quote}

    \item
    \begin{quote}
    \textit{User:} ``When a brunch menu is huge, is there a good way to tell whether a place will feel lighter and more balanced?''
    \end{quote}
\end{itemize}

\vspace{0.5em}
\textbf{OOD question (temporal trajectory, L1).}

\begin{quote}
\textit{She wants to spend a Saturday afternoon food-shopping for a cooking project. Which kind of store and approach would fit her best?}
\end{quote}

\textbf{Answer options.}
\begin{enumerate}[label=(\Alph*), leftmargin=2em]
    \item She would most likely choose a standard supermarket, stick to clearly labeled basics, and avoid anything too unfamiliar so the cooking project feels predictable and easy to manage.
    \item \textbf{She would probably head to a specialty or international market and enjoy asking questions, browsing regional pantry items, and picking up a few unfamiliar ingredients to build the meal around.}
    \item She would likely start at a regular grocery store for most items, then maybe add one stop at a specialty shop for a single recommended ingredient if it seemed approachable.
    \item She would probably pick whichever nearby grocery store is most convenient, use a recipe with easy-to-find ingredients, and keep the trip simple so it does not take too much energy.
\end{enumerate}

\textbf{Correct answer.}
\begin{quote}
\textbf{(B)} The later conversation shows that the user has shifted from preferring mainstream supermarkets to actively exploring specialty or international markets and unfamiliar ingredients. The item therefore requires using the updated preference state rather than the earlier one.
\end{quote}

\end{tcolorbox}

%%%%%%%%%%%%%%%%%%%%%%%%%%%%%%%%%%%%%%%%%%%%%%%%%%%%%%%%%%%%

% \newpage
% \input{checklist.tex}

\end{document}